\documentclass[times,twocolumn,final]{elsarticle}
\usepackage{medima_arxiv}
\usepackage{framed,multirow}
\usepackage{amssymb}
\usepackage{latexsym}
\usepackage{url}
\usepackage{xcolor}
\usepackage{graphicx}
\usepackage{verbatim}
\usepackage{amsfonts}
\usepackage{xcolor}
\usepackage{float}
\usepackage{algorithmic}
\usepackage{algorithm}
\usepackage{amsmath}
\usepackage{booktabs}
\DeclareMathOperator*{\argmax}{argmax}
\usepackage{tabulary}
\usepackage{colortbl}
\usepackage{etoolbox}
\usepackage{multicol}
\usepackage{lipsum}
\usepackage{subcaption}
\usepackage{wrapfig}
\usepackage{makecell}
\usepackage{capt-of}
\usepackage{makecell}
\usepackage{tablefootnote}
\usepackage[T1]{fontenc}
\usepackage[utf8]{inputenc}
\usepackage[font=small,labelfont=bf]{caption}
\usepackage[colorlinks=true,linkcolor=blue,urlcolor=blue,citecolor=blue]{hyperref}

\newcommand\blfootnote[1]{%
  \begingroup
  \renewcommand\thefootnote{}\footnote{#1}%
  \addtocounter{footnote}{-1}%
  \endgroup
}

\definecolor{newcolor}{rgb}{.8,.349,.1}

\journal{}

\begin{document}

\verso{Yiqiu Shen \textit{et~al.}}

\begin{frontmatter}

\title{An interpretable classifier for high-resolution breast cancer screening images utilizing weakly supervised localization}%

\author[1]{Yiqiu Shen}
\author[1]{Nan Wu}
\author[1]{Jason Phang}
\author[1]{Jungkyu Park}
\author[1]{Kangning Liu}
\author[4]{Sudarshini Tyagi}
\author[2,5]{Laura Heacock}
\author[2,3,5]{S. Gene Kim}
\author[2,3,5]{Linda Moy}
\author[1,4,6]{Kyunghyun Cho}
\author[2,3,1]{Krzysztof J. Geras}

\address[1]{Center for Data Science, New York University}
\address[2]{Department of Radiology, NYU School of Medicine}
\address[3]{Center for Advanced Imaging Innovation and Research, NYU Langone Health}
\address[4]{Department of Computer Science, Courant Institute, New York University}
\address[5]{Perlmutter Cancer Center, NYU Langone Health}
\address[6]{CIFAR Associate Fellow}

\begin{abstract}
 Medical images differ from natural images in significantly higher resolutions and smaller regions of interest. Because of these differences, neural network architectures that work well for natural images might not be applicable to medical image analysis. In this work, we extend the globally-aware multiple instance classifier, a framework we proposed to address these unique properties of medical images. This model first uses a low-capacity, yet memory-efficient, network on the whole image to identify the most informative regions. It then applies another higher-capacity network to collect details from chosen regions. Finally, it employs a fusion module that aggregates global and local information to make a final prediction. While existing methods often require lesion segmentation during training, our model is trained with only image-level labels and can generate pixel-level saliency maps indicating possible malignant findings. We apply the model to screening mammography interpretation: predicting the presence or absence of benign and malignant lesions. On the NYU Breast Cancer Screening Dataset, consisting of more than one million images, our model achieves an AUC of 0.93 in classifying breasts with malignant findings, outperforming ResNet-34 and Faster R-CNN. Compared to ResNet-34, our model is 4.1x faster for inference while using 78.4\% less GPU memory. Furthermore, we demonstrate, in a reader study, that our model surpasses radiologist-level AUC by a margin of 0.11. The proposed model is \href{https://www.github.com/nyukat/GMIC}{available online}.
\end{abstract}

\begin{keyword}
\KWD deep learning\sep breast cancer screening\sep weakly supervised localization\sep high-resolution image classification
\end{keyword}

\end{frontmatter}

\blfootnote{This paper is an extension of work originally presented at the 10th International Workshop on Machine Learning in Medical Imaging~\citep{shen2019globally}.}

\vspace{-10mm}
\section{Introduction}

\begin{figure*}[ht]
  \centering
 \includegraphics[width=0.99\textwidth,trim={0 0 0 0},clip]{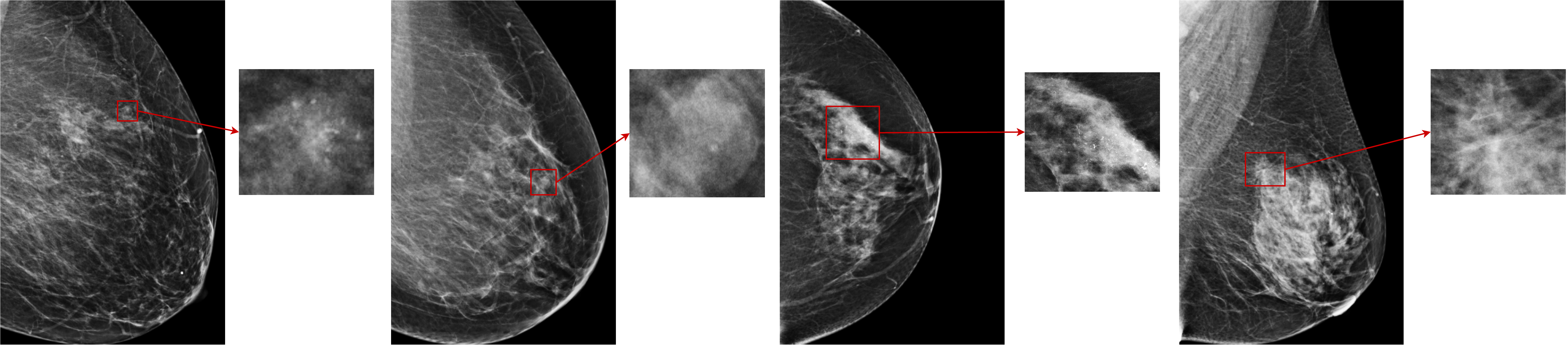}
  \caption{Four examples of breasts that were biopsied along with the annotated findings. The breasts (from left to right) were diagnosed with benign calcifications, a benign mass, malignant calcifications, and malignant architectural distortion. While microcalcifications are common in both benign and malignant findings, their presence in a ductal distribution, such as in the third example, is a strong indicator of malignancy.}
  \label{fig:lesion_example}
\end{figure*}

Breast cancer is the second leading cause of cancer-related death among women in the United States~\citep{desantis2017breast}. It was estimated that 268,600 women would be diagnosed with breast cancer and 41,760 would die in 2019~\citep{siegel2019cancer}. Screening mammography, a low-dose X-ray examination, is a major tool for early detection of breast cancer. A standard screening mammogram consists of two high-resolution X-rays of each breast, taken from the side (the “mediolateral” or MLO view) and from above (the “craniocaudal” or CC view) for a total of four images. Radiologists, physicians specialized in the interpretation of medical images, analyze screening mammograms for tissue abnormalities that may indicate breast cancer. Any detected abnormality leads to additional diagnostic imaging and possible tissue biopsy. A radiologist assigns a standardized assessment to each screening mammogram per the American College of Radiology Breast Imaging Reporting and Data System (BI-RADS), with specific follow-up recommendations for each category~\citep{liberman2002breast}. 

Screening mammography interpretation is a particularly challenging task because mammograms are in very high resolutions while most asymptomatic cancer lesions are small, sparsely distributed over the breast and may present as subtle changes in the breast tissue pattern. While randomized clinical trials have shown that screening mammography has significantly reduced breast cancer mortality~\citep{duffy2002impact,RN40}, it is associated with limitations such as false positive recalls for additional imaging and subsequent false positive biopsies which result in benign, non-cancerous findings. About 10\% to 20\% of women who have an abnormal screening mammogram are recommended to undergo a biopsy. Only 20\% to 40\% of these biopsies yield a diagnosis of cancer~\citep{kopans2015open}.

To tackle these limitations, convolutional neural networks (CNN) have been applied to assist radiologists in the analysis of screening mammography~\citep{zhu2017deep, kim2018applying,mammo, ribli2018detecting, wu2019deep, mckinney2020international}. An overwhelming majority of existing studies on this task utilize models that were originally designed for natural images. For instance, VGGNet~\citep{simonyan2014very}, designed for object classification on ImageNet~\citep{deng2009imagenet}, has been applied to breast density classification~\citep{wu2018breast} and Faster R-CNN~\citep{ren2015faster} has been adapted to localize suspicious findings in mammograms~\citep{ribli2018detecting,fevry2019improving}.

Screening mammography is inherently different from typical natural images from a few perspectives. First of all, as illustrated in Figure \ref{fig:lesion_example}, regions of interest (ROI) in mammography images, such as masses, asymmetries, and microcalcifications, are often smaller in comparison to the salient objects in natural images. Moreover, as suggested in multiple clinical studies~\citep{van1998mammographic,pereira2009spatial,wei2011association}, both the local details, such as lesion shape, and global structure, such as overall breast fibroglandular tissue density and pattern, are essential for accurate diagnosis. For instance, while microcalcifications are common in both benign and malignant findings, their presence in a ductal distribution, such as in the third example of Figure~\ref{fig:lesion_example}, is a strong indicator of malignancy. This is in contrast to typical natural images where objects outside the most salient regions provide little information towards predicting the label of the image. In addition, mammography images are usually of much higher resolutions than typical natural images. The most accurate deep CNN architectures for natural images are not applicable to mammography images due to the limited size of GPU memory. 

To address the aforementioned issues, in this work, we extend and comprehensively evaluate the globally-aware multiple instance classifier (GMIC), whose preliminary version we proposed in~\cite{shen2019globally}. GMIC first applies a low-capacity, yet memory-efficient, global module on the whole image to generate saliency maps that provide coarse localization of possible benign/malignant findings. As a result, GMIC is able to process screening mammography images in their original resolutions while keeping GPU memory manageable. In order to capture subtle patterns contained in small ROIs, GMIC then identifies the most informative regions in the image and utilizes a local module with high-capacity to extract fine-grained visual details from these regions. Finally, it employs a fusion module that aggregates information from both global context and local details to predict the presence or absence of benign and malignant lesions in a breast. The specific contributions of this work are the following:
\begin{itemize}
    \item We extend the original architecture~\citep{shen2019globally} with a fusion module. The fusion module improves classification performance by effectively combining information from both global and local features. In Section \ref{sec:ablation_study}, we demonstrate that the fusion module renders more accurate predictions than our original design.

    \item We apply the improved model to the task of screening mammography interpretation: predicting the presence or absence of benign and malignant lesions in a breast. We trained and tested our model on the NYU Breast Cancer Screening Dataset consisting of 229,426 high-resolution screening mammograms~\citep{NYU_dataset}. On a held-out test set of 14,148 exams, GMIC achieves an AUC of 0.93 in identifying breasts with malignant findings, outperforming baseline approaches including ResNet-34~\citep{he2016deep}, Faster R-CNN~\citep{fevry2019improving} and DMV-CNN~\citep{wu2019deep}. 
    
    \item We demonstrate the clinical potential of the GMIC by comparing the improved model to human experts. In the reader study, we show that it surpasses a radiologist-level classification performance: the AUC for the proposed model was greater than the average AUC for radiologists by a margin of 0.11, reducing the error approximately by half. In addition, we experiment with hybrid models that combine predictions from both GMIC and each of the radiologists separately. At radiologists' sensitivity ($62.1\%$), the hybrid models achieve an average specificity of $91.9\%$ improving radiologists' average specificity by $6.3\%$. 
    
    \item The proposed model is able to localize breast lesions in a weakly supervised manner, unlike existing approaches that rely on pixel-level lesion annotations \citep{ribli2018detecting,fevry2019improving, wu2019deep}. In Section~\ref{sec:localization_performance}, we demonstrate that the regions highlighted by the saliency maps indeed correlate with the objects of interest. 
    
   \item We demonstrate that the proposed model is computationally efficient. GMIC requires significantly less memory and is much faster to train than standard image classification models such as ResNet-34~\citep{he2016deep} and Faster R-CNN~\citep{ren2015faster}. Benchmarked on high-resolution screening mammography images, GMIC has $\mathbf{28.8\%}$ fewer parameters, uses $\mathbf{78.4\%}$ less GPU memory, is \textbf{4.1x} faster during inference and \textbf{5.6x} faster during training, as compared to ResNet-34, while being more accurate.
   
  \item We conduct a comprehensive ablation study that evaluates the effectiveness of each component of GMIC. Moreover, we empirically measure how much performance can be improved by ensembling GMIC with Faster R-CNN and ResNet-34. In addition, we also experiment with utilizing segmentation labels to enhance GMIC. In both experiments, we find that the improvement is marginal, suggesting that, for a large training set, image-level labels alone are sufficient for GMIC to reach favorable performance.
\end{itemize}

\section{Methods}

\begin{figure*}
  \centering
 \includegraphics[width=0.6\textwidth,trim={0 0 100pt 0}]{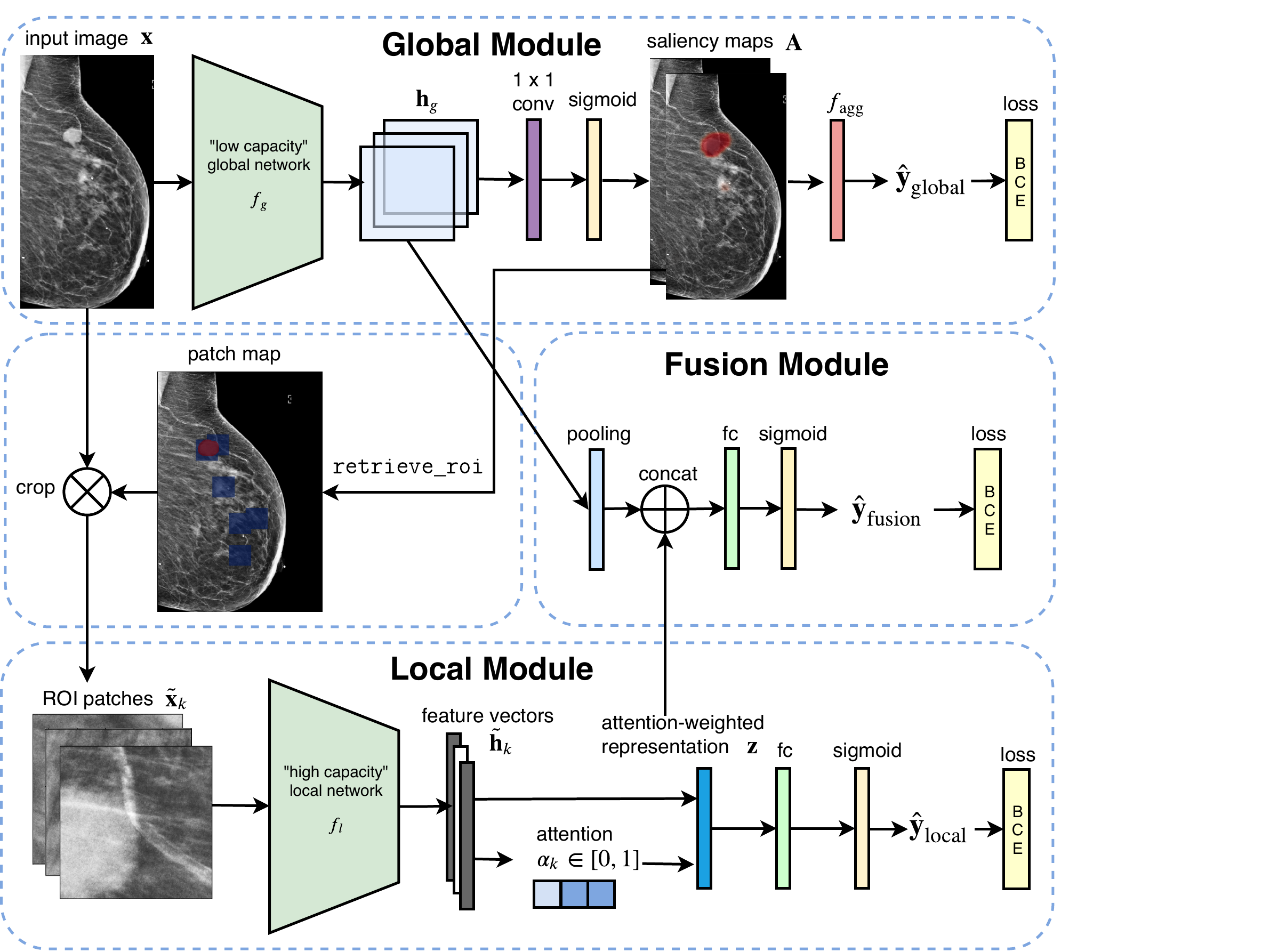}\qquad\qquad
  \includegraphics[width=0.184\textwidth]{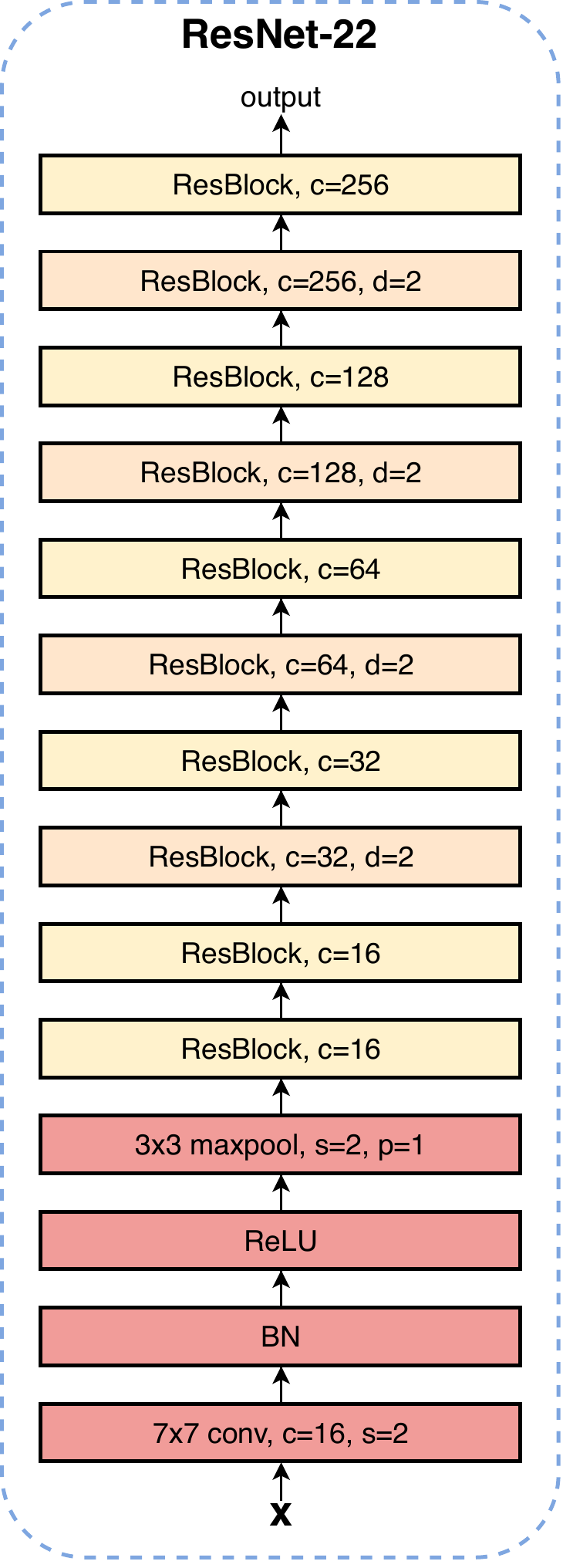}
  \caption{Overall architecture of GMIC (left) and architecture of ResNet-22 (right). The patch map indicates positions of ROI patches (blue squares) on the input. In ResNet-22, we use $c$, $s$, and $p$ to denote number of output channels, strides and size of padding. ``ResBlock, c=32, d=2'' denotes a vanilla ResBlock proposed in~\cite{he2016identity} with 32 output channels and a downsample skip connection that reduces the resolution with a factor of 2. In comparison to canonical ResNet architectures~\citep{he2016deep}, ResNet-22 has one more residual block and only a quarter of the filters in each convolution layer. Narrowing network width decreases the total number of hidden units which reduces GPU memory consumption.}
  \label{fig:overall_plot}
\end{figure*}

We frame the task of screening mammography interpretation as a multi-label classification problem: given a grayscale image $\mathbf{x} \in \mathbb{R}^{H,W}$, we predict the image-level label $\mathbf{y} =  \begin{bmatrix} y^b \\ y^m \end{bmatrix}$, where $y^b, y^m \in \{0,1\}$ indicate whether any benign/malignant lesion is present in $\mathbf{x}$.

\subsection{Globally-Aware Classification Framework}
\label{sec:overall_structure}
As shown in Figure \ref{fig:overall_plot}, we propose a classification framework that resembles the diagnostic procedure of a radiologist. We first use a global network $f_g$ to extract a feature map $\mathbf{h}_g$ from the input image $\mathbf{x}$, i.e. we compute
\begin{equation}
    \mathbf{h}_g = f_g(\mathbf{x}),
\end{equation}
which is analogous to a radiologist roughly scanning through the entire image to obtain a holistic view.

We then apply a $1\times1$ convolution layer with sigmoid non-linearity to transform $\mathbf{h}_g$ into two saliency maps $\mathbf{A}^b, \mathbf{A}^m \in \mathbb{R}^{h,w}$ indicating approximate locations of benign and malignant lesions.\footnote{Depending on the implementation of $f_g$, the resolutions of the saliency maps ($h,w$) are usually smaller than the resolution of the input image ($H,W$). In this work, we set $h=46$, $w=30$, $H=2944$, and $W=1920$.} Each element $\mathbf{A}^c_{i,j} \in [0,1]$ where $c \in \{b, m\}$, denotes the contribution of spatial location $(i,j)$ towards predicting the presence of benign/malignant lesions. Let $\mathbf{A}$ denote the concatenation of $\mathbf{A}^b$ and $\mathbf{A}^m$. That is, we compute $\mathbf{A}$ as
\begin{equation}
    \mathbf{A} = \text{sigm}(\text{conv}_{1\times1}(\mathbf{h}_g)).
\end{equation}

Due to limited GPU memory, in prior work, input images $\mathbf{x}$ are usually down-sampled \citep{guan2018diagnose, yao2018weakly, zhong2019attention}. For mammography images, however, down-sampling distorts important visual details such as lesion margins and blurs small microcalcifications. Instead of sacrificing the input resolution, we control memory consumption by reducing the complexity of the \emph{global} network $f_g$. Because of its constrained capacity, $f_g$ may not be able to capture all subtle patterns contained in the images at all scales. To compensate for this, we utilize a high-capacity \emph{local} network $f_l$ to extract fine-grained details from a set of informative regions. In the second stage, we use $\mathbf{A}$ to retrieve $K$ most informative patches from $\mathbf{x}$:
\begin{equation}
    \{\tilde{\mathbf{x}}_k\} = \texttt{retrieve\_roi}(\mathbf{A}),
\end{equation}
where $\texttt{retrieve\_roi}$ denotes a heuristic patch-selection procedure described later. This procedure can be seen as an analogue to a radiologist concentrating on areas that might correspond to lesions. The fine-grained visual features $\{\tilde{\mathbf{h}}_k\}$ contained in all chosen patches $\{\tilde{\mathbf{x}}_k\}$ are then processed using $f_l$ and are aggregated into a vector $\mathbf{z}$ by an aggregator $f_a$. That is,
\begin{equation}
    \tilde{\mathbf{h}}_k = f_l(\tilde{\mathbf{x}}_k) \quad\text{and}\quad \mathbf{z} = f_a(\{\tilde{\mathbf{h}}_k\}).
\end{equation}
Finally, a fusion network $f_\text{fusion}$ combines information from both global structure $\mathbf{h}_g$ and local details $\mathbf{z}$ to produce a prediction $\hat{\mathbf{y}}$. This is analogous to modelling a radiologist comprehensively considering the global and local
information to render a full diagnosis as
\begin{equation}
    \hat{\mathbf{y}} = f_\text{fusion}(\mathbf{h}_g, \mathbf{z}).
\end{equation}

\subsection{Model Parameterizaiton} \label{sec:model_parameterization}

\paragraph{Generating the Saliency Maps}
To process high-resolution images while keeping GPU memory consumption manageable, we parameterize $f_g$ as a ResNet-22~\citep{wu2019deep} whose architecture is shown in Figure \ref{fig:overall_plot}. In comparison to canonical ResNet architectures~\citep{he2016deep}, ResNet-22 has one more residual block and only a quarter of the filters in each convolution layer. As suggested by~\cite{tan2019efficientnet}, a deeper CNN has larger receptive fields and can capture richer and more complex features in high-resolution images. Narrowing network width can decrease the total number of hidden units which reduces GPU memory consumption.

It is difficult to define a loss function that directly compares saliency maps $\mathbf{A}$ and the cancer label $\mathbf{y}$, since $\mathbf{y}$ does not contain localization information. In order to train $f_g$, we use an aggregation function $f_\text{agg}(\mathbf{A}^c): \mathbb{R}^{h,w} \mapsto [0,1]$ to transform a saliency map into an image-level class prediction:
\begin{equation}
    \hat{\mathbf{y}}_\text{global}^c = f_\text{agg}(\mathbf{A}^c).
\end{equation}
With $f_\text{agg}$ we can train $f_g$ by backpropagating the gradient of the classification loss between $\mathbf{y}$ and $\hat{\mathbf{y}}_\text{global}$. The design of $f_\text{agg}(\mathbf{A}^c)$ has been extensively studied~\citep{durand2017wildcat}. Global average pooling (GAP) would dilute the prediction as most of the spatial locations in $\mathbf{A}^c$ correspond to background and provide little training signal. On the other hand, with global max pooling (GMP), the gradient is backpropagated through a single spatial location, which makes the learning process slow and unstable. In our work, we propose, \textit{top $t\%$ pooling}, which is a soft balance between GAP and GMP. Namely, we define the aggregation function as
\begin{equation}
    f_{\text{agg}}(\mathbf{A}^c) = \frac{1}{|H^+|}\sum_{(i,j) \in H^+} \mathbf{A}^c_{i,j},
\end{equation}
where $H^+$ denotes the set containing locations of top $t\%$ values in $\mathbf{A}^c$, where $t$ is a hyperparameter. In all experiments, we tune $t$ using a procedure described in Section~\ref{sec:classification_performance}. In fact, GAP and GMP can be viewed as two extremes of \textit{top $t\%$ pooling}. GMP is equivalent to setting $t = \frac{1}{h \times w}$ and GAP is equivalent to setting $t = 100\%$. In Section~\ref{sec:ablation_study}, we study the impact of $t$ and empirically demonstrate that our parameterization of $f_\text{agg}$ achieves performance superior to GAP and GMP.

\paragraph{Acquiring ROI Patches}
We designed a greedy algorithm (Algorithm~\ref{alg:roi}) to retrieve $K$ patches as proposals for ROIs, $\tilde{\mathbf{x}}_k \in \mathbb{R}^{h_c,w_c}$, from the input $\mathbf{x}$, where $w_c = h_c = 256$ in all experiments. In each iteration, \texttt{retrieve\_roi} greedily selects the rectangular bounding box that maximizes the criterion defined in line 7. The algorithm then maps each selected bounding box to its corresponding location on the input image. The reset rule in line 12 explicitly ensures that extracted ROI patches do not significantly overlap with each other. In Section~\ref{sec:ablation_study}, we show how the classification performance is impacted by $K$.

\renewcommand{\algorithmicrequire}{\textbf{Input:}}
\renewcommand{\algorithmicensure}{\textbf{Output:}}
\begin{algorithm}[t]
    \caption{\texttt{retrieve\_roi}}
    \label{alg:roi}
    \begin{algorithmic}[1]
        \REQUIRE  $\mathbf{x} \in \mathbb{R}^{H,W}$, $\mathbf{A} \in \mathbb{R}^{h,w,2}$, $K$
        \ENSURE $ O = \{ \tilde{\mathbf{x}}_k |  \tilde{\mathbf{x}}_k \in \mathbb{R}^{h_c,w_c} \}$
        \STATE{$O = \emptyset$}
        \FOR{each class $c \in \{\text{benign}, \text{malignant}\}$}
            \STATE{$\mathbf{\tilde{A}}^c = \text{min-max-normalization}(\mathbf{A}^c)$}
         \ENDFOR\\
         \STATE{$ \mathbf{A}^{*} = \sum_{c} \tilde{\mathbf{A}}^c$}
         \STATE{$l$ denotes an arbitrary $h_c \frac{h}{H} \times w_c \frac{w}{W}$ rectangular patch on $\mathbf{A}^{*}$}
         \STATE $\text{criterion}(l, \mathbf{A}^{*}) = \sum_{(i,j) \in l} \mathbf{A}^{*}[i,j]$
        \FOR{each $1,2,...,K$}
            \STATE{$l^* = \argmax_{l} \text{criterion}(l, \mathbf{A}^{*})$}
            \STATE{$L = $ position of $l^*$ in $\mathbf{x}$}
            \STATE{$O = O \cup \{L\}$}
            \STATE{ $\forall (i,j) \in l^*$, set $\mathbf{A}^*[i,j]=0$}
        \ENDFOR
        \RETURN $O$
    \end{algorithmic}
\end{algorithm}

\paragraph{Utilizing Information from Patches}
With \texttt{retrieve\_roi}, we can focus learning on a selected set of small yet informative patches $\{\tilde{\mathbf{x}}_k\}$. We can now apply a local network $f_l$ with higher capacity (wider or deeper) that is able to utilize fine-grained visual features, to extract a representation $\tilde{\mathbf{h}}_k \in \mathbb{R}^L$ from every patch $\tilde{\mathbf{x}}_k$. We experiment with several parameterizations of $f_l$ including ResNet-18, ResNet-34 and ResNet-50.

Since ROI patches are retrieved using coarse saliency maps, the information relevant for classification carried in each patch varies significantly. To address this issue, we use the Gated Attention Mechanism (GA)~\citep{ilse2018attention}, allowing the model to selectively incorporate information from all patches. Compared to other common attention mechanisms~\citep{bahdanau2014neural,luong2015effective}, GA uses the sigmoid function to provide a learnable non-linearity which increases model flexibility. An attention score $\alpha_k$ is computed on each patch:
\begin{equation}
    \alpha_k = \frac{\text{exp}\{\mathbf{w}^\intercal (\text{tanh}(\mathbf{V}\mathbf{\tilde{h}}_k^{\intercal}) \odot \text{sigm}(\mathbf{U}\mathbf{\tilde{h}}_k^{\intercal}) )\}}{\sum^K_{j=1}\text{exp}\{\mathbf{w}^\intercal (\text{tanh}(\mathbf{V}\mathbf{\tilde{h}}_j^{\intercal}) \odot \text{sigm}(\mathbf{U}\mathbf{\tilde{h}}_j^{\intercal}) )\}},
\end{equation}
where $\odot$ denotes an element-wise multiplication and $\mathbf{w} \in \mathbb{R}^{L}$, $\mathbf{V} \in \mathbb{R}^{L \times M}$, $\mathbf{U} \in \mathbb{R}^{L \times M}$ are learnable parameters. In all experiments, we set $L = 512$ and $M = 128$. This process yields an attention-weighted representation
\begin{equation}
    \mathbf{z} = \sum_{k=1}^{K} \alpha_k \tilde{\mathbf{h}}_k,
\end{equation}
where the attention score $\alpha_k \in [0,1]$ indicates the relevance of each patch $\tilde{\mathbf{x}}_k$. The representation $\mathbf{z}$ is then passed to a fully connected layer with sigmoid activation to generate a prediction
\begin{equation}
    \hat{\mathbf{y}}_{\text{local}} = \text{sigm}(\mathbf{w_{\text{local}}}^T \mathbf{z}),
\end{equation}
where $ \mathbf{w}_{\text{local}} \in \mathbb{R}^{L \times 2}$ are learnable parameters.

\paragraph{Information Fusion} To combine information from both saliency maps and ROI patches, we apply a global max pooling on $\mathbf{h}_g$ and concatenate it with $\mathbf{z}$. The concatenated representation is then fed into a fully connected layer with sigmoid activation to produce the final prediction:
\begin{equation}
    \hat{\mathbf{y}}_\text{fusion} = \text{sigm}(\mathbf{w}_f[\text{GMP}(\mathbf{h}_g), \mathbf{z}]^{\intercal})
\end{equation}
where GMP denotes the global max pooling operator and $\mathbf{w}_f$ are learnable parameters.

\subsection{Learning the parameters of GMIC} \label{sec:learning}
In order to constrain the saliency maps to only highlight important regions, we impose the $L1$ regularization on $\mathbf{A}^c$ to make the saliency maps sparser:
\begin{equation}
L_\text{reg}(\mathbf{A}^c) = \sum_{(i,j)} |\mathbf{A}^c_{i,j}|.
\end{equation}
Despite the relative complexity of our proposed framework, the model can be trained end-to-end using stochastic gradient descent with following loss function, defined for a single training example as:
\begin{equation}
    \begin{split}
        L(\mathbf{y}, \hat{\mathbf{y}}) = \sum_{c \in \{b,m\}} \text{BCE}(\mathbf{y}^c, \hat{\mathbf{y}}_{\text{local}}^c) + \text{BCE}(\mathbf{y}^c, \hat{\mathbf{y}}_{\text{global}}^c) 
        \\+ \text{BCE}(\mathbf{y}^c, \hat{\mathbf{y}}_{\text{fusion}}^c) 
        +  \beta L_\text{reg}(\mathbf{A}^c),
    \end{split}
\end{equation}
where $\text{BCE}$ is the binary cross-entropy and $\beta$ is a hyperparameter.

\section{Experiments and Results}
To demonstrate the effectiveness of GMIC on high-resolution image classification, we evaluate it on the task of screening mammography interpretation: predicting the presence or absence of benign and malignant findings in a breast. We compare GMIC to a previous ResNet-like network dedicated to mammography~\citep{wu2019deep} as well as to the standard ResNet-34~\citep{he2016deep} and Faster-RCNN~\citep{ren2015faster,fevry2019improving} in terms of classification accuracy, number of parameters, computation time, and GPU memory consumption. In addition, we also evaluate the localization performance of GMIC by qualitatively and quantitatively comparing the resulting saliency maps with the ground truth segmentation provided by the radiologists.

\subsection{The NYU Breast Cancer Screening Dataset}

\begin{figure}[t]
\begin{tabular}{c c}
  \reflectbox{\includegraphics[width=0.22\textwidth]{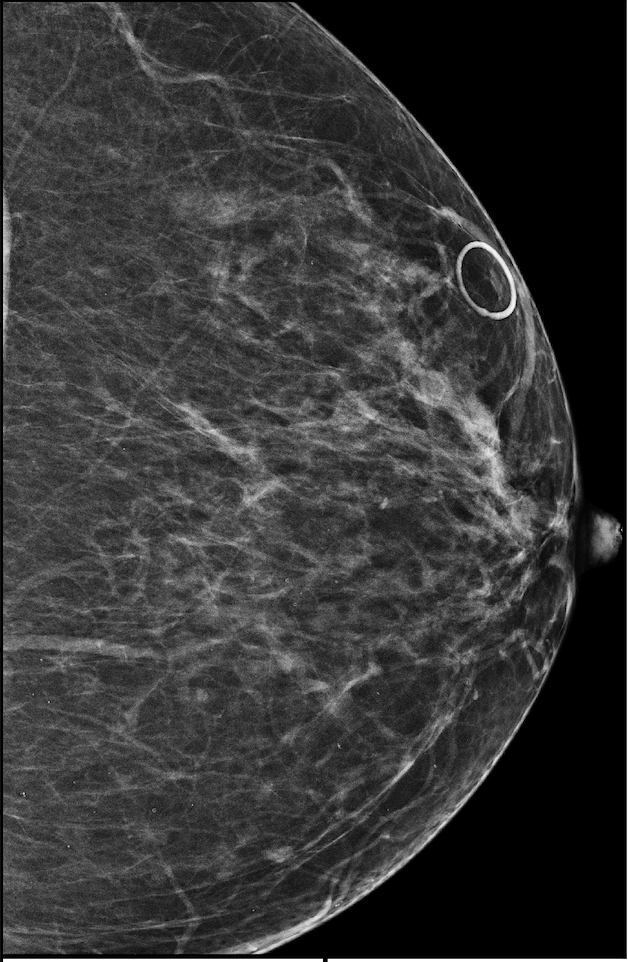}} & \includegraphics[width=0.22\textwidth]{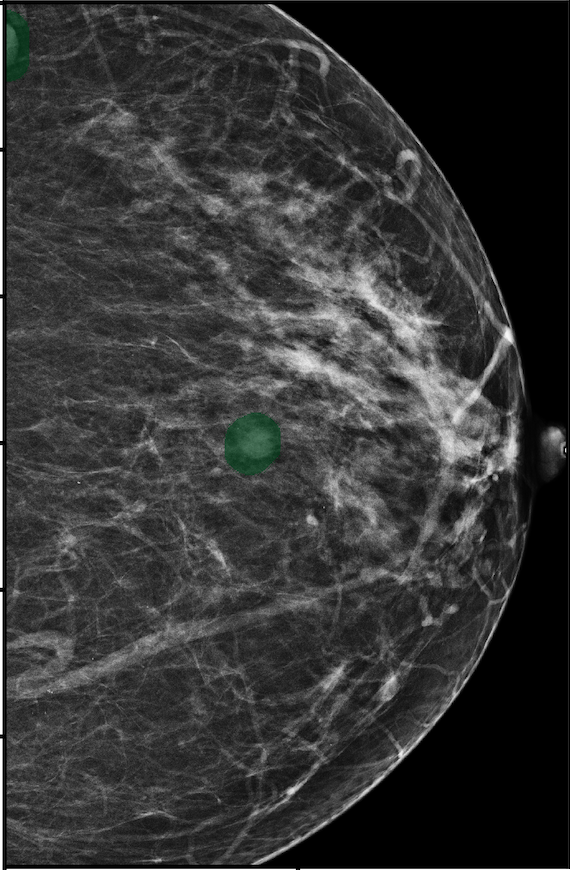}\\
    \footnotesize{R-CC} & \footnotesize{L-CC}\\
    \reflectbox{\includegraphics[width=0.22\textwidth]{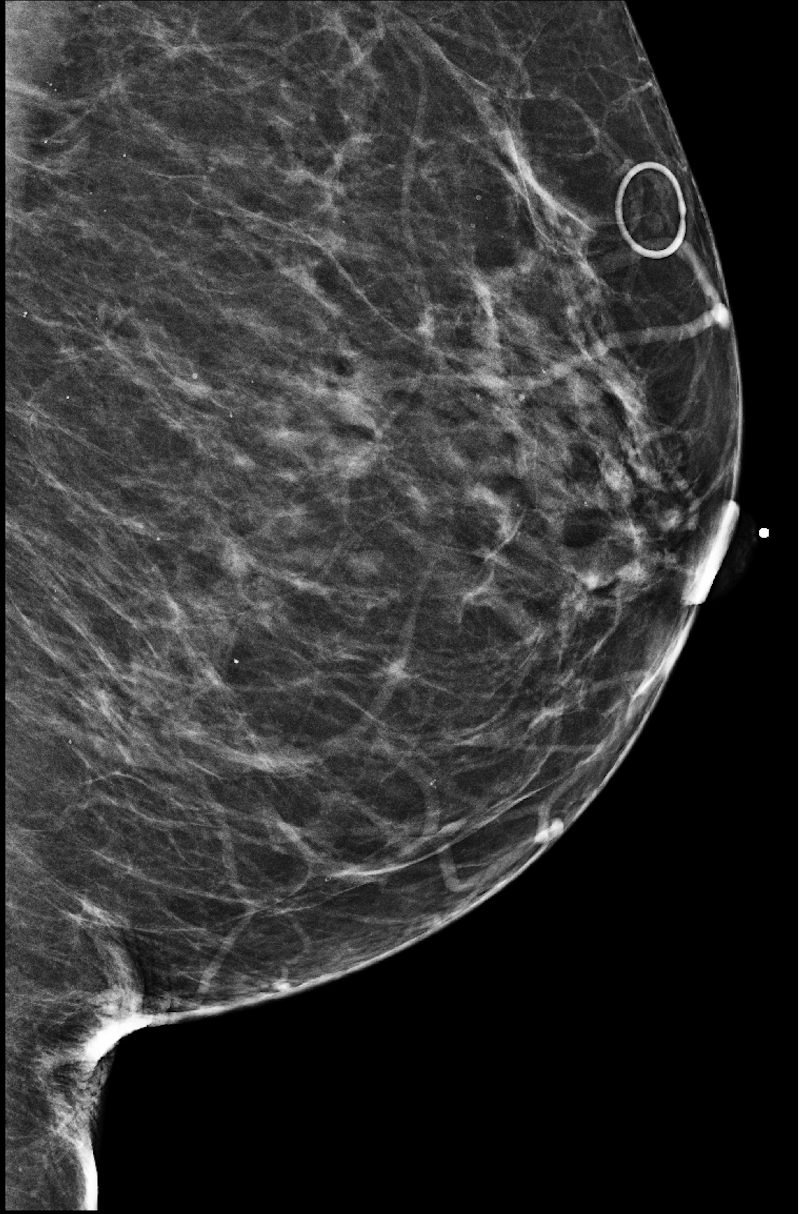}} &
    \includegraphics[width=0.22\textwidth]{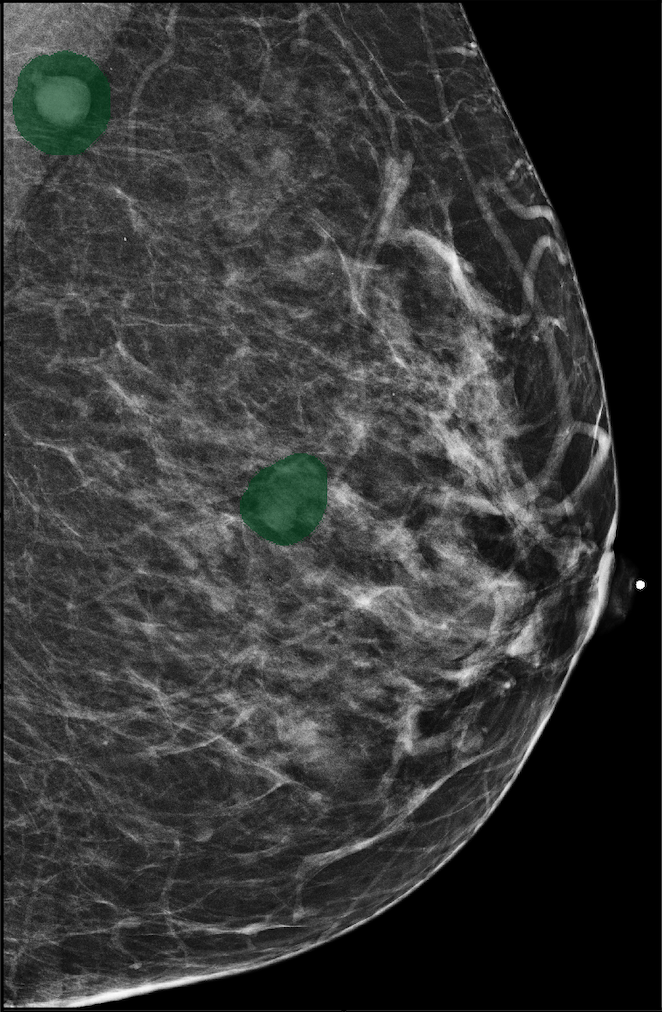} \\
    \footnotesize{R-MLO} & \footnotesize{L-MLO} \\
\end{tabular} 
\caption{Example screening mammography exam. Each exam is associated with four images that correspond to the CC and MLO view of both left and right breast. The left breast is diagnosed with benign findings which are highlighted in green.}
\label{fig:data_plot}
\end{figure}

The NYU Breast Cancer Screening Dataset~\citep{NYU_dataset} includes 229,426 exams (1,001,093 images) from 141,472 patients.\footnote{Our retrospective study was approved by our institutional review board and was compliant with the Health Insurance Portability and Accountability Act. Informed consent was waived.} Each exam contains at least four images which correspond to the four standard views used in screening mammography: R-CC (right craniocaudal), L-CC (left craniocaudal), R-MLO (right mediolateral oblique) and L-MLO (left mediolateral oblique). An example is shown in Figure \ref{fig:data_plot}.

Across the entire dataset (458,852 breasts), malignant findings were present in 985 breasts ($0.21\%$) and benign findings in 5,556 breasts ($1.22\%$). All findings are confirmed by at least one biopsy performed within 120 days of the screening mammogram. For the remaining screening exams that were not matched with a biopsy, we assigned labels corresponding to the absence of malignant and benign findings in both breasts. In each exam, the two views of the same breast share the same label.

For all exams matched with biopsies, we asked a group of radiologists (provided with the corresponding pathology reports) to retrospectively indicate the location of the biopsied lesions. This way we obtained the segmentation labels: $\mathbf{M}^b, \mathbf{M}^m \in \{0,1\}^{H \times W}$ where $\mathbf{M}^{b/m}_{i,j} = 1$ if pixel $i,j$ belongs to the benign/malignant findings. An example of such a segmentation is shown in Figure~\ref{fig:data_plot}. In all experiments (except for experiments in Section \ref{sec:ablation_study} that assess the benefits of utilizing segmentation labels), segmentation labels are only used for evaluation. We found that, according to the radiologists, approximately $32.8\%$ of exams were mammographically occult, i.e., the lesions that were biopsied were not visible on mammography, even retrospectively, and were identified using other imaging modalities: ultrasound or MRI.  

\subsection{Experimental Setup and Evaluation Metrics}
The dataset is divided into disjoint training (186,816), validation (28,462) and test (14,148) sets. All images are cropped to $2944 \times 1920$ pixels and normalized to have zero mean and unit standard deviation. We adopt the same pre-processing and augmentation (random cropping, size noise) as \cite{wu2019deep}. During test phase, we similarly apply data augmentation and average predictions over 10 random augmentations to compute the prediction for a given image. No data augmentation is used during validation. Since the classes of the images in the dataset are imbalanced, we adopted the following sampling strategy during training. In each epoch, we train the model using all exams that contain at least one benign or malignant finding and an equal number of randomly sampled negative exams. During the training phase, we also randomly rotate the selected ROI patches by $\{0, 90, 180, 270\}$ degrees with equal probability. No rotation to the patches is applied during validation and test phase.

As each breast is associated with two images (CC and MLO views) and our model generates a prediction for each image, we define breast-level predictions as the average of the two image-level predictions. For classification performance, we report area under the ROC curve (AUC) on the breast-level. In the reader study, we also use area under the precision-recall curve (PRAUC) to compare radiologists and the proposed model. We computed the radiologists’ sensitivity which served as prediction threshold to derive the specificity of GMIC. To assess statistical significance, we performed Student's t-test and used binomial proportion confidence intervals for specificity. To quantitatively evaluate our model's localization ability, we calculate the Dice similarity coefficient (DSC). The DSC values we report are computed as an average over images for which segmentation labels are available (i.e. images from breasts which have biopsied findings which were not mammographically occult).

In addition to accuracy, computation time and memory efficiency are also important for medical image analysis. To measure memory efficiency, we report the peak GPU memory usage during training as in \cite{canziani2016analysis}. Similar to \cite{schlemper2019attention}, we also report the run-time performance by recording the total number of floating-point operations (FLOPs) during inference and elapsed time for forward and backward propagation. Both memory and run-time statistics are measured by benchmarking each model on a single exam (4 images), averaged across 100 exams. All experiments are conducted on an NVIDIA Tesla V100 GPU.

\subsubsection{Implementation Details}
In all experiments, we parameterize $f_g$ as a ResNet-22 whose architecture is shown in Figure \ref{fig:overall_plot}. We pretrain $f_g$ on BI-RADS labels as described in \cite{high_resolution} and \cite{wu2019deep}. For $f_l$, we experiment with three different architectures with varying levels of complexity (ResNet-18, ResNet-34, ResNet-50). For each image, we extract $K = 6$ ROI patches. In all experiments (except the ablation study described in Section~\ref{sec:ablation_study}), we only used image-level labels to train GMIC. For all experiments, the training loss is optimized using Adam \citep{kingma2014adam} with learning rate fine-tuned as described in Section \ref{sec:classification_performance}. Our PyTorch \citep{paszke2017automatic} implementation (the code and the trained weights of the model) is available at \href{https://github.com/nyukat/GMIC}{https://github.com/nyukat/GMIC}.


\subsection{Classification Performance} \label{sec:classification_performance}

\paragraph{Baselines} The proposed model is compared against three baselines. We first trained ResNet-34~\citep{he2016deep} to predict the presence of malignant and benign findings in a breast. In fact, ResNet-34 is the highest capacity model among the ResNet architectures that can process a mammography in its original resolution while fitting in the memory of an NVIDIA Tesla V100 GPU. We also experimented with a variant of ResNet-34 by replacing the fully connected classification layer with a $1\times1$ convolutional layer and \textit{top $t\%$ pooling} as the aggregation function. In addition, we compared our model with Deep Multi-view CNN (DMV-CNN) proposed by \cite{wu2019deep} which has two versions. In the vanilla version, DMV-CNN applies a ResNet-based model on four standard views to generate two breast-level predictions for each exam. DMV-CNN can also be enhanced with pixel-level heatmaps generated by a patch-level classifier. However, training the patch-level classifier requires hand-annotated segmentation labels. Lastly, we also compared GMIC with the work of \cite{fevry2019improving} which trains a Faster R-CNN \citep{ren2015faster} that utilizes segmentation labels to localize anchor boxes that correspond to malignant or benign lesions. Unlike DMV-CNN and Faster R-CNN which rely on segmentation labels, GMIC can be trained with only image-level labels.

\paragraph{Hyperparameter Tuning} To make a fair comparison between model architectures, we optimize the hyperparameters with random search \citep{bergstra2012random} for both ResNet-34 baselines and GMIC. Specifically, for all models, we search for the learning rate $\eta \in 10^{[-5.5, -4]}$ on a logarithmic scale. Additionally, for GMIC and ResNet-34 with $1\times1$ filters in the last convolutional layer, we also search for the regularization weight $\beta \in 10^{[-5.5, -3.5]}$ (on a logarithmic scale) and for the pooling threshold $t \in \{1\%, 3\%, 5\%, 10\%, 20\%\}$. For all models, we train 30 separate models using hyperparameters randomly sampled from ranges described above. Each model is trained for 50 epochs, and we report the test performance using the weights from the training epoch that achieves highest validation performance.

\begin{table*}
    \centering
      \caption{Comparison of performance of GMIC and the baselines on screening mammogram interpretation. For both GMIC and ResNet-34, we reported test AUC (mean and standard deviation) of \textit{top-5} models that achieved highest validation AUC in identifying breasts with malignant findings. We also measure the total number of learnable parameters in millions, peak GPU memory usage (Mem) for training a single exam (4 images), time taken for forward (Fwd) and backward (Bwd) propagation in milliseconds, and number of floating-point operations (FLOPs) in billions.}
    \begin{tabular}{|l|c|c|c|c|c|c|}
    \hline
    Model & AUC(M) & AUC(B) & $\#$Param & Mem(GB) &  Fwd/Bwd (ms) & FLOPs \\
    \hline
    
    ResNet 34 + fc & $0.736 \pm 0.026$  & $0.684 \pm 0.015$ & 21.30M & 13.95 & 189/459 &  1622B\\
    
    \hline
    ResNet 34 + $1 \times 1$ conv & $0.889 \pm 0.015$ & $0.772 \pm 0.008$ & 21.30M & 12.58  & 201/450 & 1625B\\  \hline
    
     DMV-CNN (w/o heatmaps) & $0.827 \pm 0.008$ & $0.731 \pm 0.004$ & 6.13M & 2.4 & 38/86 & 65B   \\
    \hline
    
    DMV-CNN (w/ heatmaps)  & $0.886 \pm 0.003$   & $0.747 \pm 0.002$  & 6.13M & 2.4 & 38/86 & 65B\\ \hline
    
    Faster R-CNN  & $0.908 \pm 0.014$  & $0.761 \pm 0.008$ & 104.8M & 25.75 & 920/2019 &  -\footnotemark\\ \Xhline{3\arrayrulewidth}

    GMIC-ResNet-18 & $0.913 \pm 0.007$  & $0.791 \pm 0.005$ & 15.17M & 3.01  & 46/82 &  122B\\ \hline
    
    GMIC-ResNet-34  & $0.909 \pm 0.005$  & $0.790 \pm 0.006$ & 25.29M & 3.45  & 58/94 & 180B\\ \hline
    
    GMIC-ResNet-50  & $\mathbf{0.915} \pm 0.005$  & $\mathbf{0.797} \pm 0.003$ & 27.95M & 5.05 & 66/131 &  194B\\  \Xhline{3\arrayrulewidth}
    
    GMIC-ResNet-18-ensemble & $\mathbf{0.930}$ & 0.800 & - & -  & - &  - \\ \hline
    GMIC-ResNet-34-ensemble & 0.920 & 0.795 & - & -  & - & - \\ \hline
    GMIC-ResNet-50-ensemble & 0.927  & $\mathbf{0.805}$ & - & -  & - &  -\\
    \hline
  \end{tabular}
  \label{tb:table_AUC}
\end{table*}

\paragraph{Performance} For each network architecture, we selected the top five models (referred to as \textit{top-5}) from the hyperparameter tuning phase that achieved the highest validation AUC in identifying breasts with malignant findings and evaluated their performance on the held-out test set. In Table~\ref{tb:table_AUC}, we report the mean and the standard deviation of AUC for the \textit{top-5} models in each network architecture. In general, the GMIC model outperformed all baselines. In particular, GMIC achieved higher AUC than Faster R-CNN and DMV-CNN (with heatmaps), despite GMIC not learning with pixel-level labels. We hypothesize that GMIC's superior performance is related to its ability to efficiently integrate both global features and local details. In Section \ref{sec:ablation_study}, we empirically investigate this hypothesis with multiple ablation studies. Separately, we also observe that increasing the complexity of $f_l$ brings a small improvement in AUC. 

To further improve our results, we employed the technique of model ensembling \citep{dietterich2000ensemble}. Specifically, we averaged the predictions of the \textit{top-5} models for GMIC-ResNet-18, GMIC-ResNet-34, and GMIC-ResNet-50 to produce the overall prediction
of the ensemble. Our best ensemble model achieved an AUC of 0.930 in identifying breasts with malignant findings.

In addition, GMIC is efficient in both run-time complexity and memory usage. Compared to ResNet-34, GMIC-ResNet-18 has $28.8\%$ fewer parameters, uses $78.43\%$ less GPU memory, is 4.1x faster during inference and 5.6x faster during training. GMIC achieved even more prominent superiority in both run-time and GPU memory usage compared to Faster R-CNN. This improvement is brought forth by its design that avoids excessive computation on the whole image while selectively focusing on informative regions.

\subsection{Reader Study}
\paragraph{Organization} To evaluate the potential clinical impact of our model, we compare the performance of GMIC to the performance of radiologists using data from the reader study conducted by \cite{wu2019deep}. This study includes 14 readers: 12 attending radiologists at various level of experience (between 2-30 years), a medical resident, and a medical student. Each reader was asked to provide probability estimates as well as binary predictions of malignancy for 720 screening exams (1440 breasts). Among the 1,440 breasts, 62 breasts were associated with malignant findings and 356 breasts were associated with benign findings. Among the breasts in which there were malignant findings, there were 21 masses, 26 calcifications, 12 asymmetries and 4 architectural distortions. The radiologists were only shown images with no other data.

\begin{figure}[t]
    \begin{tabular}{c c}
        \includegraphics[width = 0.22\textwidth,trim={15pt 15pt 25pt 0}]{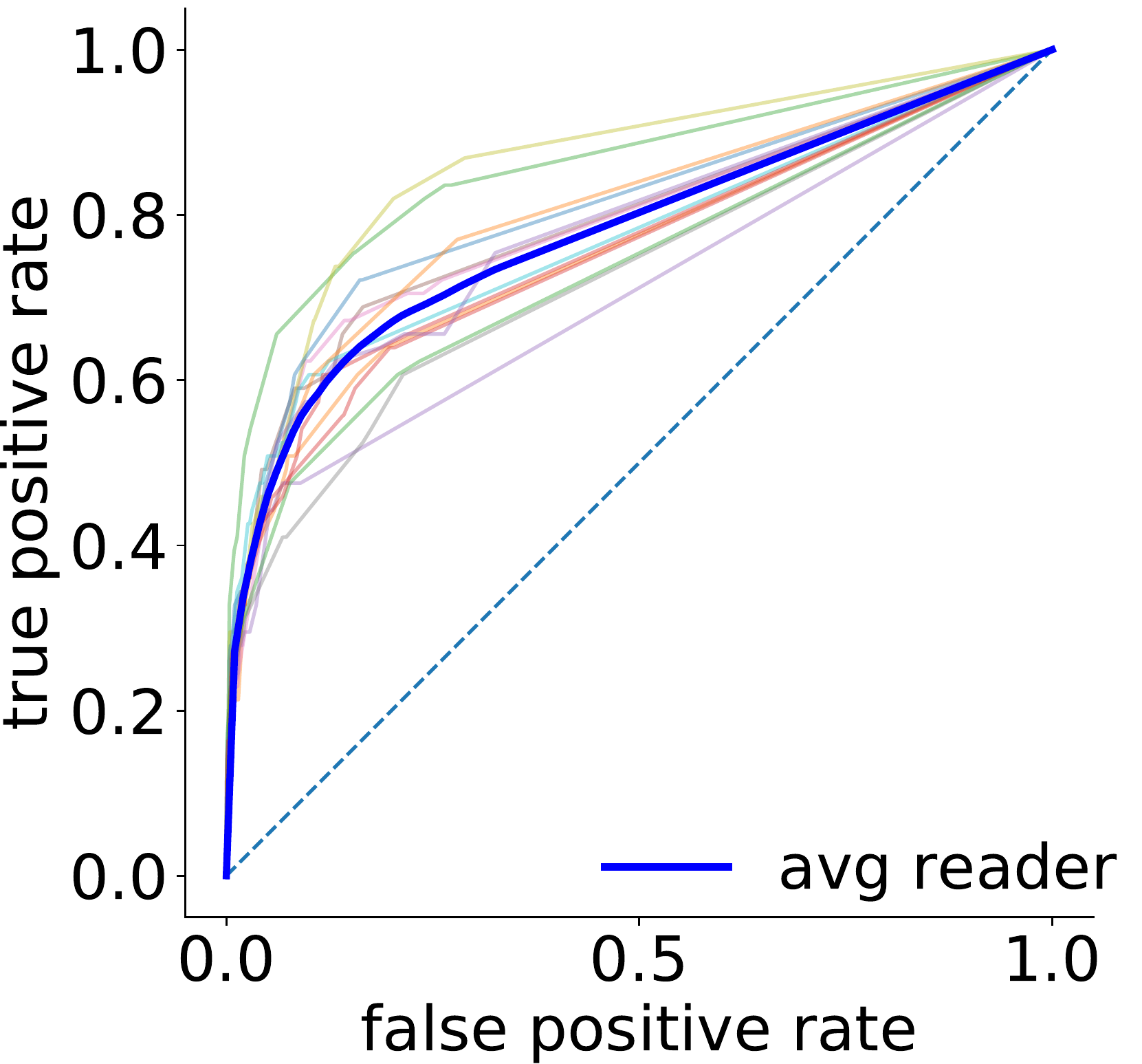} & 
        \includegraphics[width = 0.22\textwidth,trim={15pt 15pt 25pt 0}]{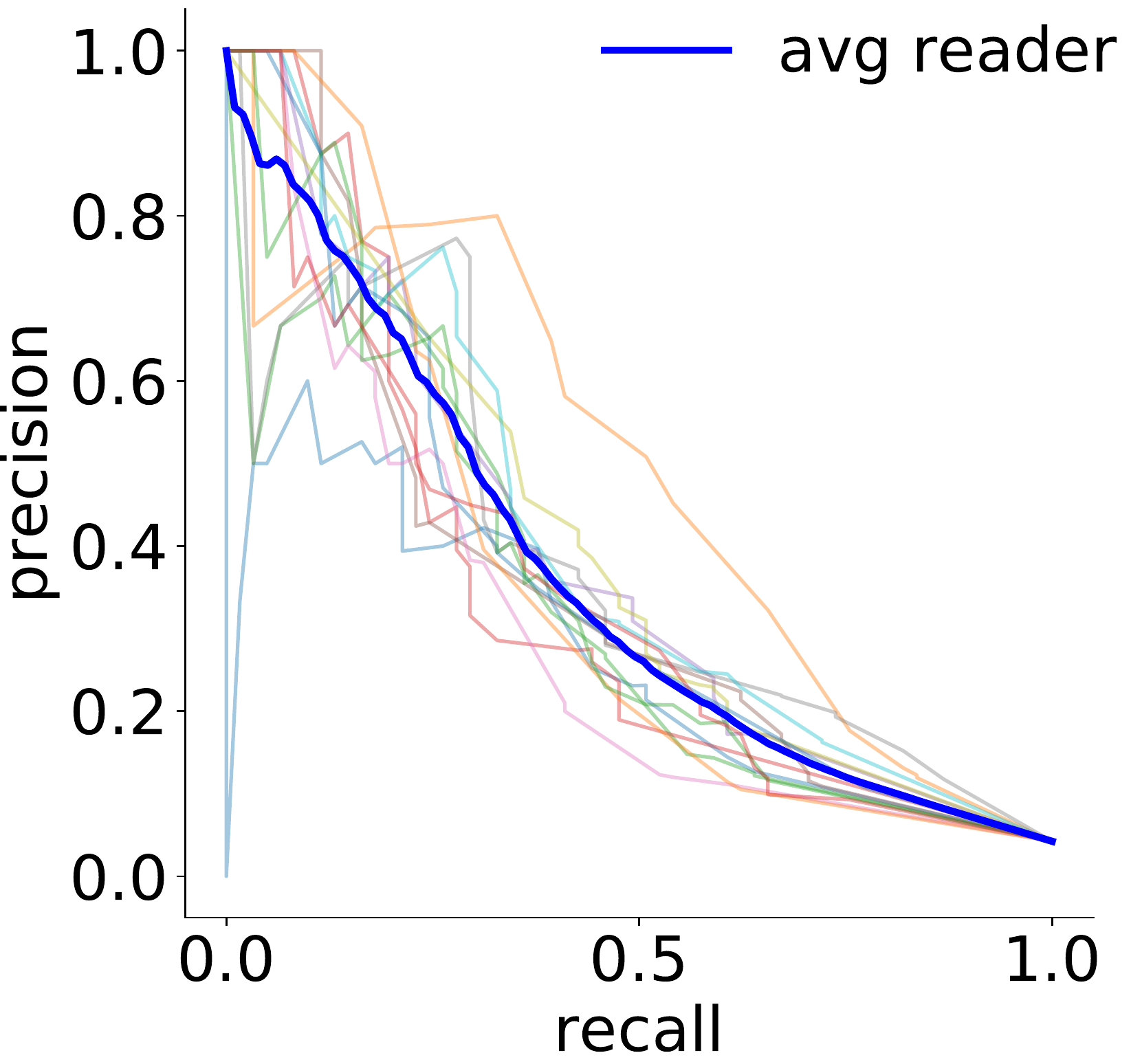}\\
    
    $\quad \quad$ \footnotesize{(a)} & $\quad \quad$  \footnotesize{(a*)}\\
    
    \includegraphics[width = 0.22\textwidth,trim={15pt 15pt 25pt 0}]{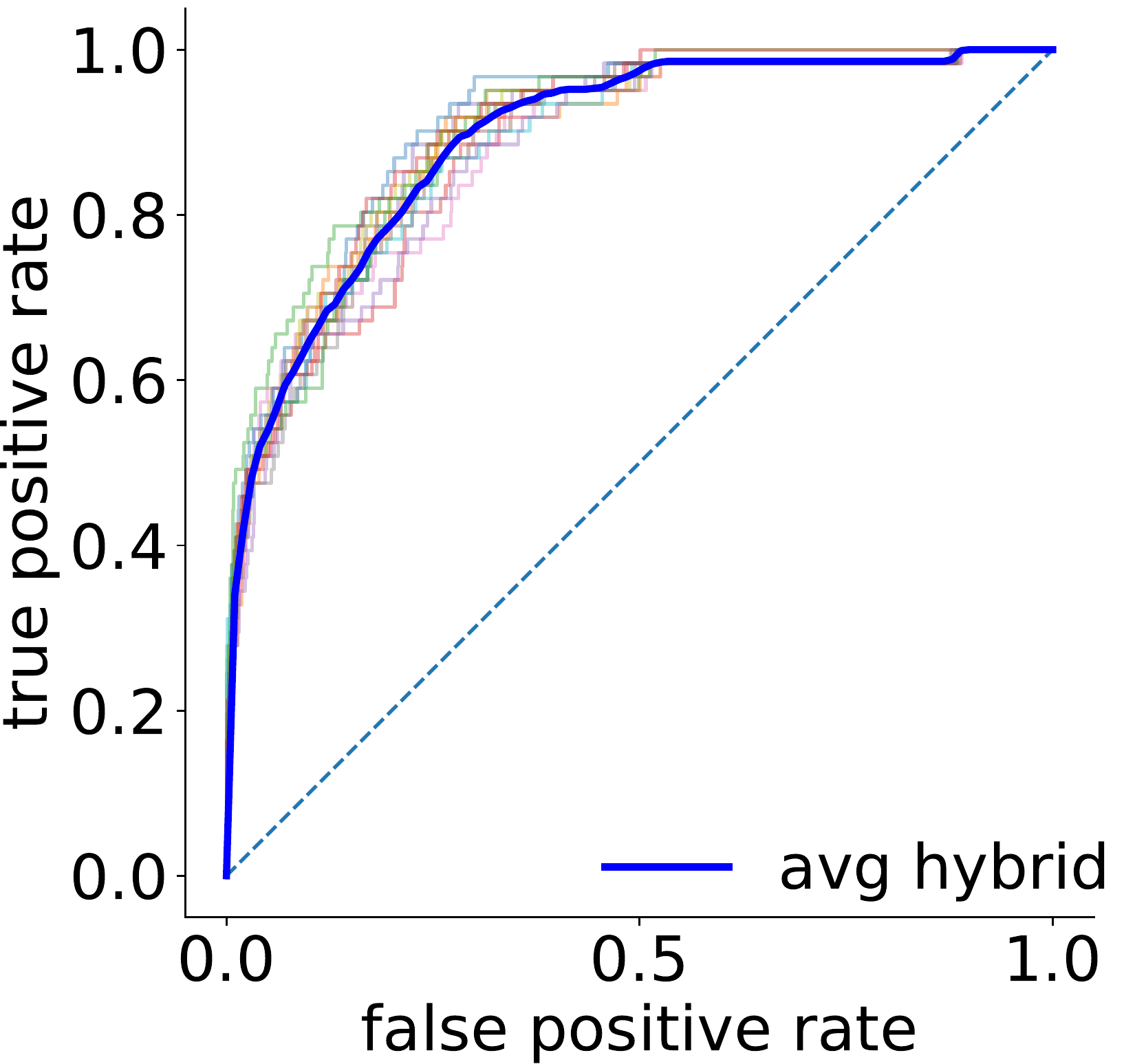} & 
        \includegraphics[width = 0.22\textwidth,trim={15pt 15pt 25pt 0}]{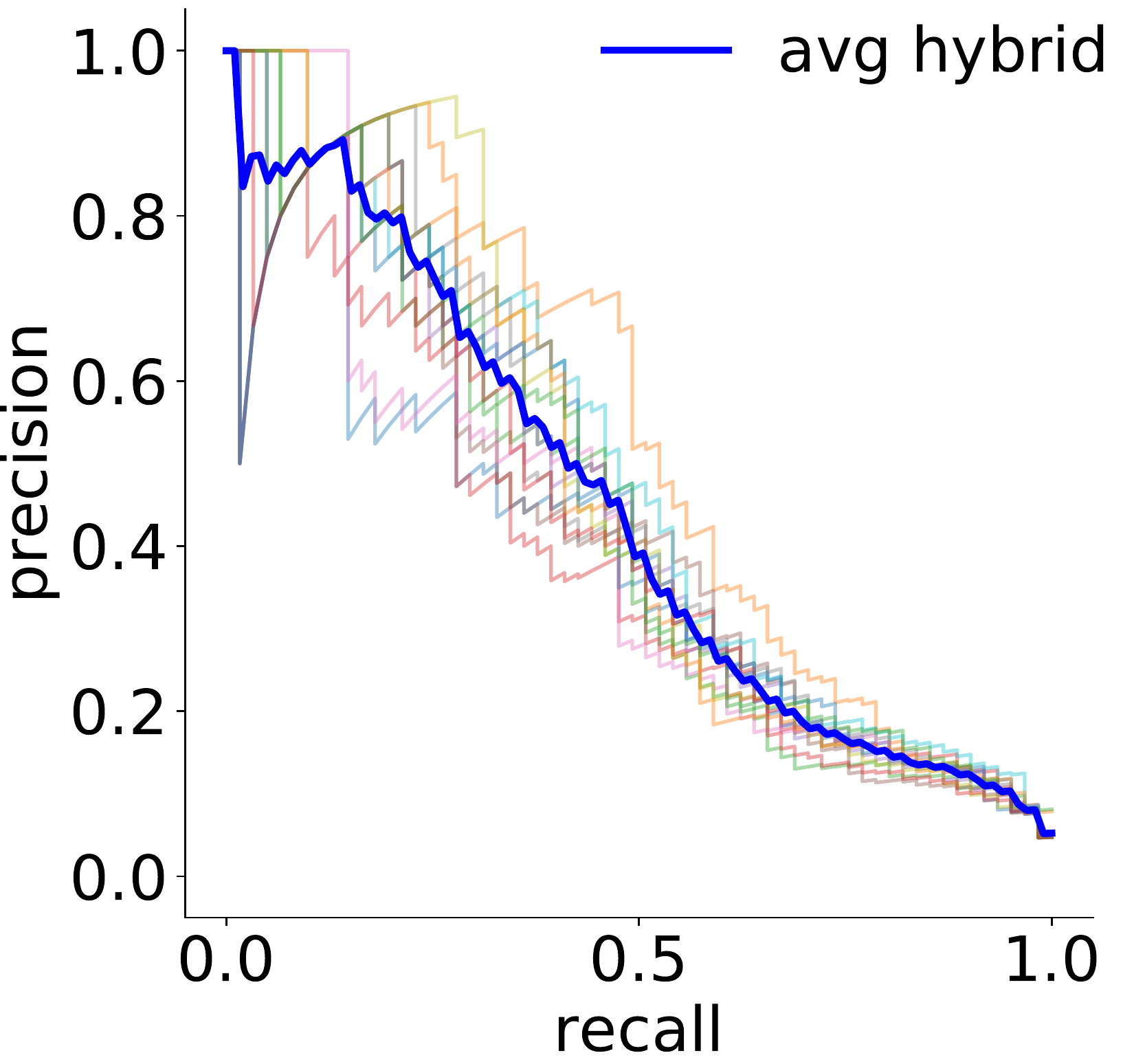}\\
    
    $\quad \quad$ \footnotesize{(b)} & $\quad \quad$ \footnotesize{(b*)}\\
    
    \includegraphics[width = 0.22\textwidth,trim={15pt 15pt 25pt 0}]{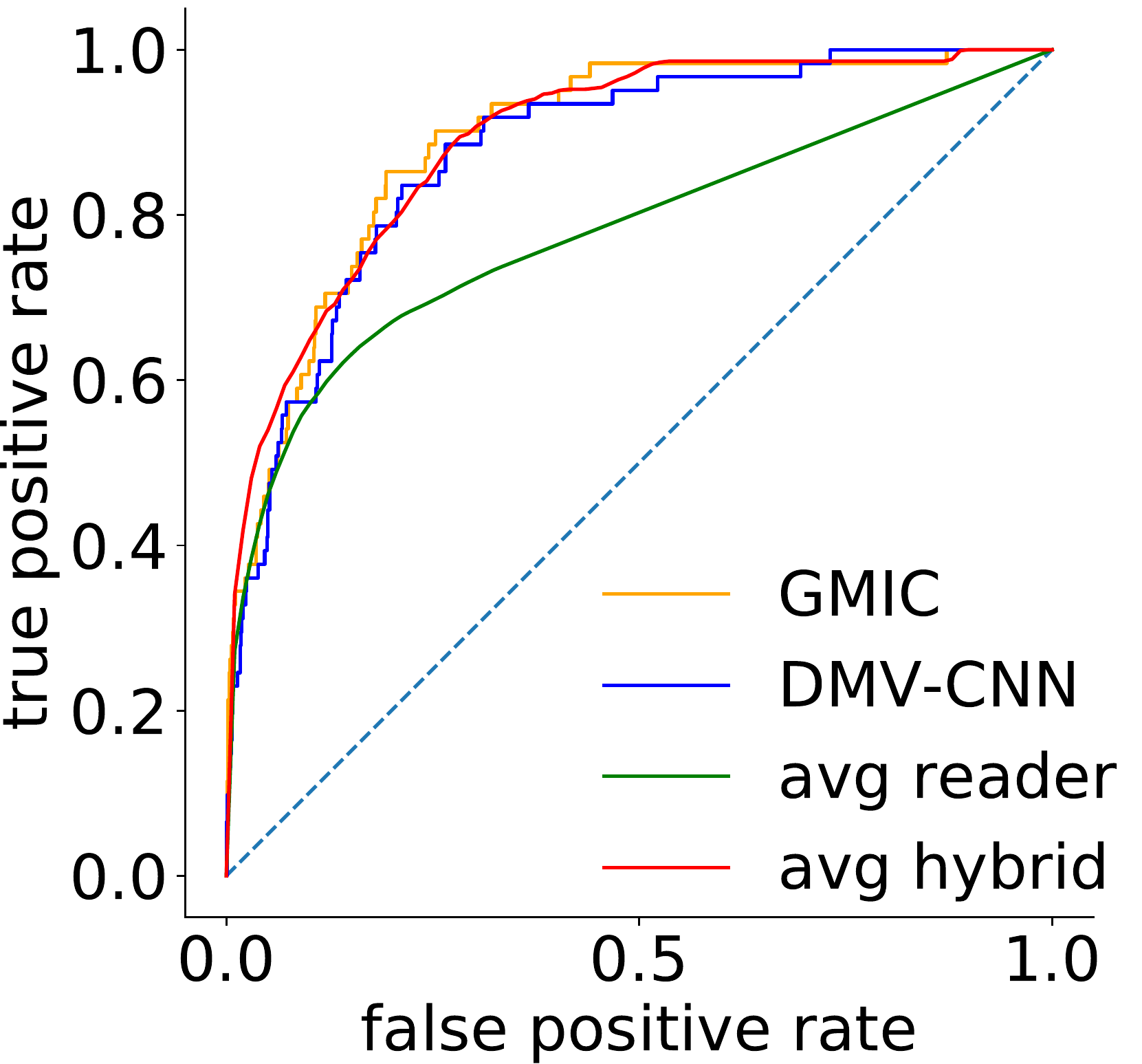} & 
        \includegraphics[width = 0.22\textwidth,trim={15pt 15pt 25pt 0}]{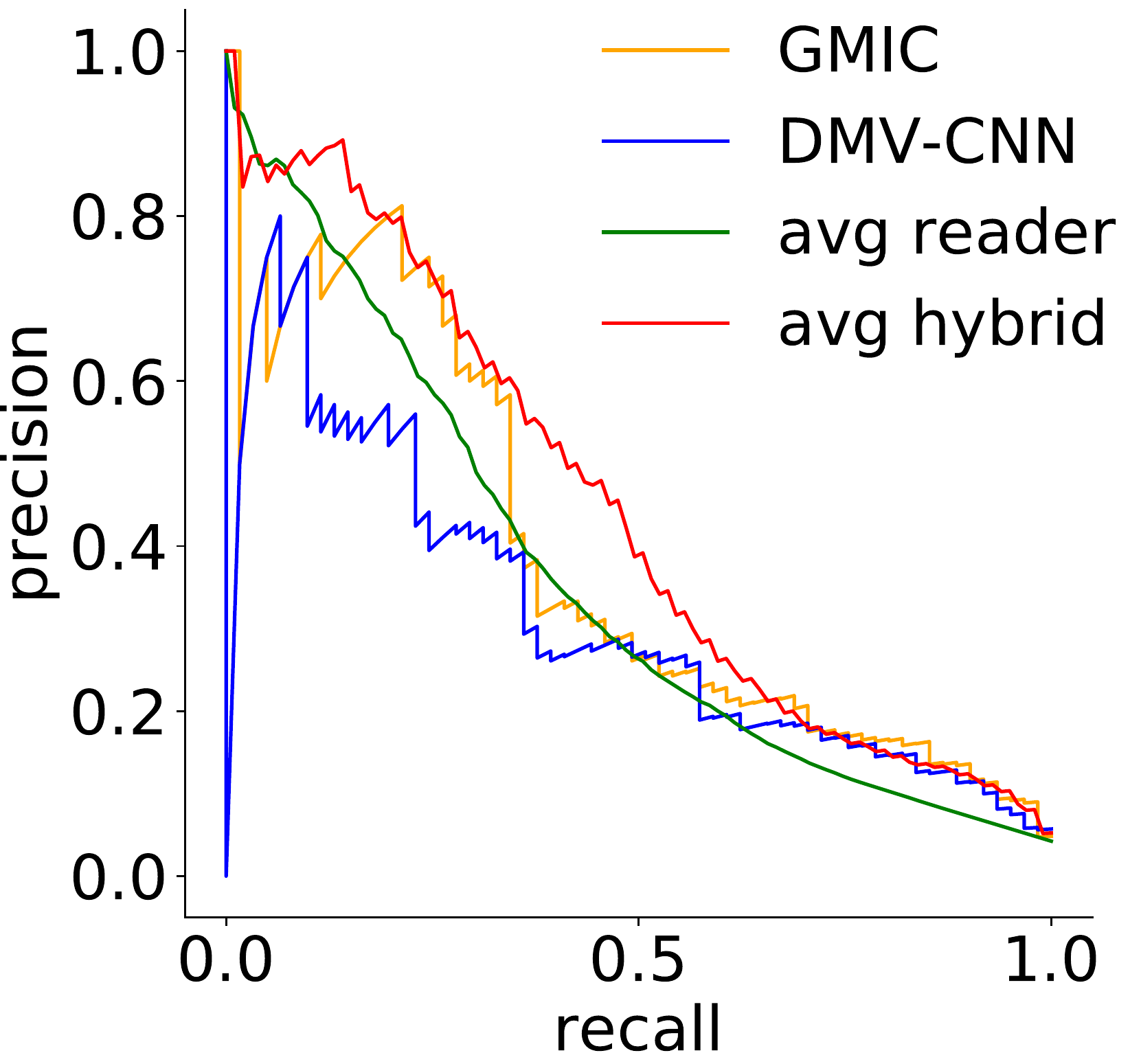}\\
    
    $\quad \quad$ \footnotesize{(c)} & $\quad \quad$ \footnotesize{(c*)}\\
    
    \end{tabular}
    
    \vspace{-2mm}
    \caption{The ROC curves ((a), (b), (c)) and the precision-recall curves ((a*), (b*), (c*)) computed on the reader study dataset. (a) \& (a*): curves for all 14 readers. We derive the ROC/PRC for the average reader by computing the average true positive rate and precision across all readers for every false positive rate and recall. (b) \& (b*): curves for hybrid models with each single reader. The curve highlighted in blue indicates the average performance of all hybrids. (c) \& (c*): comparison among the GMIC, DMV-CNN, the average reader, and average hybrid.}
    \label{fig:readerstudy}
    \vspace{-3mm}
\end{figure}

\paragraph{Comparison to Radiologists} We calculate AUC and PRAUC on the reader study dataset to measure the performance of radiologists and GMIC. We obtain GMIC's predictions by ensembling the predictions of the $\textit{top-5}$ GMIC-ResNet-18 models. In Figure~\ref{fig:readerstudy} ((a) and (a*)), we visualize the receiver operating characteristic curve (ROC) and precision-recall curve (PRC) for each individual reader using their probability estimates of malignancy. We also compared GMIC with DMV-CNN and the radiologists ((c) and (c*)). GMIC achieves an AUC of 0.891 and PRAUC of 0.39 outperforming DMV-CNN (AUC: 0.876, PRAUC: 0.318). The AUCs associated with each individual reader ranges from 0.705 to 0.860 (mean: 0.778, std: 0.0435) and the PRAUCs for readers vary from 0.244 to 0.453 (mean: 0.364, std: 0.0496). GMIC achieves a higher AUC and PRAUC than the average reader. We note that there is a limitation associated with AUC and PRAUC. While AUC and PRAUC are calculated on continuous predictions, radiologists are trained to make diagnosis by choosing from a discrete set of BI-RADS scores~\citep{d2013acr}. Indeed, even though the readers were given a possibility to predict any number between 0\% and 100\%, they chose to stick to the probability threshold corresponding to BI-RADS scores.

To compare GMIC to radiologists, we also use sensitivity and specificity as additional evaluation metrics. We first compute the radiologists' sensitivity and specificity using the data from the reader study. We then use the average specificity and sensitivity among readers as the proxy for radiologists' performance under a single-reader setting and use the statistics of the consensus reading to approximate the performance under a multi-reader setting. The predictions for the consensus reading are derived using majority voting. The 14 radiologists achieved an average specificity of $85.2\%$ (std:$5.5\%$) and average sensitivity of $62.1\%$ (std:$9\%$). The consensus reading yields a specificity of $94.6\%$ and a sensitivity of $76.8\%$. The performance of the radiologists in the reader study is lower than that for community practice radiologists’ performance~\citep{lehman2016national} which reported a sensitivity of $86.9\%$ and a specificity $88.9\%$. However, the overall sensitivity in our study falls within acceptable national performance standards~\citep{lehman2016national} and likely reflects the lack of prior imaging and other clinical data available during interpretation. At the average radiologists' sensitivity level ($62.1\%$), GMIC achieves a specificity of $90\%$ which is higher (P<0.001) than the average radiologists' specificity ($85.2\%$). At the consensus reading sensitivity level ($76.8\%$), GMIC's specificity is $83.6\%$ which is lower than consensus reading specificity ($94.6\%$). While the proposed model underperforms the consensus reading, the results demonstrate the potential value of GMIC as a second reader.

\footnotetext{The implementation of Faster RCNN by~\cite{fevry2019improving} is not compatible with our framework of FLOPs calculation.}

\begin{figure}[t]
    \begin{tabular}{c c}
        \includegraphics[width = 0.22\textwidth,trim={15pt 15pt 25pt 0}]{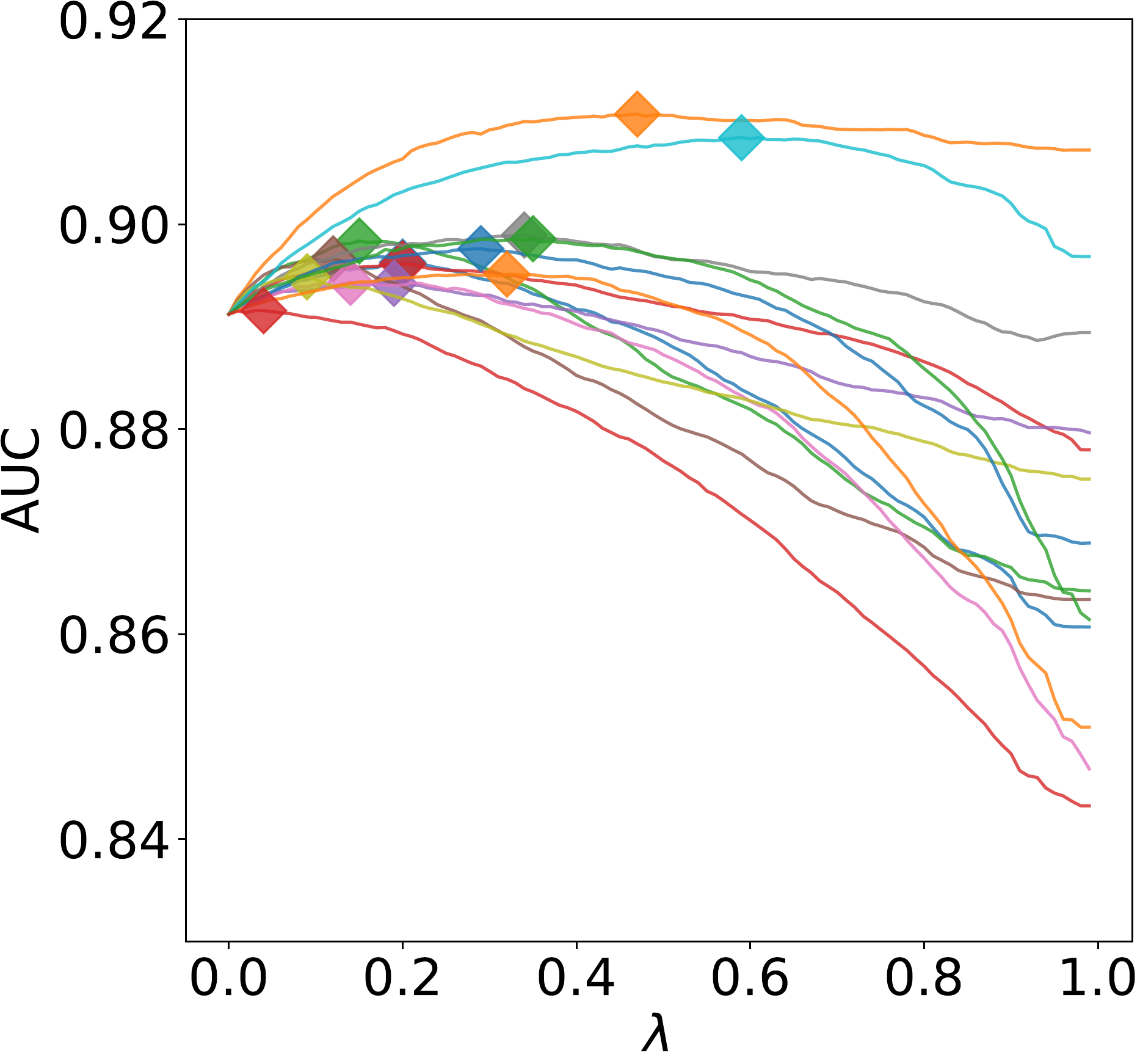} & 
        \includegraphics[width = 0.22\textwidth,trim={15pt 15pt 25pt 0}]{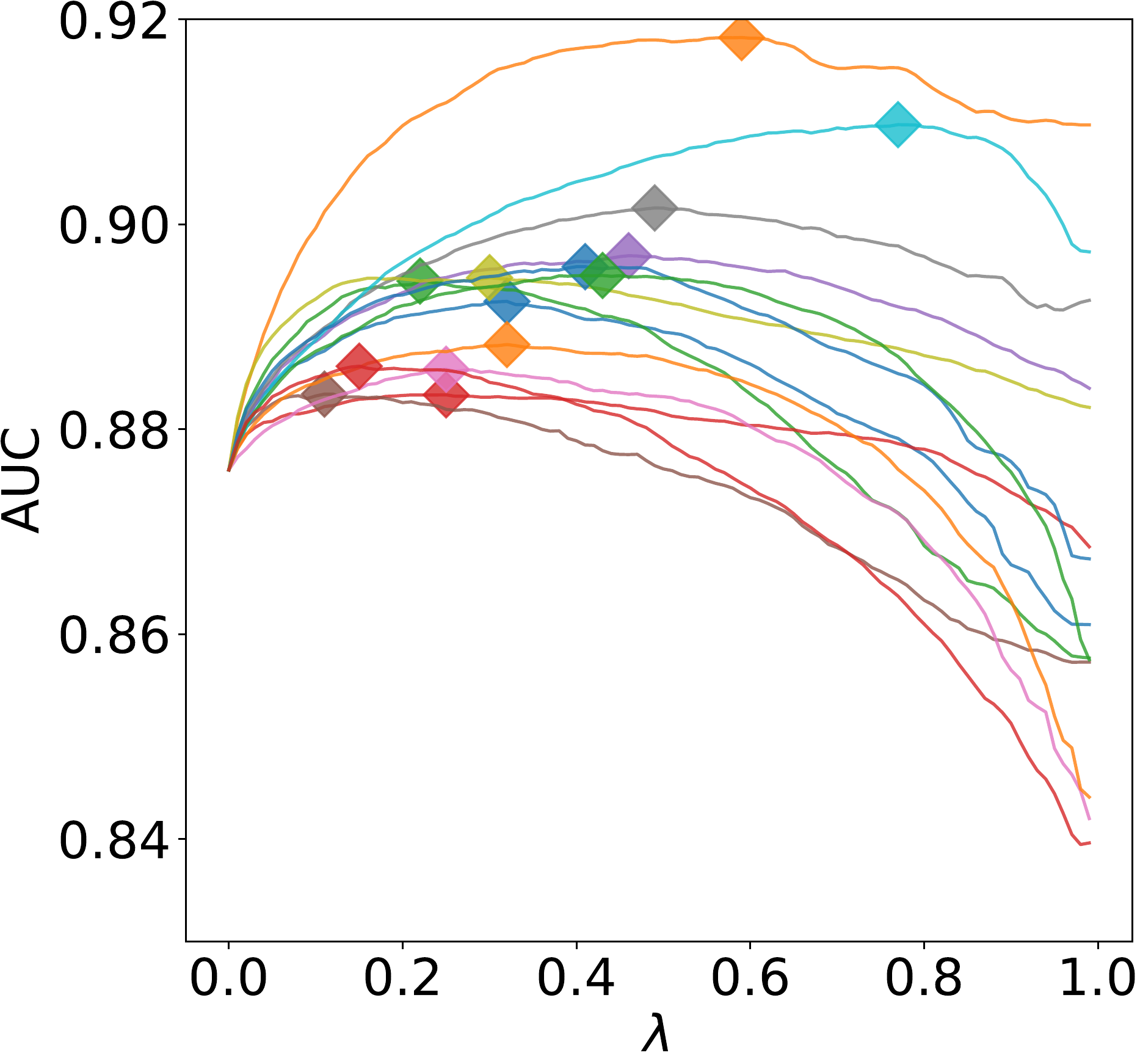}\\
    
    $\quad \quad$ \footnotesize{(a)} & $\quad \quad$  \footnotesize{(a*)}\\
    
    \includegraphics[width = 0.22\textwidth,trim={15pt 15pt 25pt 0}]{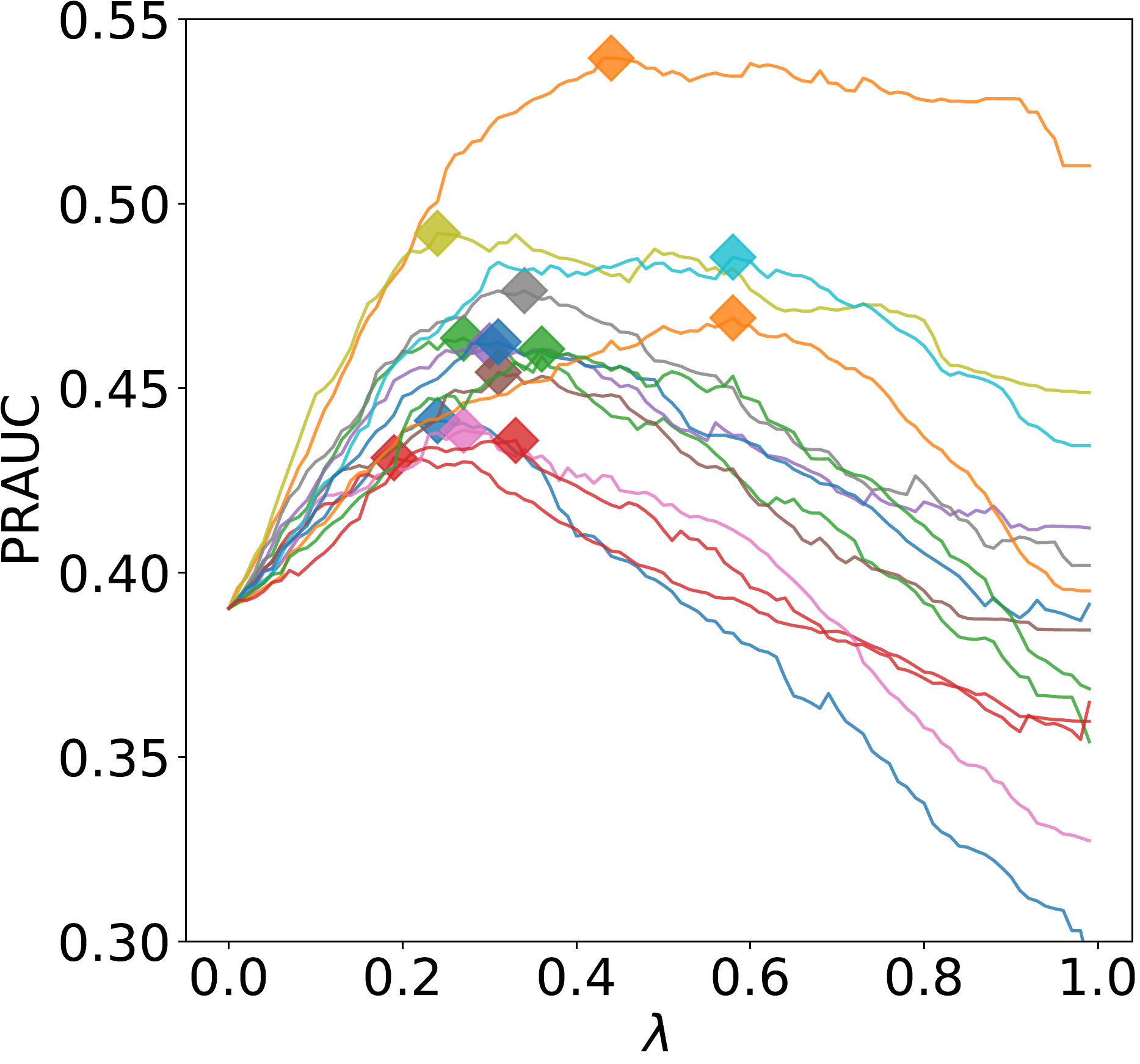} & 
        \includegraphics[width = 0.22\textwidth,trim={15pt 15pt 25pt 0}]{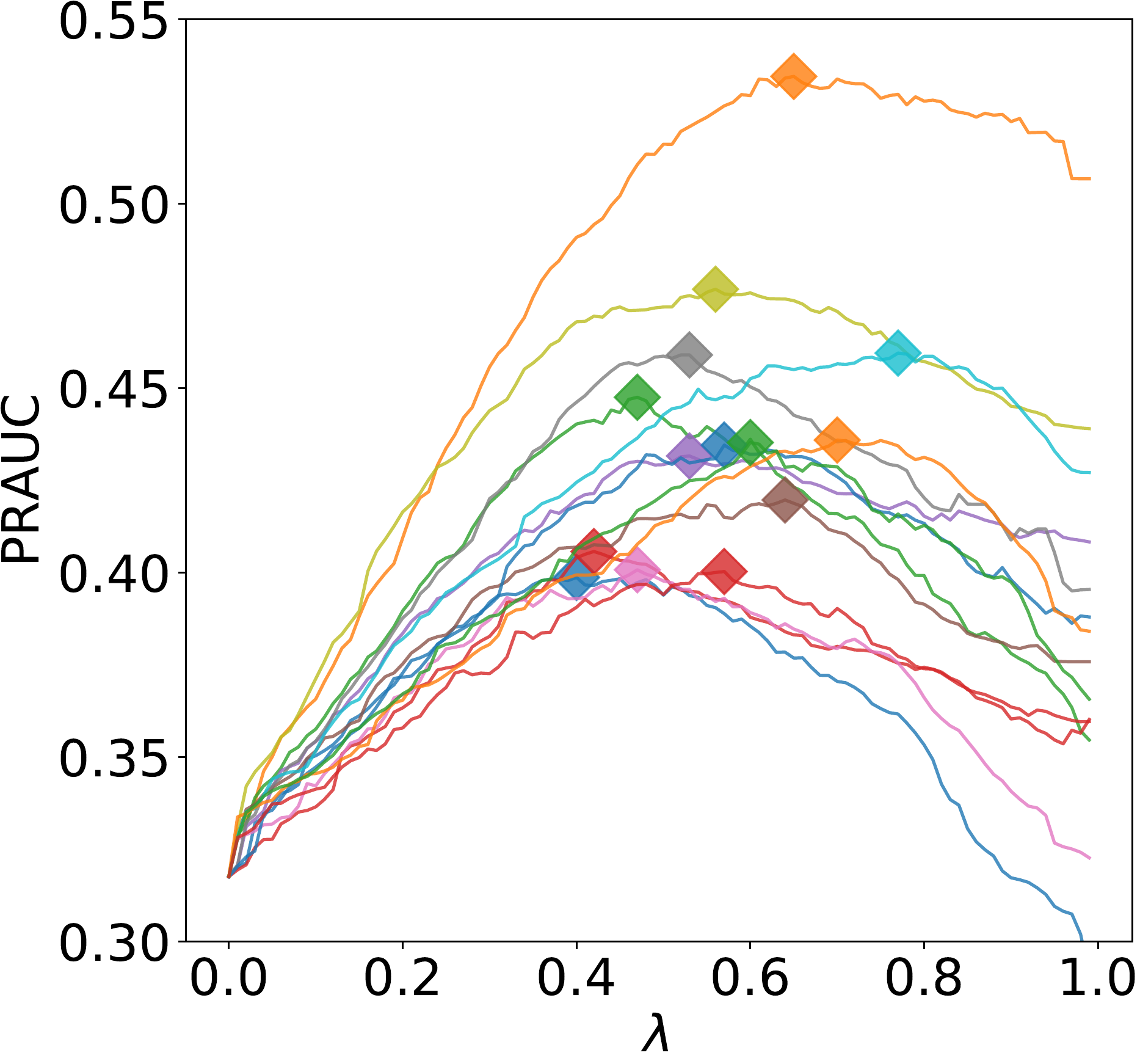}\\
    
    $\quad \quad$ \footnotesize{(b)} & $\quad \quad$  \footnotesize{(b*)}\\
    \end{tabular}
    
    \vspace{-2mm}
    \caption{AUC and PRAUC as a function of $\lambda \in [0, 1)$ for hybrids between each reader and GMIC (left)/DMV-CNN (right) ensemble. Each hybrid achieves the highest AUC/PRAUC for a different $\lambda$ (marked with $\diamondsuit$).
    }
    \label{fig:hybrid}
    \vspace{-3mm}
\end{figure}

\begin{figure}[t]
    \begin{tabular}{c c}
    \includegraphics[width = 0.22\textwidth,trim={15pt 15pt 25pt 0}]{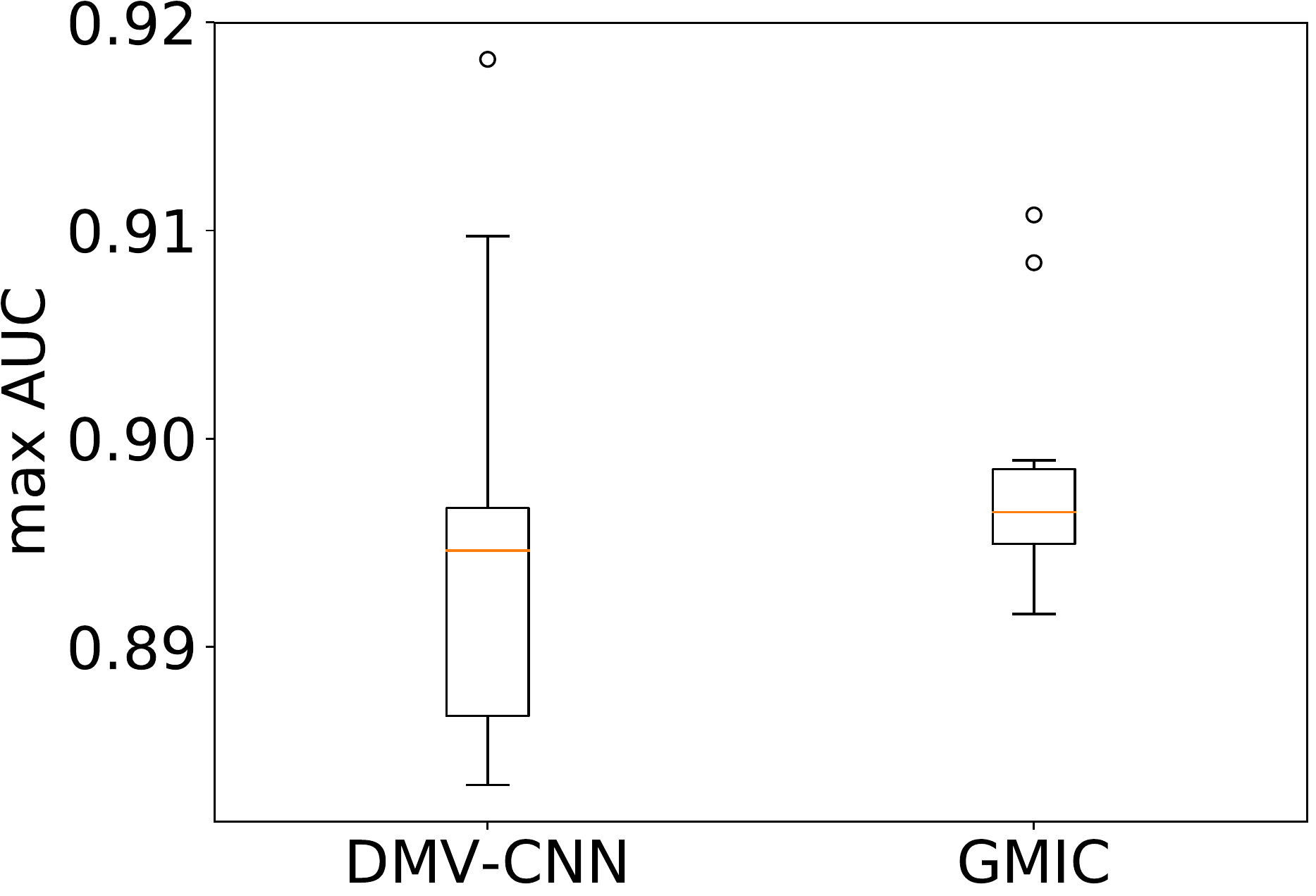} & 
        \includegraphics[width = 0.22\textwidth,trim={15pt 15pt 25pt 0}]{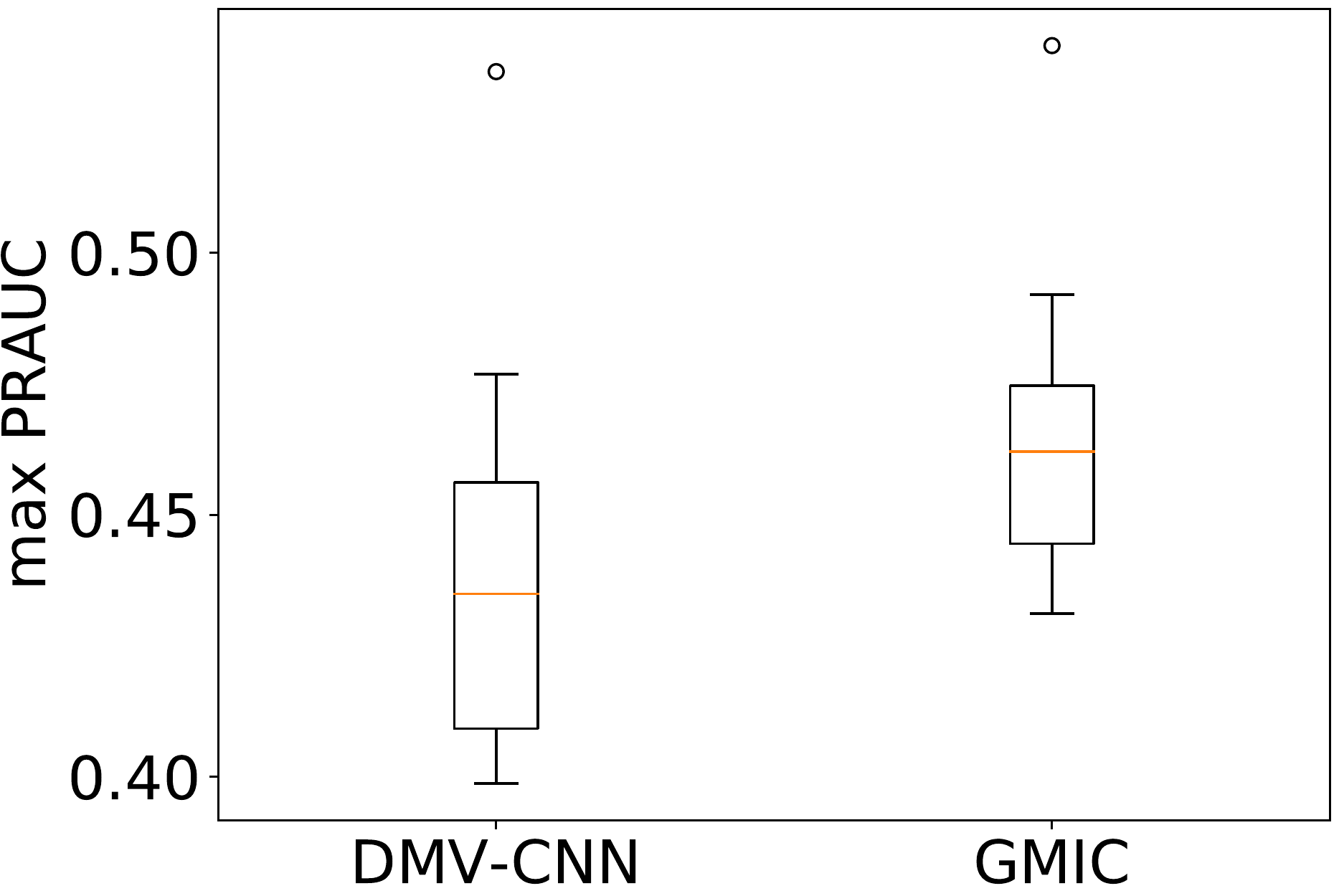}\\
    
    $\quad \quad$ \footnotesize{(a)} & $\quad \quad$  \footnotesize{(a*)}\\
    
    \includegraphics[width = 0.22\textwidth,trim={15pt 15pt 25pt 0}]{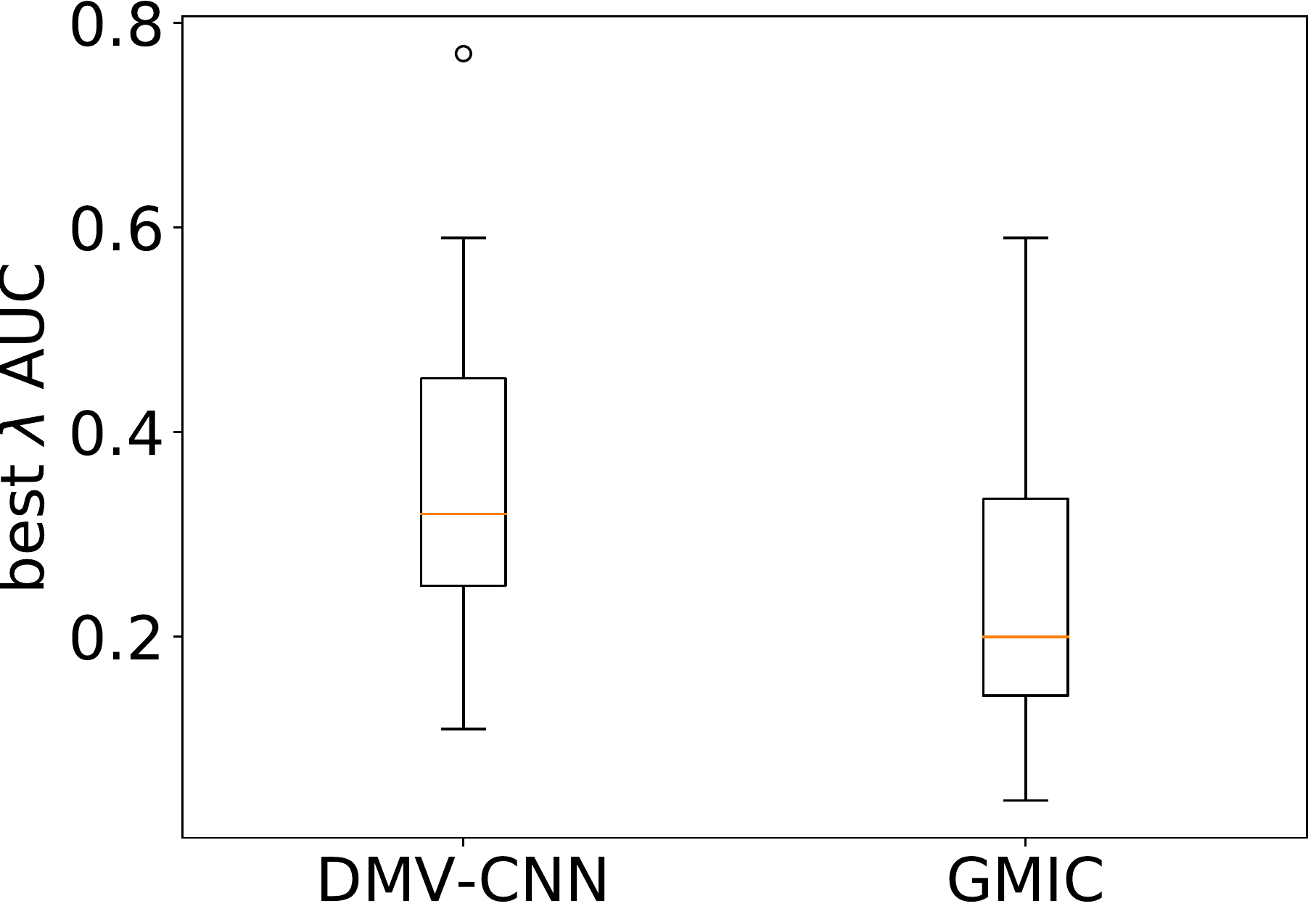} & 
        \includegraphics[width = 0.22\textwidth,trim={15pt 15pt 25pt 0}]{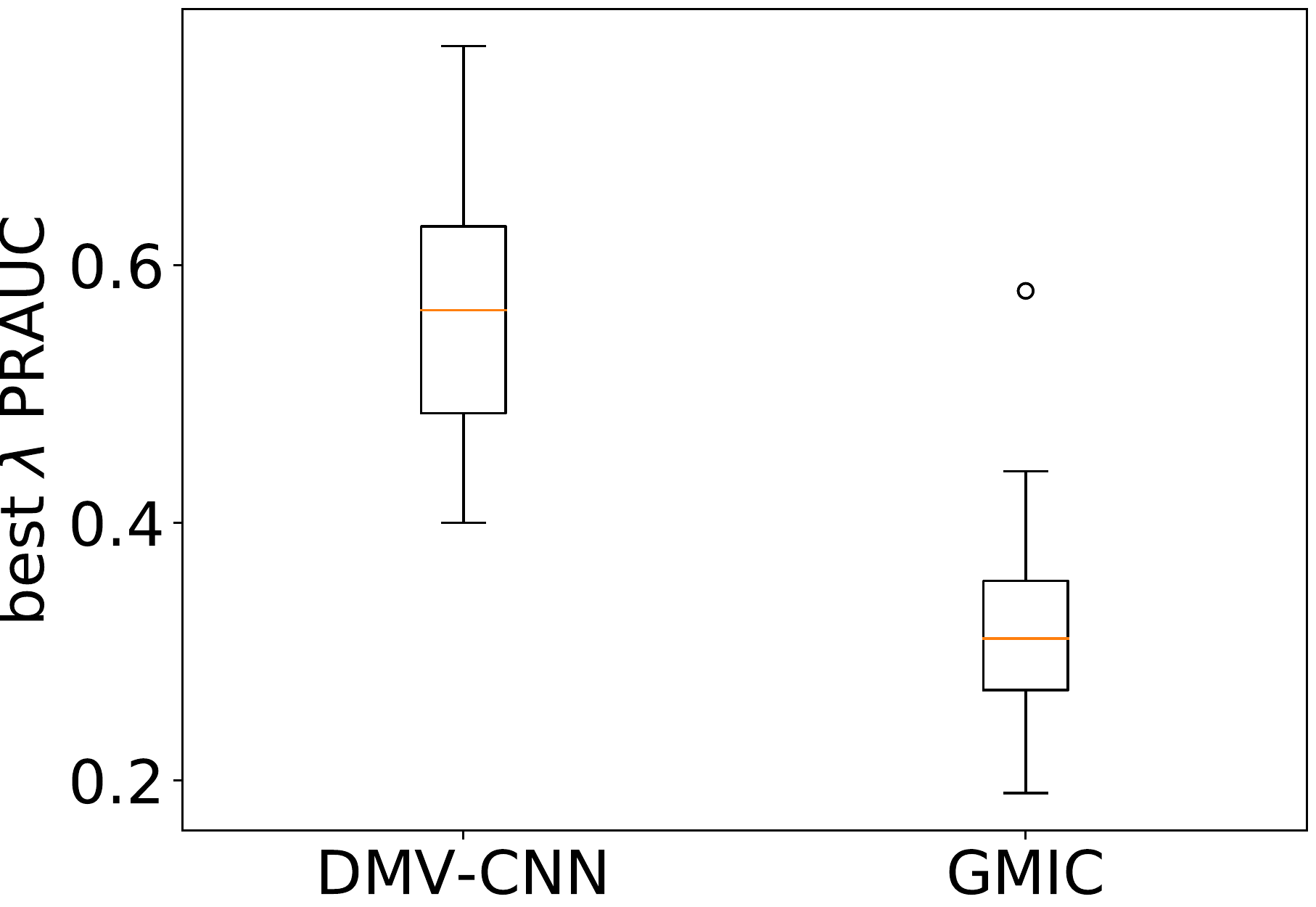}\\
    
    $\quad \quad$ \footnotesize{(b)} & $\quad \quad$  \footnotesize{(b*)}\\
    
    \end{tabular}
    
    \vspace{-2mm}
    \caption{(a) and (a*): the distribution of maximum AUC/PRAUC achieved for hybrids between each reader and GMIC/DMV-CNN ensemble. (b) and (b*): the distribution of the optimal $\lambda^{*}$ that achieves the maximum AUC/PRAUC for both GMIC/DMV-CNN hybrids. GMIC hybrids achieve higher AUC and PRAUC than DMV-CNN hybrids. Moreover, GMIC plays a more important role than DMV-CNN in the hybrid models as indicated by the distribution of $\lambda^*$.}
    \label{fig:hybrid_lambda}
    \vspace{-3mm}
\end{figure}

\paragraph{Human-machine Hybrid} To further demonstrate the clinical potential of GMIC, we create a hybrid model whose predictions are a linear combination of predictions from each reader and the model: $\hat{\mathbf{y}}_\text{hybrid} =  \lambda \hat{\mathbf{y}}_\text{reader} + (1 - \lambda) \hat{\mathbf{y}}_\text{GMIC}$. We compute the AUC and PRAUC of the hybrid models by setting $\lambda = 0.5$. We note that $\lambda=0.5$ is not the optimal value for all hybrid models. On the other hand, the performance obtained by retroactively fine-tuning $\lambda$ on the reader study is not transferable to realistic clinical settings. Therefore, we chose $\lambda = 0.5$ as the most natural way of aggregating two sets of predictions when not having prior knowledge of their quality. In Figure \ref{fig:readerstudy} ((b) and (b*)), we visualize the ROC and PRC curves of the hybrid models ($\lambda = 0.5$) which on average achieve an AUC of 0.892 (std: 0.009) and an PRAUC of 0.449 (std: 0.036), improving radiologists' mean AUC by 0.114 and mean PRAUC by 0.085. For each of the hybrid models, we also calculate its specificity at the average radiologists' sensitivity ($62.1\%$). The 14 hybrid models achieve an average specificity of $91.5\%$ (std:$1.8\%$) which is higher than (P < 0.001) the average radiologists' specificity ($85.2\%$). These results indicate that our model captures different aspects of the task compared to radiologists and can be used as a tool to assist in interpreting breast cancer screening exams.

In addition, in Figure \ref{fig:hybrid}, we visualize the AUC and PRAUC achieved by combining predictions from each of these 14 readers with GMIC ((a) and (b)) and DMV-CNN ((a*) and (b*)) with varying $\lambda$. The diamond mark on each curve indicates the $\lambda^*$ that achieves the highest AUC/PRAUC. As shown in the plot, the predictions from all radiologists could be improved ($\lambda^* < 1.0$) by incorporating predictions from GMIC. More specifically, as shown in Figure~\ref{fig:hybrid_lambda} ((a) and (a*)), with the optimal $\lambda^{*}$, GMIC hybrids achieves a mean AUC of $0.898\pm0.005$ and mean PRAUC of $0.465\pm0.03$ both of which are higher than the counterparts of DMV-CNN hybrids (AUC:$0.895\pm0.01$, PRAUC:$0.439\pm0.035$). In addition, we compare the distribution of $\lambda^*$ for GMIC and DMV-CNN. The average value of $\lambda^*$ associated with GMIC hybrid models to achieve maximum AUC/PRAUC is $0.25\pm0.15/0.34\pm0.11$ which is lower than DMV-CNN ($0.34\pm0.15/0.59\pm0.12$). This result shows that, the more accurate the model used in the human-machine hybrid is, the more weight is attached to its predictions.
    
\begin{figure*}
    \centering
\includegraphics[width=0.95\textwidth]{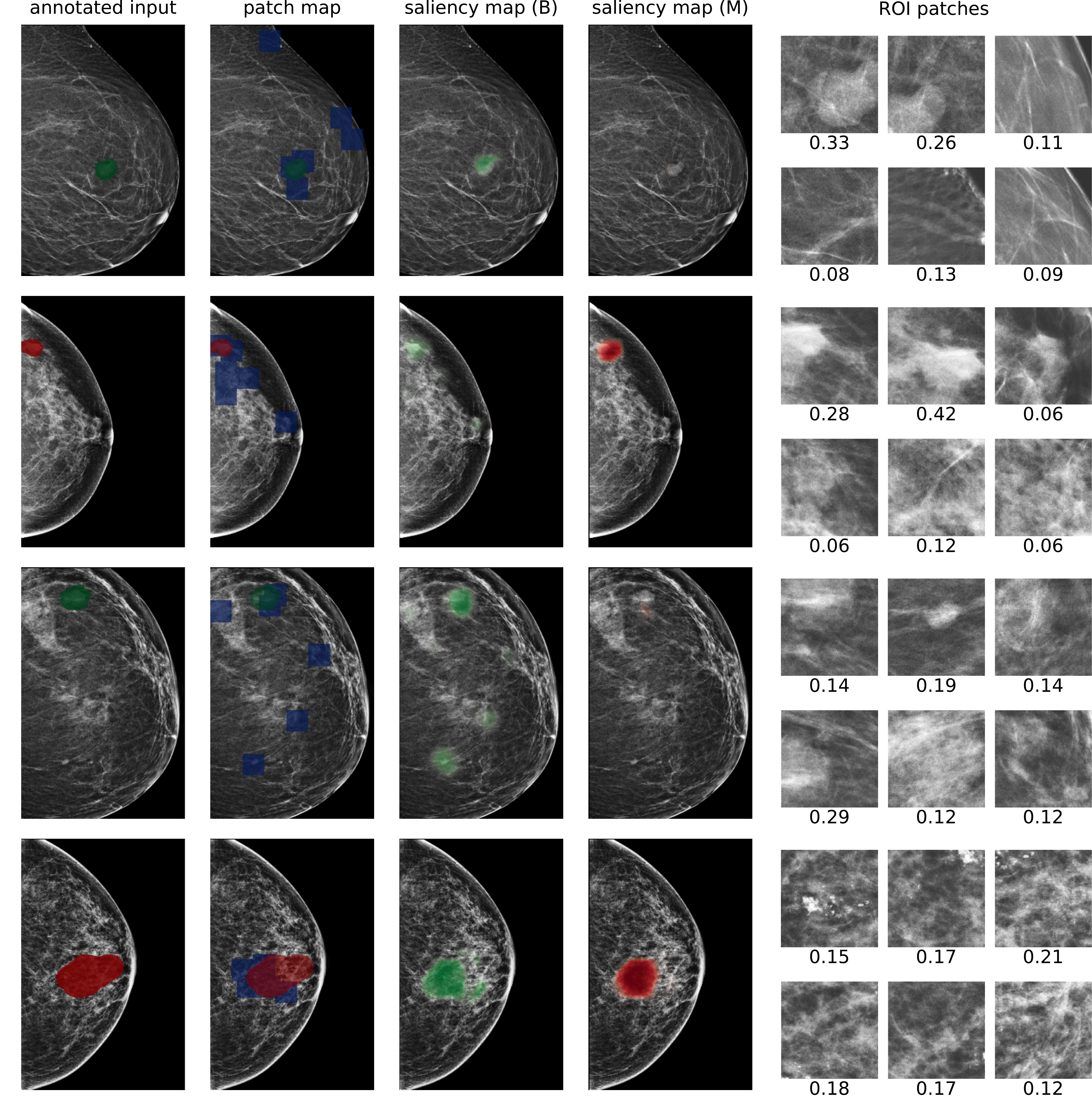}
    \caption{Visualization of results for four examples. From left to right: input images annotated with segmentation labels (green=benign, red=malignant), locations of ROI patches (blue squares), saliency map for benign class, saliency map for malignant class, and ROI patches with their attention scores. The top example contains a circumscribed oval mass in the left upper breast middle depth which was diagnosed as a benign fibroadenoma by ultrasound biopsy. The second example contains an irregular mass in the right lateral breast posterior depth which was diagnosed as an invasive ductal carinoma by ultrasound biopsy. In the third example, the benign saliency map identifies (from up to bottom) (a) a circumscribed oval mass in the lateral breast middle depth, (b) a smaller circumscribed oval mass in the media breast, and (c) an asymmetry in the left central breast middle depth. Ultrasound-guided biopsy of the finding shown in (a) yielded benign fibroadenoma. The medial breast mass (b) was recommended for short-term follow-up by the breast radiologist. The central breast asymmetry (c) was imaging-proven stable on multiple prior mammograms and benign. The bottom example contains segmental coarse heterogeneous calcifications in the right central breast middle depth. Stereotactic biopsy yielded high grade ductal carcinoma in situ. We provide additional visualizations of exams with benign and malignant findings in the Appendix (Figure~\ref{fig:add_vis_ben} and Figure~\ref{fig:add_vis_mal}).}
  \label{vis_plot}
\end{figure*}

\subsection{Localization Performance} \label{sec:localization_performance}
To evaluate the localization performance of GMIC, we select the model with the highest DSC for malignancy localization using the validation set. During inference, we upsample saliency maps using nearest neighbour interpolation to match the resolution of the input image. Our best localization model achieves a mean test DSC of 0.325 (std:0.231) for localization of malignant lesions and 0.240 (std:0.175) for localization of benign lesions. The best localization model achieves an AUC of 0.886/0.78 on classifying malignant/benign lesions. We observe that localization and classification performance are not perfectly correlated. The trade-off between classification and localization has been discussed in the weakly supervised object detection literature~\citep{feng2017discriminative,sedai2018deep,yao2018weakly}. 

In Figure~\ref{vis_plot}, we visualize saliency maps for four samples selected from the test set. In the first two examples, the saliency maps are highly activated on the annotated lesions, suggesting that our model is able to detect suspicious lesions without pixel-level supervision. Moreover, the attention $\alpha_k$ is highly concentrated on ROI patches that overlap with the annotated lesions. In the third example, the saliency map for benign findings identifies three abnormalities. Although only the top abnormality was escalated for biopsy and hence annotated by radiologists, the radiologist's report confirms that the two non-biopsied findings have a high probability of benignity and a low probability of malignancy. In the fourth example, we illustrate a case when there is some level of disagreement between our model and the annotation in the dataset. The malignancy saliency map only highlights part of a large malignant lesion with segmental coarse heterogeneous calcifications. This behavior is related to the design of $f_{\text{agg}}$: a fixed pooling threshold $t$ cannot be optimal for all sizes of ROI. The impact of $f_\text{agg}$ is further studied in \ref{sec:ablation_study}. This example also illustrates that while human experts are asked to annotate the entire lesion, CNNs tend to emphasize only the most informative regions. While no benign lesion is present, the benign saliency map still highlights regions similar to that in the malignancy saliency map, but with a lower probability than the malignancy saliency map. In fact, calcifications with this morphology and distribution can also result from benign pathophysiology~\citep{liberman2002breast}.

\begin{figure*}[h!]
    \centering
\includegraphics[width=0.18\textwidth]{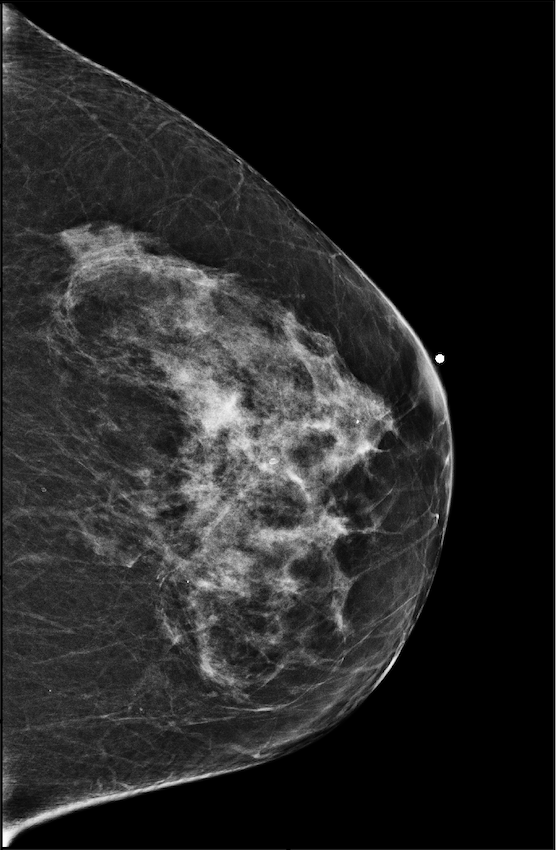} \hspace{3pt}
\includegraphics[width=0.18\textwidth]{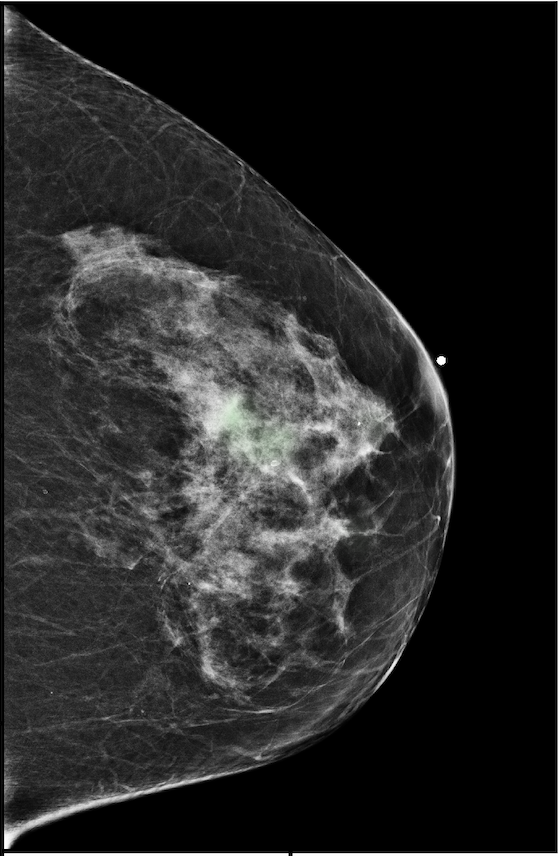} \hspace{3pt}
\includegraphics[width=0.18\textwidth]{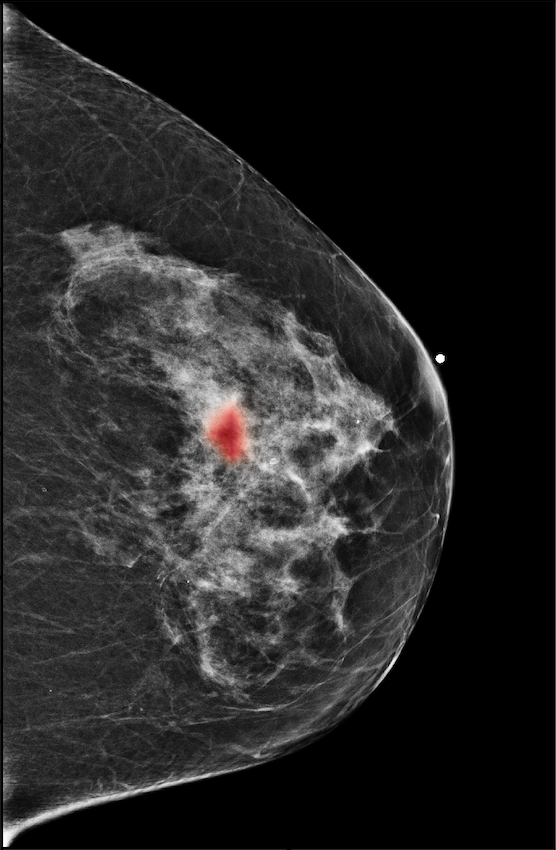} \hspace{3pt}
\includegraphics[width=0.309\textwidth]{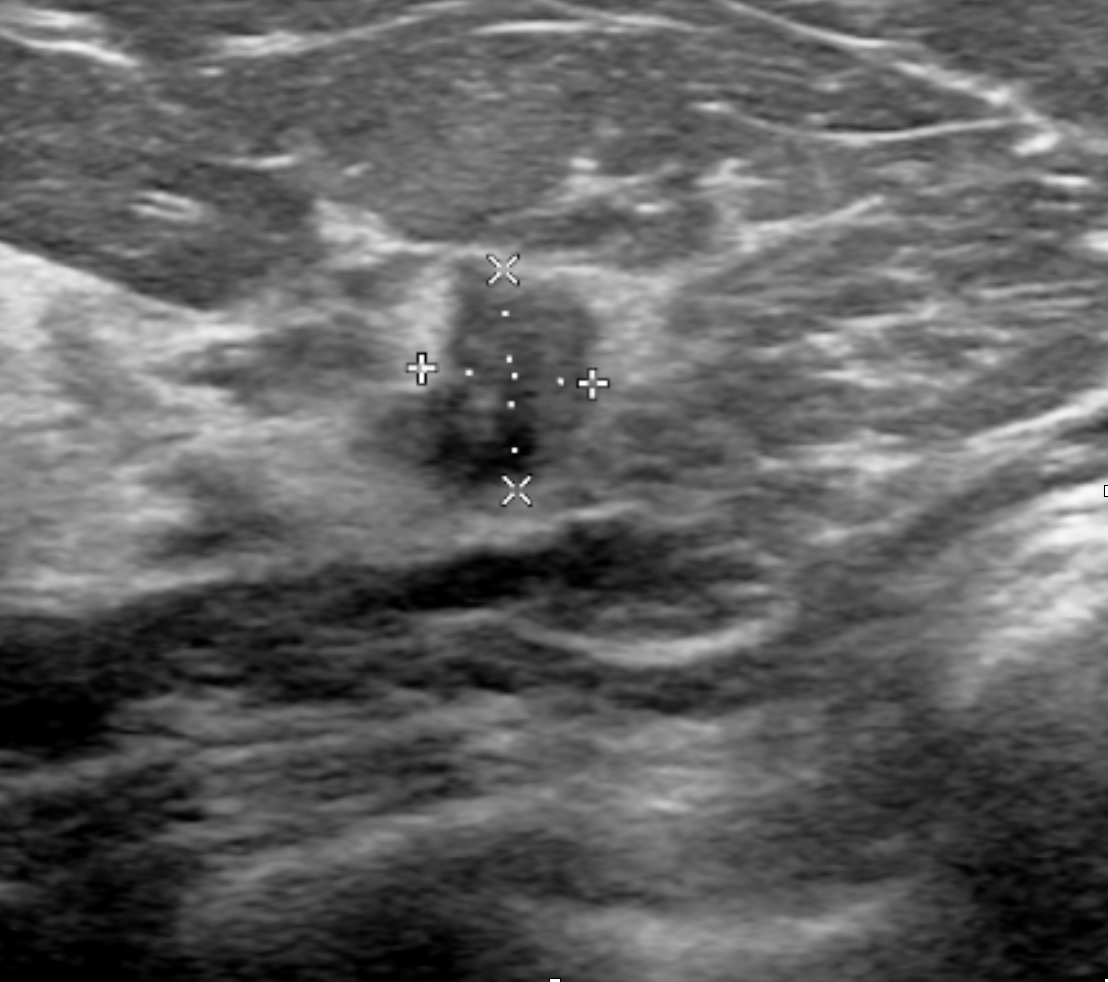}
 \caption{A mammographically occult example with a biopsy-proven malignant finding. From left to right: the original image, the saliency map for benign findings, the saliency map for malignant findings, and the sagittal ultrasound image of this patient. While the asymmetry in the left lateral breast posterior depth was intepreted as benign by the radiologist, a subsequent screening ultrasound and ultrasound-guided biopsy yielded mammographically-occult moderate grade invasive ductal carcinoma. On saliency maps, this area shows a weak probability of benignity and a high probability of malignancy.}
  \label{fig:occult}
\end{figure*}

In addition, we observe that GMIC is able to provide meaningful localization when the lesions are hardly visible to radiologists in the image. In Figure~\ref{fig:occult}, we illustrate a mammographically occult mammogram of a 59-year old patient with no family history of breast cancer and dense breasts. There is an asymmetry in the left lateral breast posterior depth which appears stable compared to prior mammograms and was determined to be benign by the reading radiologist. However, the saliency map of malignant findings successfully identifies the malignant lesion on the screening mammogram. Same day screening ultrasound (sagittal image) demonstrated a 1.2 cm irregular mass; ultrasound biopsy yielded moderate grade invasive ductal carcinoma. 

\begin{figure*}
    \centering
    \setlength{\tabcolsep}{3pt}
    \begin{tabular}{c c c c c c}
    
    Ground Truth & GMP & top $3\%$ & top $10\%$ & top $20\%$ & GAP \\
  
  \includegraphics[width=0.15\textwidth,height=120pt]{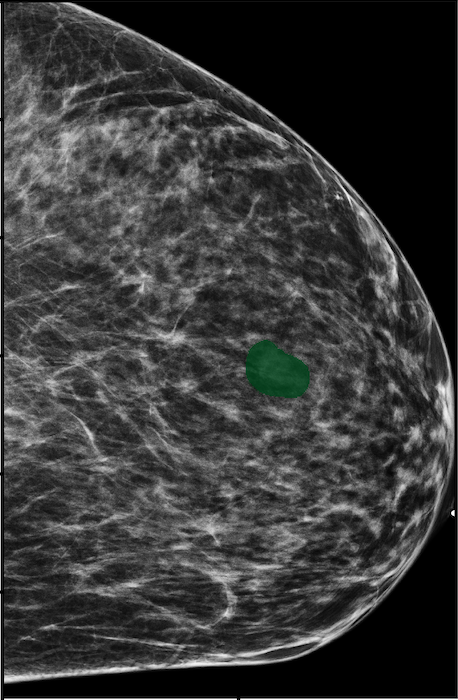} &
  \includegraphics[width=0.15\textwidth,height=120pt]{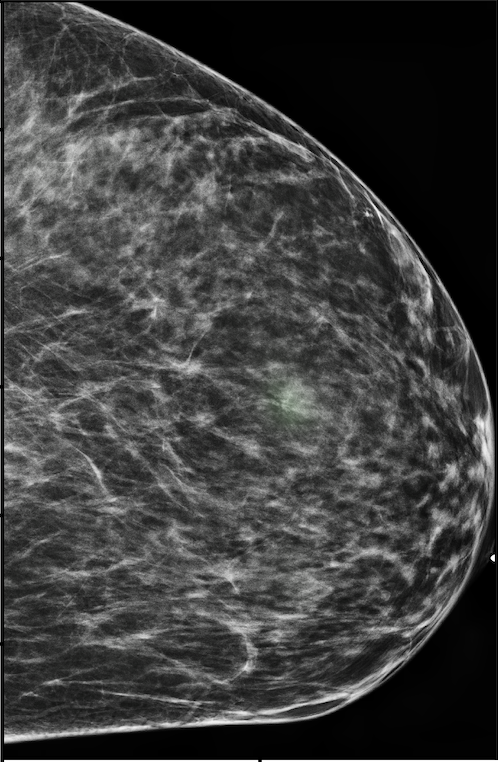} & \includegraphics[width=0.15\textwidth,height=120pt]{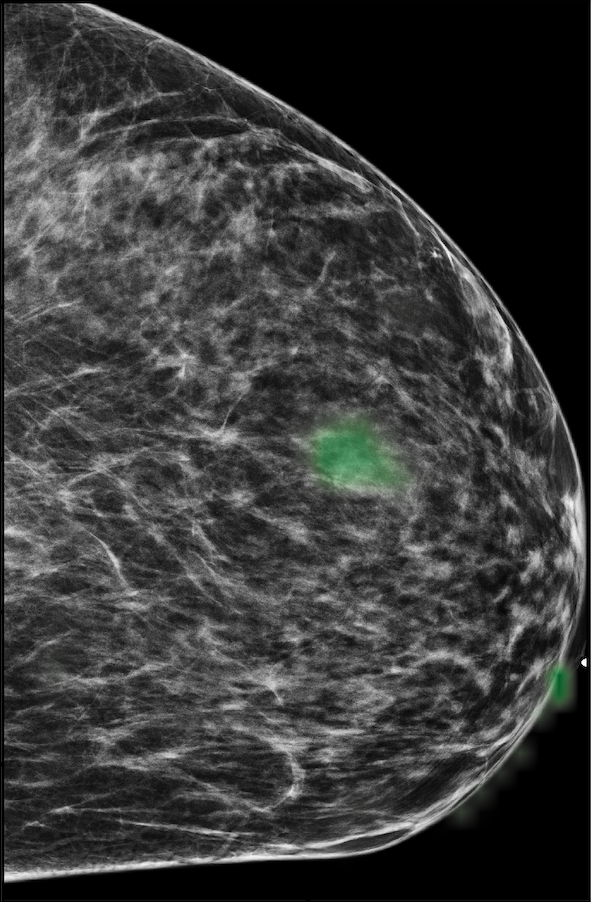} & 
  \includegraphics[width=0.15\textwidth,height=120pt]{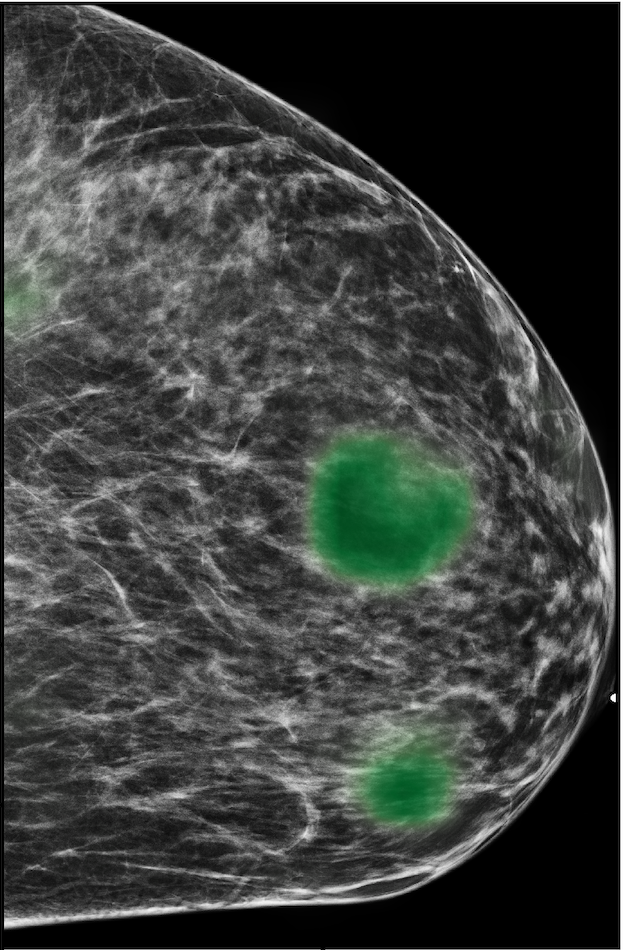} & 
  \includegraphics[width=0.15\textwidth,height=120pt]{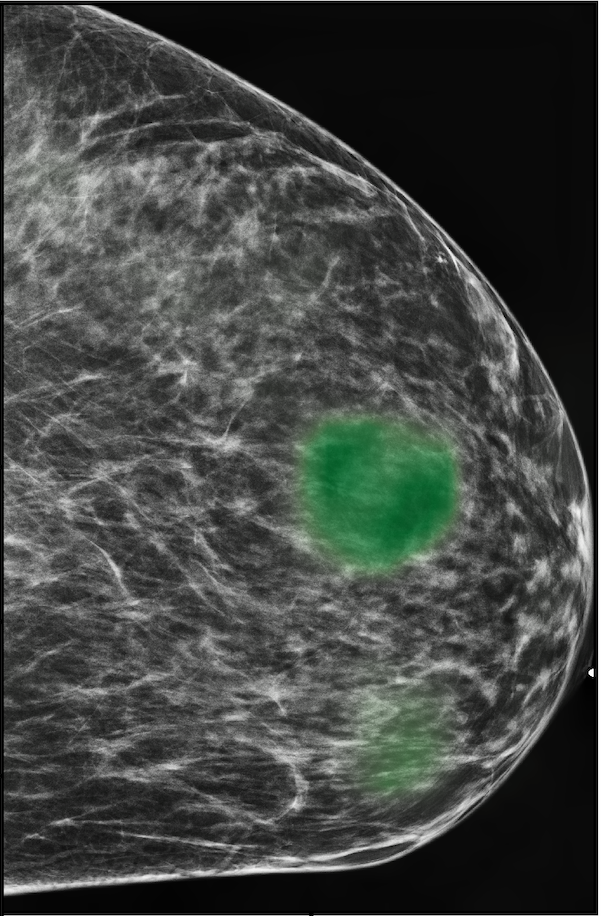} & 
  \includegraphics[width=0.15\textwidth,height=120pt]{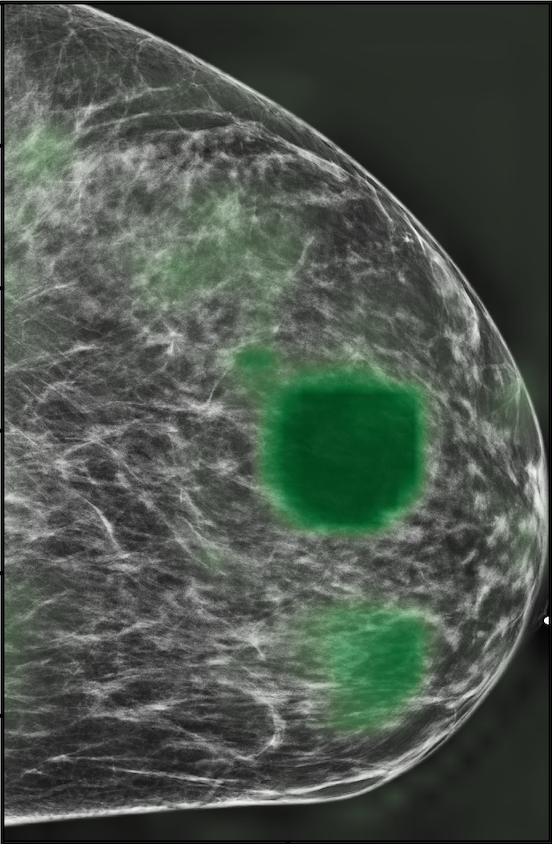}  \\
  
    & 0.33 & 0.52 & 0.39 & 0.38 & 0.18\\

  \includegraphics[width=0.15\textwidth,height=120pt]{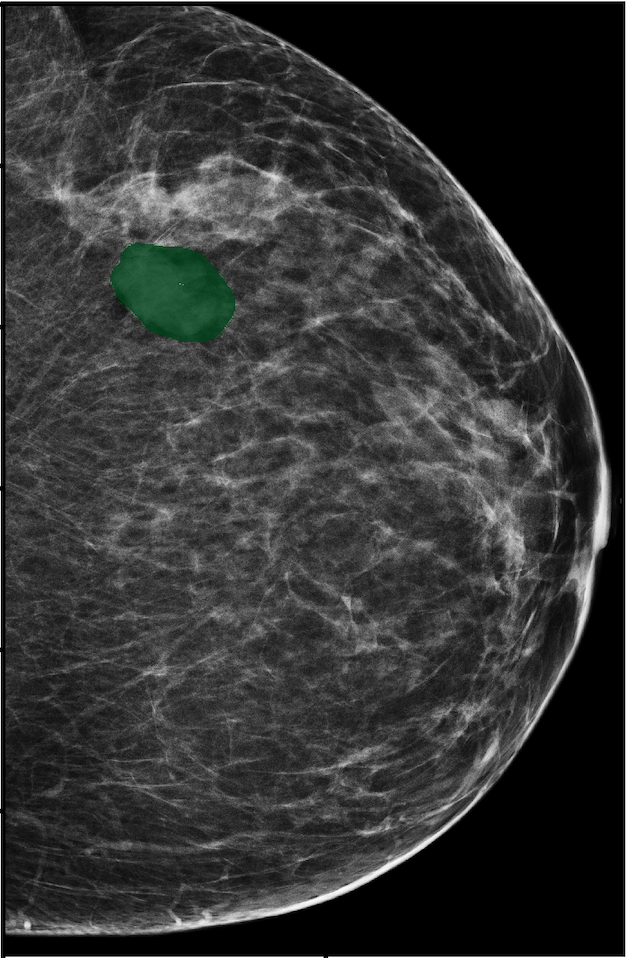} &
  \includegraphics[width=0.15\textwidth,height=120pt]{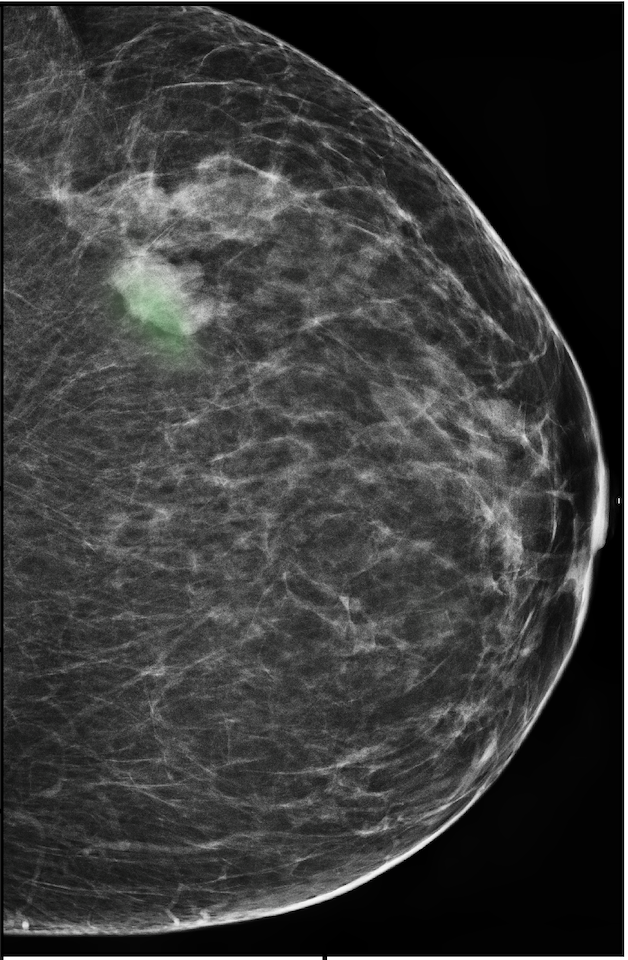} & \includegraphics[width=0.15\textwidth,height=120pt]{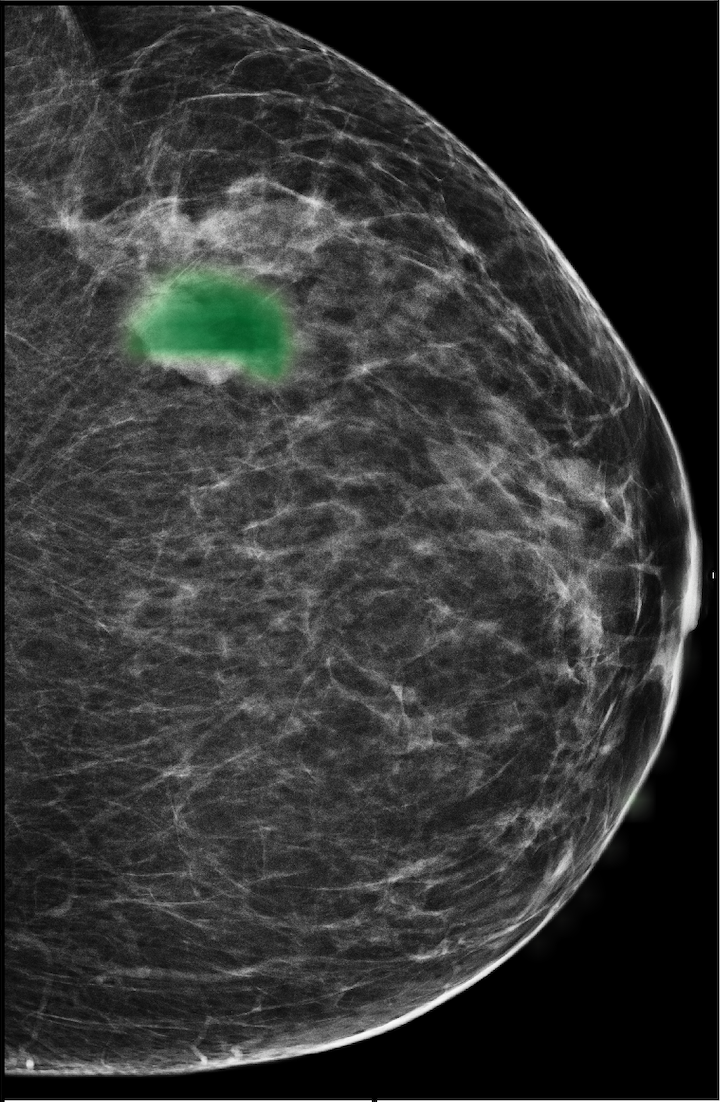} & 
  \includegraphics[width=0.15\textwidth,height=120pt]{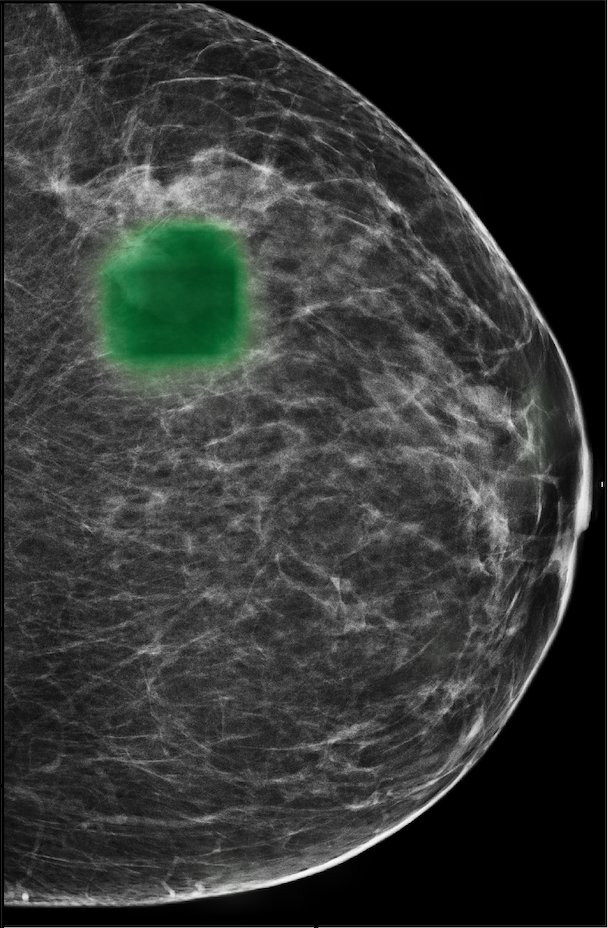} & 
  \includegraphics[width=0.15\textwidth,height=120pt]{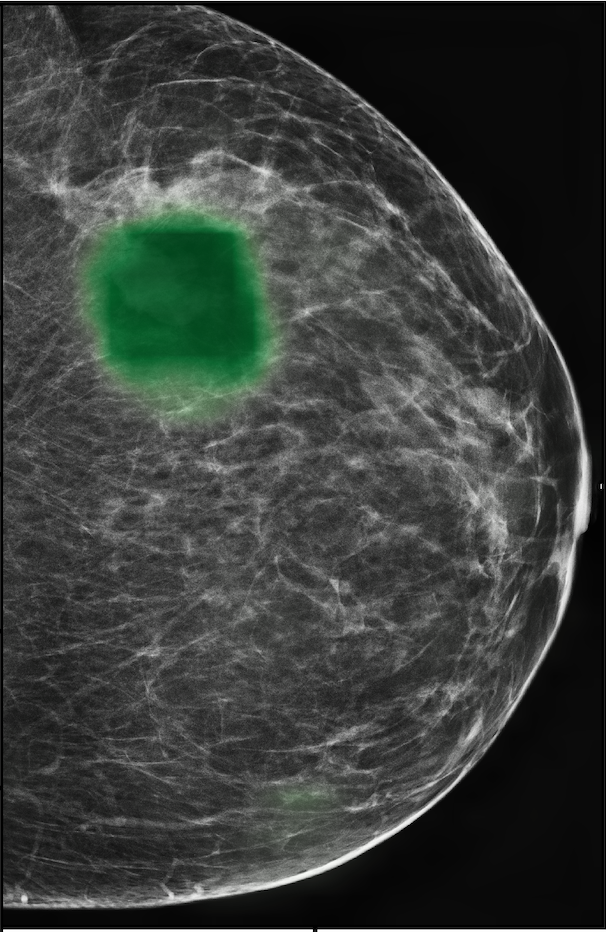} & 
  \includegraphics[width=0.15\textwidth,height=120pt]{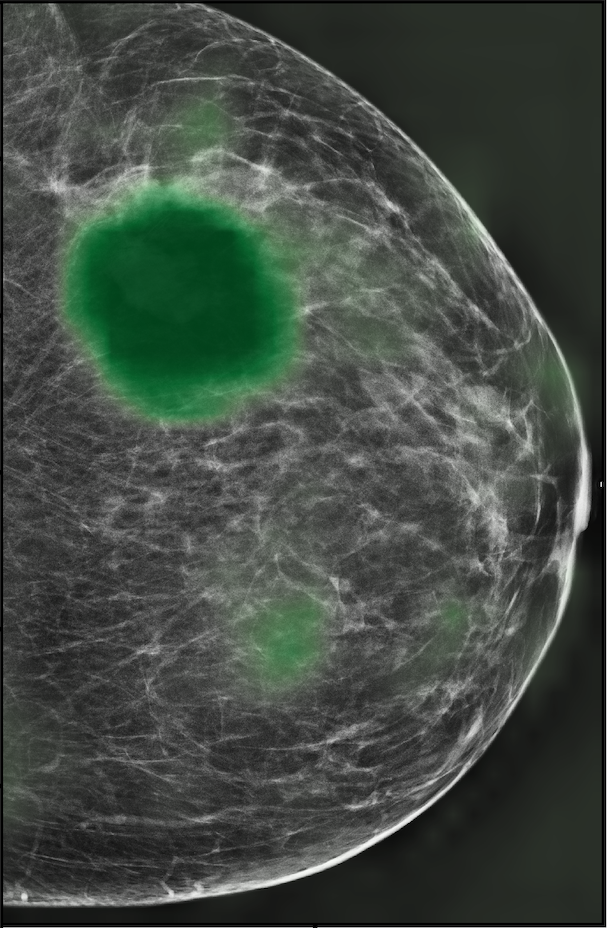}  \\
  
  & 0.34 & 0.73 & 0.62 & 0.49 & 0.30\\
  
  \includegraphics[width=0.15\textwidth,height=120pt]{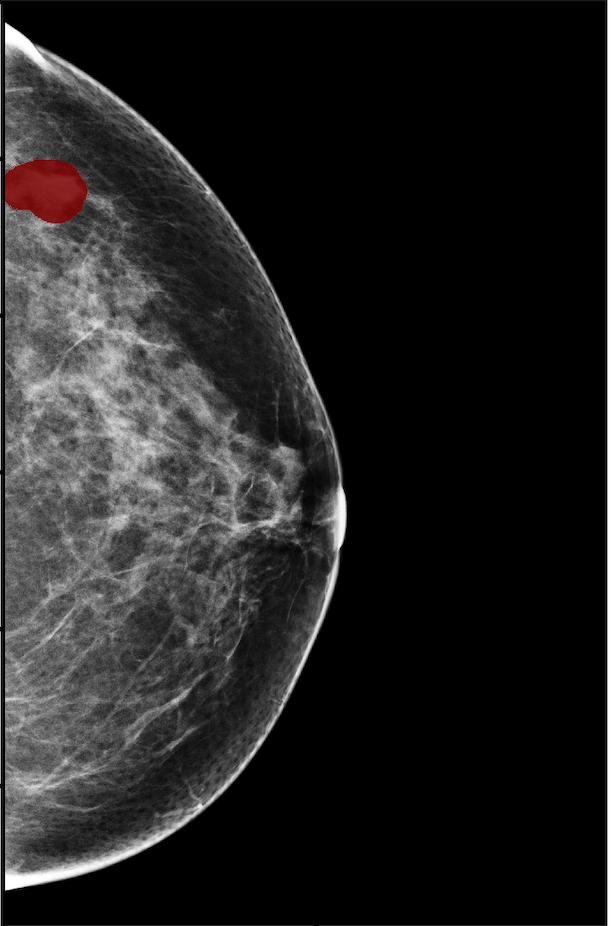} &
  \includegraphics[width=0.15\textwidth,height=120pt]{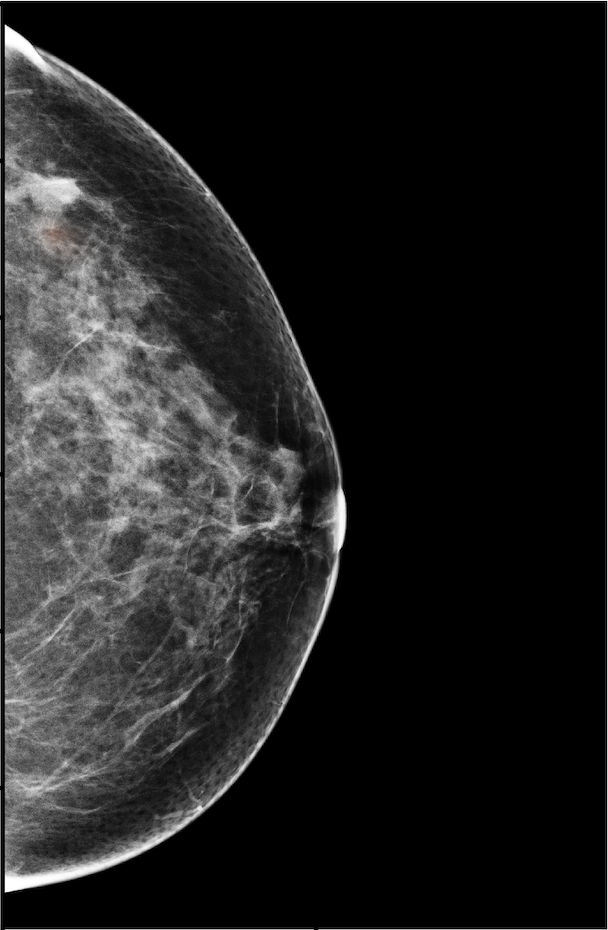} & \includegraphics[width=0.15\textwidth,height=120pt]{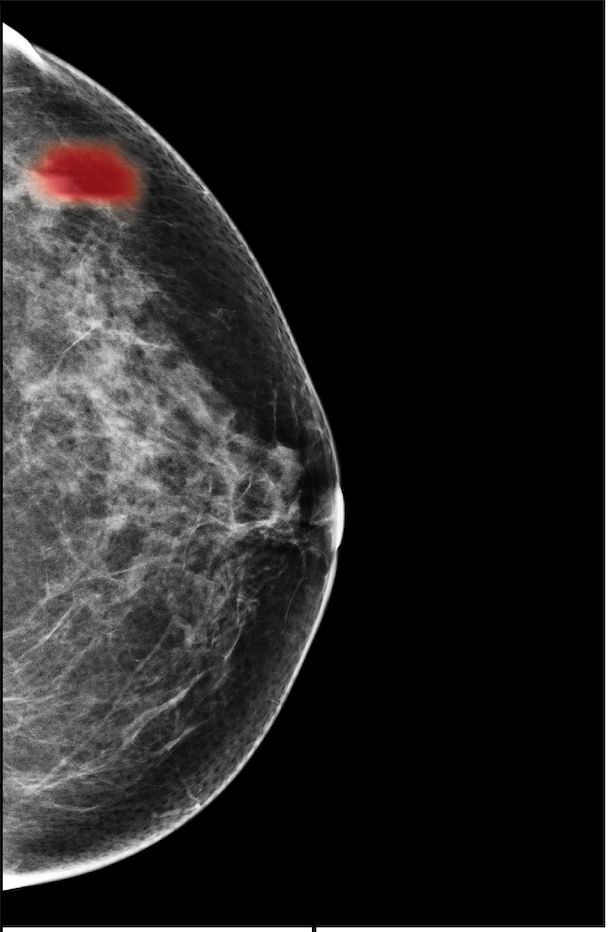} & 
  \includegraphics[width=0.15\textwidth,height=120pt]{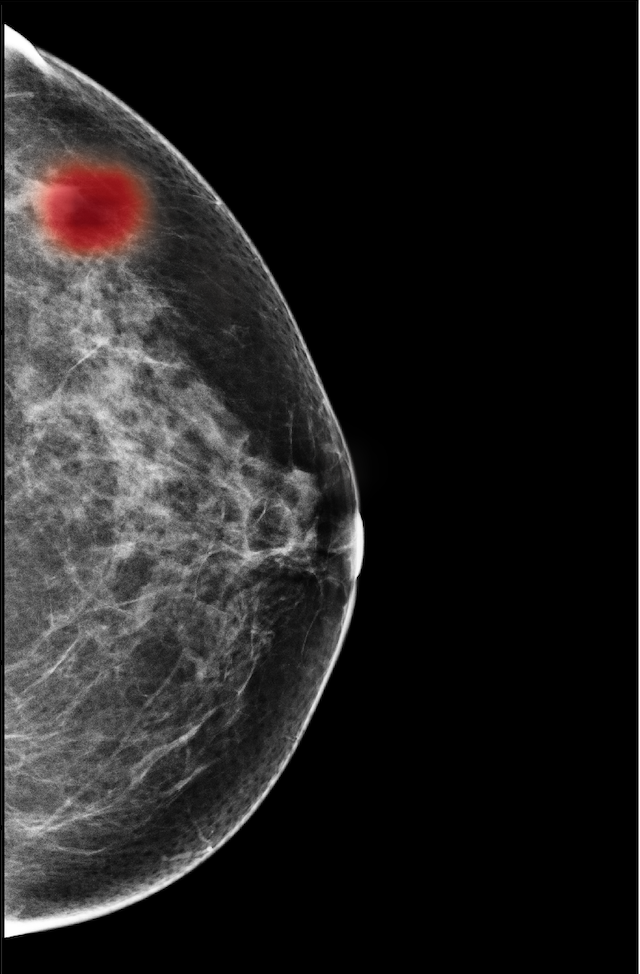} & 
  \includegraphics[width=0.15\textwidth,height=120pt]{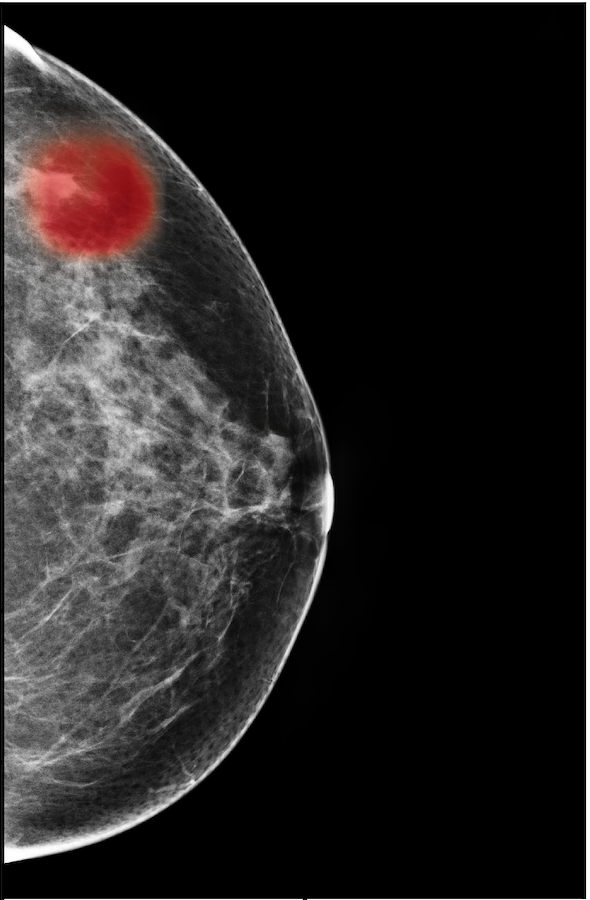} & 
  \includegraphics[width=0.15\textwidth,height=120pt]{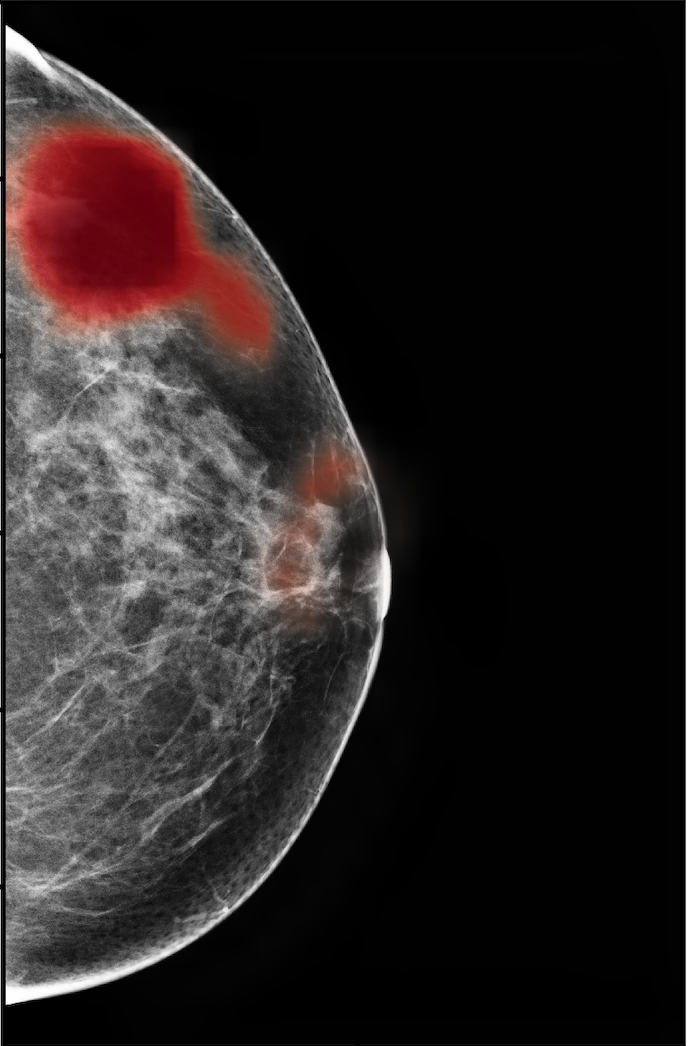}  \\
    & 0.04 & 0.42 & 0.50 & 0.38 & 0.27\\
  
  \includegraphics[width=0.15\textwidth,height=120pt]{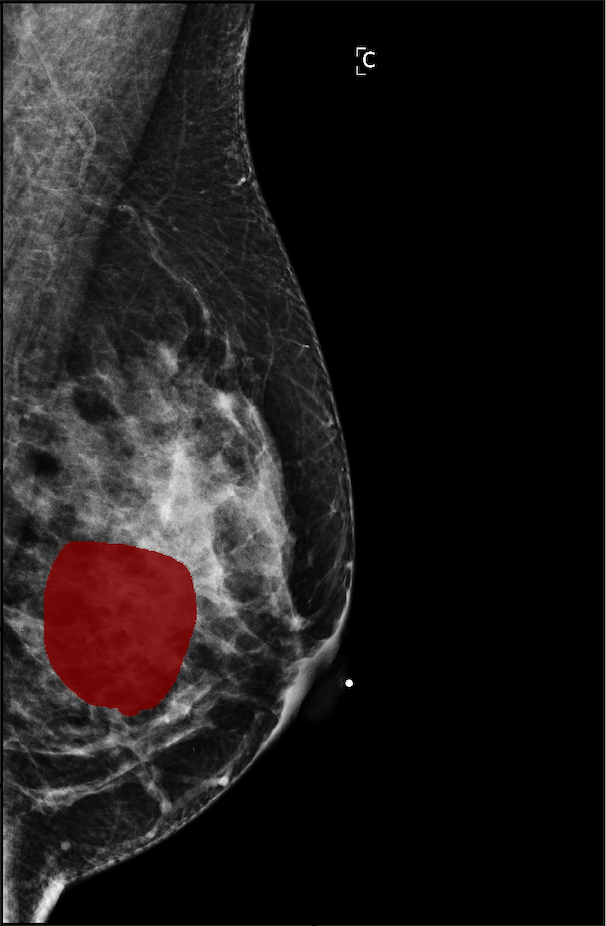} &
  \includegraphics[width=0.15\textwidth,height=120pt]{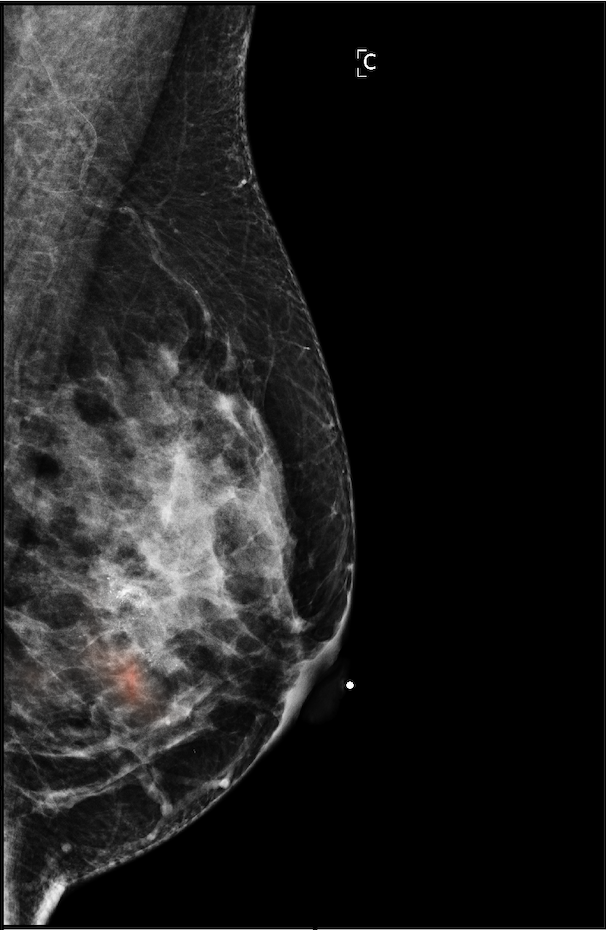} & \includegraphics[width=0.15\textwidth,height=120pt]{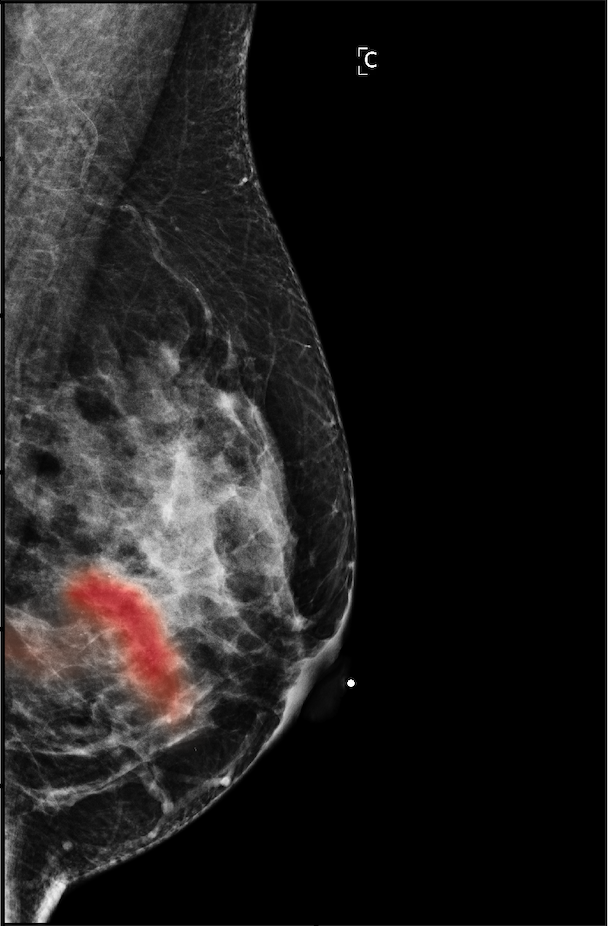} & 
  \includegraphics[width=0.15\textwidth,height=120pt]{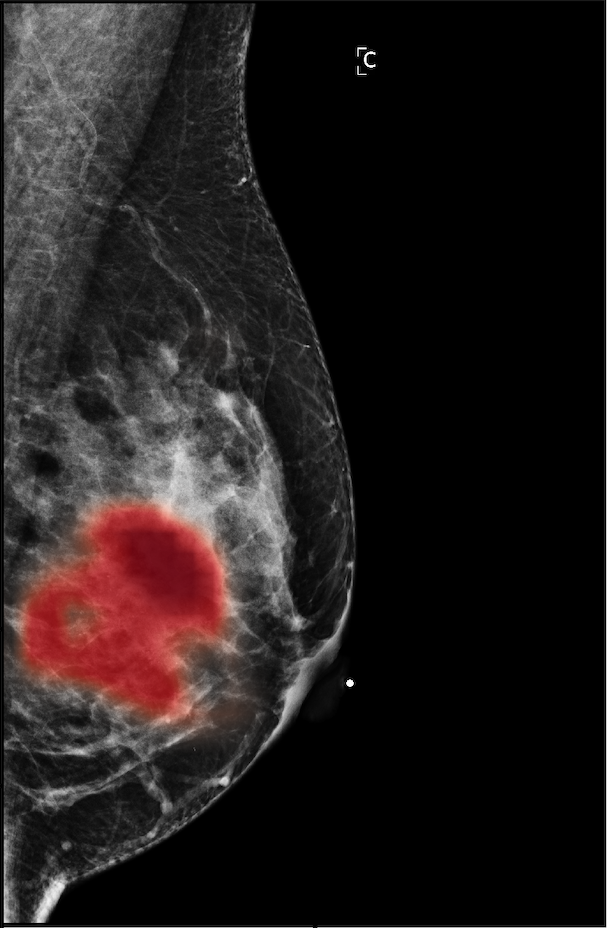} & 
  \includegraphics[width=0.15\textwidth,height=120pt]{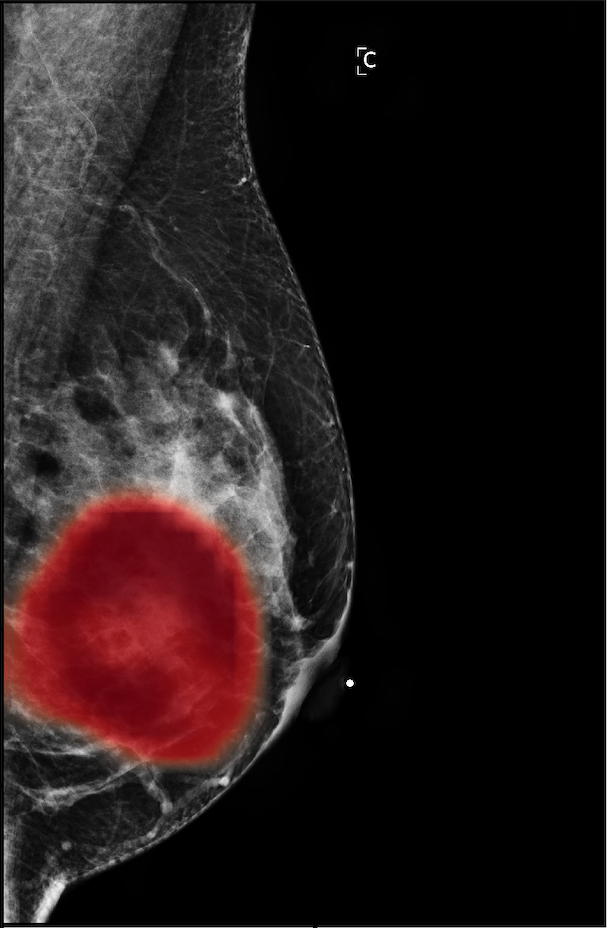} & 
  \includegraphics[width=0.15\textwidth,height=120pt]{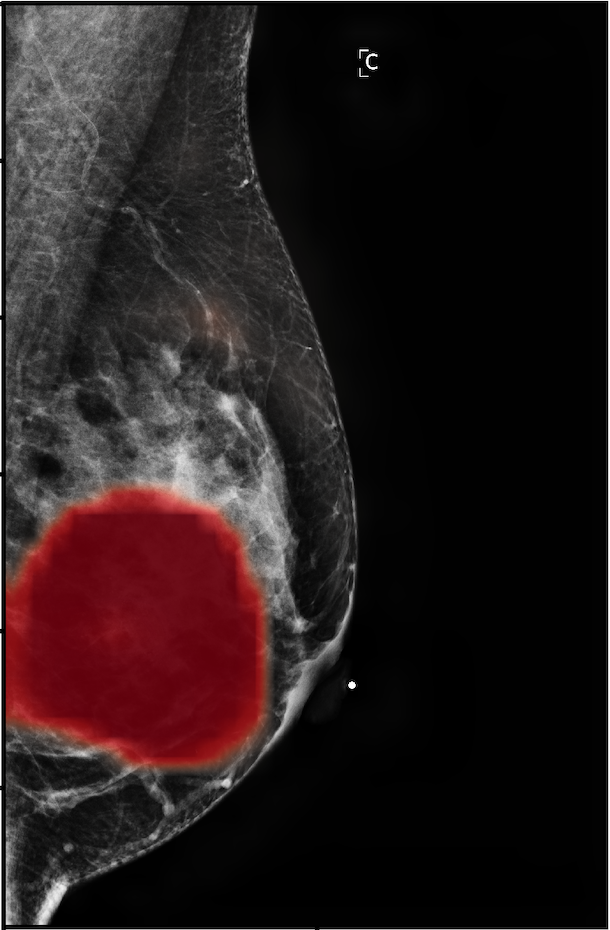}  \\
  & 0.14 & 0.53 & 0.77 & 0.63 & 0.55\\
    \end{tabular}  
\caption{In this figure we illustrate the effect of $t$ in the pooling function on the saliency maps. From left to right: the mammogram with ground truth segmentation and the saliency map generated using GMP, top $3\%$ pooling, top $10\%$ pooling, top $20\%$ pooling, and GAP. The corresponding DSC is specified below each saliency map. A benign lesion is found in the top two examples. A malignant lesion is found in the bottom two examples.}
  \label{fg:top_k_vis}
\end{figure*}

\subsection{Ablation Study}
\label{sec:ablation_study}
We performed ablation studies to explore the effectiveness of global module, local module, fusion module, patch-level attention, and the proposed \textit{top t\% pooling}. In addition, we also assess how much performance of GMIC could be improved by utilizing the pixel-level labels and ensembling GMIC with DMV-CNN and Faster R-CNN. All ablation experiments are based on the GMIC-ResNet-18 model.

\begin{table}
    \centering
      \caption{Ablation study: effectiveness of incorporating both global and local features. We report the mean and standard deviation of the test AUC for \textit{top-5} GMIC-ResNet-18. We experimented with 4 GMIC variants that use $\hat{\mathbf{y}}_\text{global}$, $\hat{\mathbf{y}}_\text{local}$, the average of $\hat{\mathbf{y}}_\text{global}$ and $\hat{\mathbf{y}}_\text{local}$, and $\hat{\mathbf{y}}_\text{fusion}$ as predictions. The proposed design that uses $\hat{\mathbf{y}}_\text{fusion}$ as predictions outperforms all variants.}
    \begin{tabular}{|c|c|c|}
    \hline
     Prediction & AUC(M) & AUC(B) \\ \hline
      $\hat{\mathbf{y}}_\text{global}$ & $0.892 \pm 0.009$ & $0.776 \pm 0.004$ \\
    \hline
    $\hat{\mathbf{y}}_\text{local}$ &  $0.897 \pm 0.004$ & $0.778 \pm 0.005$ \\
    \hline
    $\frac{1}{2}(\hat{\mathbf{y}}_\text{local} + \hat{\mathbf{y}}_\text{global})$  &  $0.905 \pm  0.006$ & $0.785 \pm 0.004$ \\
    \hline
    $\hat{\mathbf{y}}_\text{fusion}$  &  $\mathbf{0.913} \pm 0.007$ & $\mathbf{0.791} \pm 0.005$ \\
    \hline
  \end{tabular}
  \label{prediction_tb}
\end{table}

\paragraph{Synergy of Global and Local Information} In the preliminary version of GMIC~\citep{shen2019globally}, the final prediction is defined as $\frac{1}{2}(\hat{\mathbf{y}}_\text{global} + \hat{\mathbf{y}}_\text{local})$. In this work, we enhance GMIC with a fusion module that combines signals from both global features and local details. To empirically evaluate the effectiveness of the fusion module, we compared the performance achieved using only global features ($\hat{\mathbf{y}}_\text{global}$), only local patches ($\hat{\mathbf{y}}_\text{local}$), the average prediction of two modules ($\frac{1}{2}(\hat{\mathbf{y}}_\text{global} + \hat{\mathbf{y}}_\text{local})$), and the fusion of the two ($\hat{\mathbf{y}}_\text{fusion}$). As shown in Table \ref{prediction_tb}, $\hat{\mathbf{y}}_\text{fusion}$ achieved a higher AUC consistently for classifying both benign and malignant lesions than either $\hat{\mathbf{y}}_\text{global}$ or $\hat{\mathbf{y}}_\text{local}$. This result suggests that the fusion module helps GMIC to aggregate signals from both global and local module. Moreover, $\hat{\mathbf{y}}_\text{fusion}$ also outperforms the ensemble prediction $\frac{1}{2}(\hat{\mathbf{y}}_\text{local} + \hat{\mathbf{y}}_\text{global})$, which further demonstrates that the fusion module promotes an effective synergy beyond an ensembling effect created from averaging predictions over two sets of parameters.

\begin{table}[ht]
    \centering
      \caption{To evaluate the effectiveness of the patch-wise attention, we compare the proposed model with the variant (uniform) that always assigns equal attention to all patches. To investigate the importance of the localization information in the saliency maps, we trained another variant (random) that randomly selects patches from the input image. We use GMIC-ResNet-18 model with \textit{top 3\% pooling} as the base model. The performance of the local module ($\hat{\mathbf{y}}_\text{local}$) is reported.}
      \resizebox{\columnwidth}{!}{
    \begin{tabular}{|c|c|c|c|}
    \hline
     Attention & ROI patches & AUC(M)  & AUC(B)  \\ \hline
    uniform & \texttt{retrieve\_roi} & $0.874 \pm 0.008$ & $0.776 \pm 0.007$ \\ \hline
    gated & random &  $0.629 \pm 0.042$ & $0.658 \pm 0.011$ \\ \Xhline{3\arrayrulewidth}
    gated & \texttt{retrieve\_roi} & $\mathbf{0.898} \pm 0.01$ & $\mathbf{0.78} \pm 0.008$  \\ \hline
  \end{tabular}}
  \label{attn_roi_tb}
\end{table}

\begin{table}[htb!]
    \centering
      \caption{Ablation study: effect of different choice of aggregation function. We report the performance achieved by parameterizing $f_\text{agg}$ as global average pooling (GAP), global maximum pooling (GMP), and \textit{top $t\%$ pooling}. For each setting, we trained five GMIC-ResNet-18 models and report the mean and standard deviation of AUC and DSC.}
    \resizebox{\columnwidth}{!}{
    \begin{tabular}{|l|c|c|c|c|}
    \hline
     $f_\text{agg}$ & AUC(M) & AUC(B) & DSC(M) & DSC(B)\\ \hline
      GMP & $0.890 \pm 0.02$  & $0.785 \pm 0.012$ & $0.127 \pm 0.052 $ & $0.103 \pm 0.060$\\ \hline
      
      $t=1\%$ & $0.906 \pm 0.01$  & $0.784 \pm 0.007$ & $0.190 \pm 0.030$ &  $0.147 \pm 0.053$\\ \hline
      
      $t=2\%$ & $\textbf{0.916} \pm 0.009$  & $0.790 \pm 0.007$ & $0.203 \pm 0.013$ &$0.191 \pm 0.042$\\ \hline
      
      $t=3\%$ & $0.913 \pm 0.007$  & $\textbf{0.791} \pm 0.004$ & $\textbf{0.228} \pm 0.036$ &  $0.178 \pm 0.041$\\ \hline
      
      $t=5\%$ & $0.912 \pm 0.009$  & $0.790 \pm 0.002$ & $0.172 \pm 0.004$  & $\textbf{0.194} \pm 0.027$ \\ \hline
      
      $t=10\%$ & $0.914 \pm 0.005$  & $\textbf{0.791} \pm 0.008$ & $0.156 \pm 0.050$ & $0.182 \pm 0.028$\\ \hline
      
      $t=20\%$ & $0.907 \pm 0.017$ & $0.785 \pm 0.008$ & $0.126 \pm 0.048$  & $0.182 \pm 0.040$\\ \hline
      
      GAP & $0.903 \pm 0.02$  & $0.783 \pm 0.012$ & $0.065 \pm 0.006$  & $0.181 \pm 0.011$\\ \hline
  \end{tabular}}
  \label{tb:topk_tb}
\end{table}

\paragraph{ROI Proposals and Patch-wise Attention} GMIC applies two mechanisms to control the quality of patches provided to the local module. First, the \texttt{retrieve\_roi} algorithm utilizes localization information from the saliency maps and greedily selects informative patches of the input image. Those selected patches are then weighted using the Gated Attention network. To evaluate the effectiveness of both mechanisms, we trained two variants: one (uniform) that always assigns equal attention score to each patch and another (random) that randomly samples patches without using the saliency map. As shown in Table~\ref{attn_roi_tb}, if patch-wise attention is disabled, the AUC of classifying malignant lesions decreases from 0.898 to 0.874. If the \texttt{retrieve\_roi} algorithm is replaced with random sampling, the local module suffers from a significant performance decrease. These results suggest that both the patch-wise attention and \texttt{retrieve\_roi} procedure are essential for the local module to make accurate predictions.

\paragraph{Aggregation Function} In order to study the the impact of the aggregation function, we experimented with 8 parameterizations of $f_\text{agg}$ including GAP, GMP, and \textit{top $t\%$ pooling} with $t \in \{1,2,3,5,10,20\}$. For each parameterization, we fixed other hyperparameters and trained five GMIC-ResNet-18 models with randomly initialized weights. In table \ref{tb:topk_tb}, we report the AUC and DSC achieved by each value of $t$. GMIC-ResNet-18 achieves the highest AUC on identifying malignant cases when using \textit{top $t\%$ pooling} with $t=2$. The performance of \textit{top $t\%$ pooling} decreases as $t$ moves away from $2$ and converges to that of GAP/GMP when $t$ is large/small. This observation is consistent with the intuition that GAP and GMP are two extremes of \textit{top $t\%$ pooling}. We observe a similar but less pronounced trend on the AUC of identifying benign cases.

GMIC-ResNet-18 also obtains better localization performance with \textit{top $t\%$ pooling} than with GAP or GMP. The highest DSC for localizing malignant and benign lesions is achieved when $t$ is set to $3\%$ and $5\%$ respectively. To further study the effect of $t$, we visualize the saliency maps for four examples selected from the test set. As illustrated in Figure \ref{fg:top_k_vis}, when $t$ is small, the saliency maps tend to highlight a small area. When $t$ is large, the highlighted region grows. Ideally, the choice of $t$ should reflect the true size of lesions contained in the image and different images could use different $t$. In future research, we propose to learn $t$ using information within the image.

\paragraph{Number of ROI patches} 
\begin{figure}[h!]
    \includegraphics[width = 0.45\textwidth]{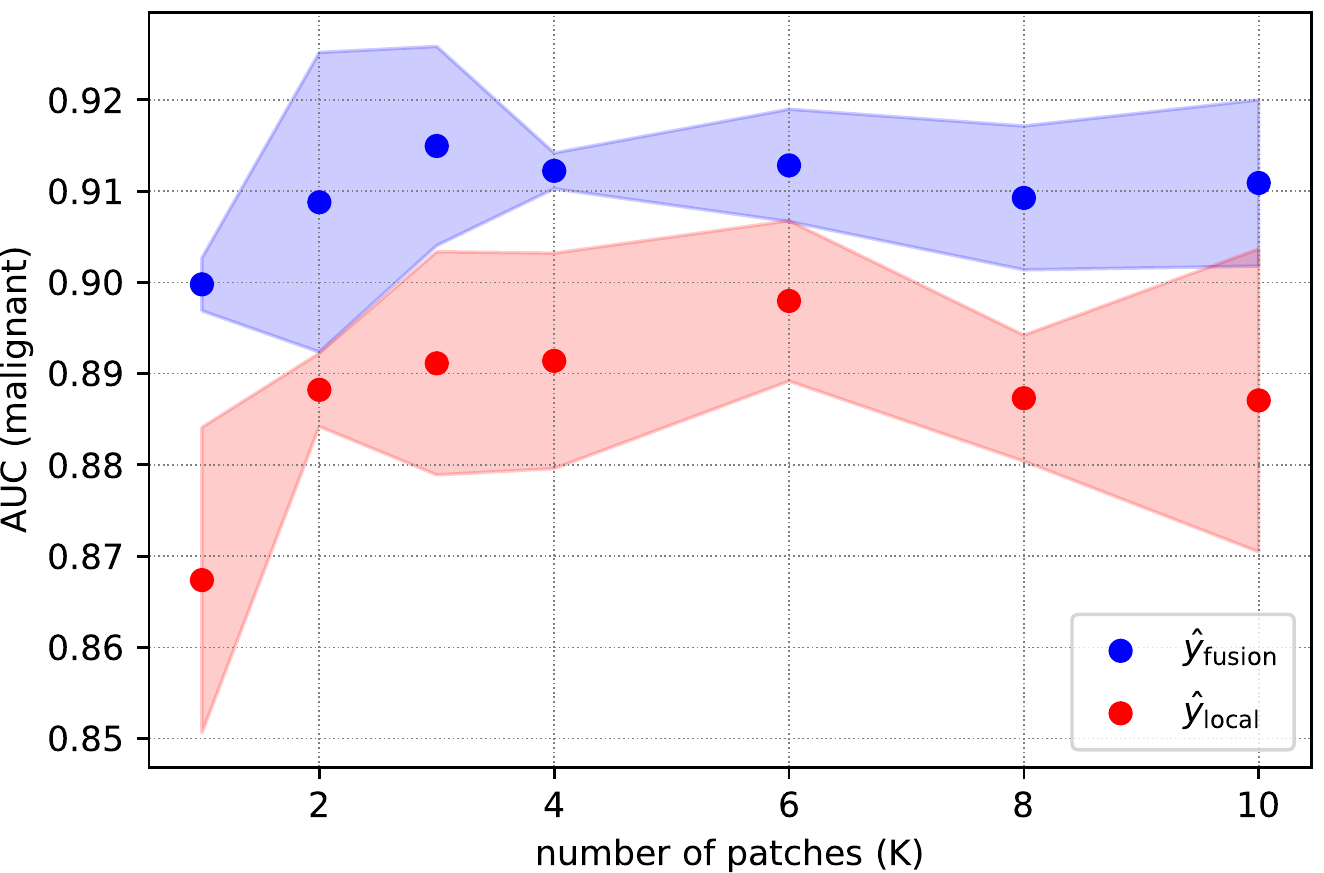}  \\
    \includegraphics[width = 0.45\textwidth]{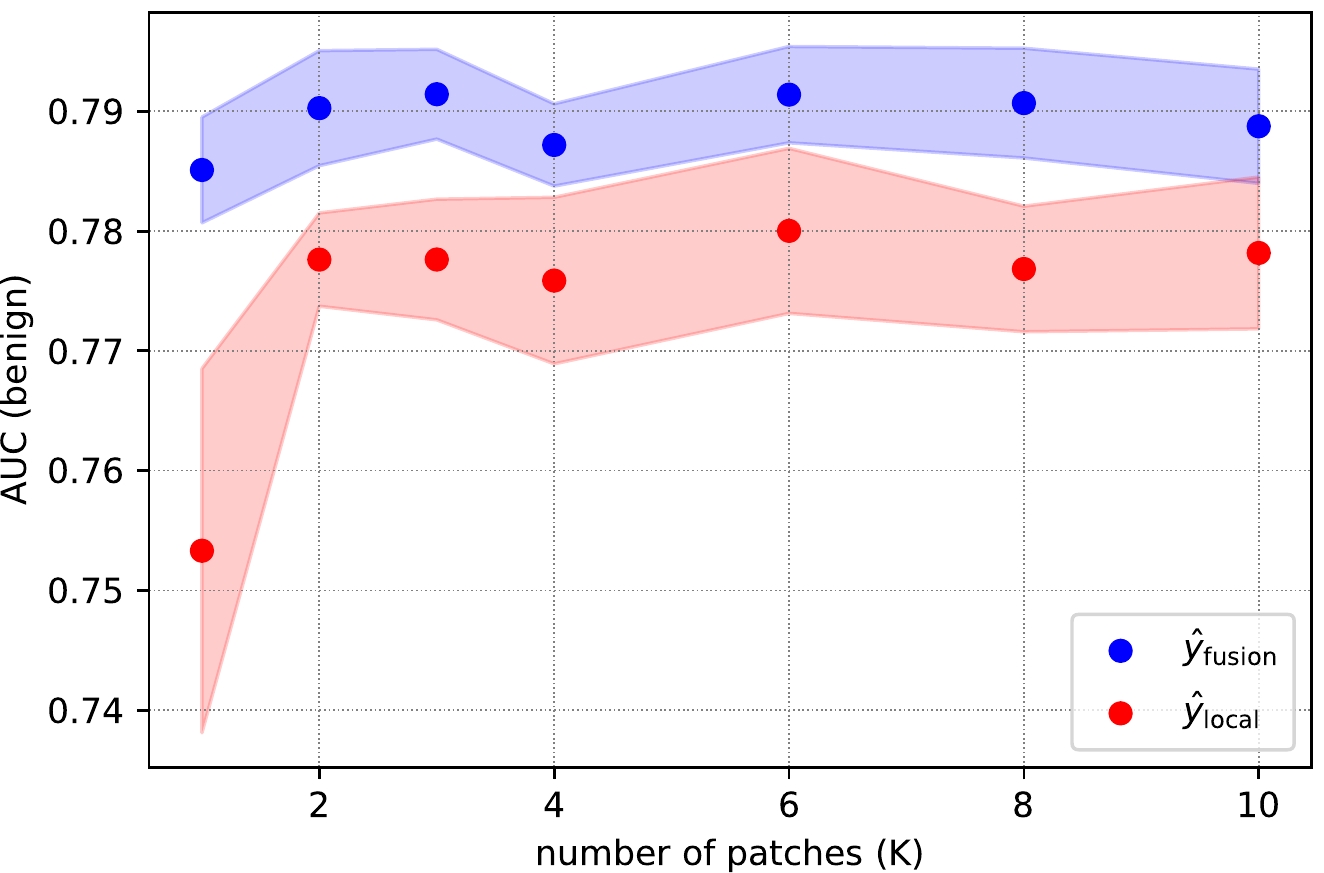}
    \vspace{-2mm}
    \caption{The classification performance of GMIC-ResNet-18 with a varying number of patches $K \in \{1,2,3,4,6,8,10\}$. For each $K$, we trained five models and reported the mean and the standard deviation of test AUC on classifying malignant (top) and benign (bottom) lesions. We show the performance of both $\hat{\mathbf{y}}_\text{fusion}$ and $\hat{\mathbf{y}}_\text{local}$. The performance saturates for $K > 3$.}
    \label{fig:ab_pnum}
    \vspace{-3mm}
\end{figure}

We experimented with GMIC varying the number of patches $K \in \{1,2,3,4,6,8,10\}$. For each setting, we trained five GMIC-ResNet-18 models with \textit{top $t\%$ pooling} ($t = 3\%$). In Figure \ref{fig:ab_pnum}, we illustrate the mean and the standard deviation of AUC achieved by $\hat{\mathbf{y}}_\text{fusion}$ and $\hat{\mathbf{y}}_\text{local}$ on classifying benign and malignant lesions. Increasing $K$ improves the classification performance when $K$ is small. The improvement is more evident on $\hat{\mathbf{y}}_\text{local}$ than $\hat{\mathbf{y}}_\text{fusion}$, because $\hat{\mathbf{y}}_\text{fusion}$ also utilizes global features. However, for $K > 3$, the classification performance saturates. This observation demonstrates a trend of diminishing marginal return of incorporating additional ROI patches.

\paragraph{Utilizing Segmentation Labels} 

\begin{figure}[ht]
    
    \includegraphics[width = 0.16\textwidth,trim={0 0 0 0}]{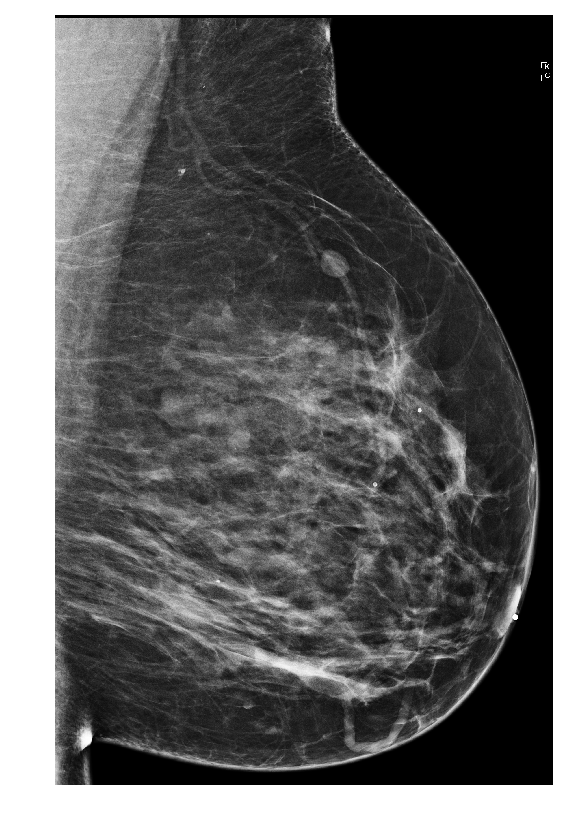}  
    \hspace{-7pt}
    \includegraphics[width = 0.16\textwidth,trim={0 0 0 0}]{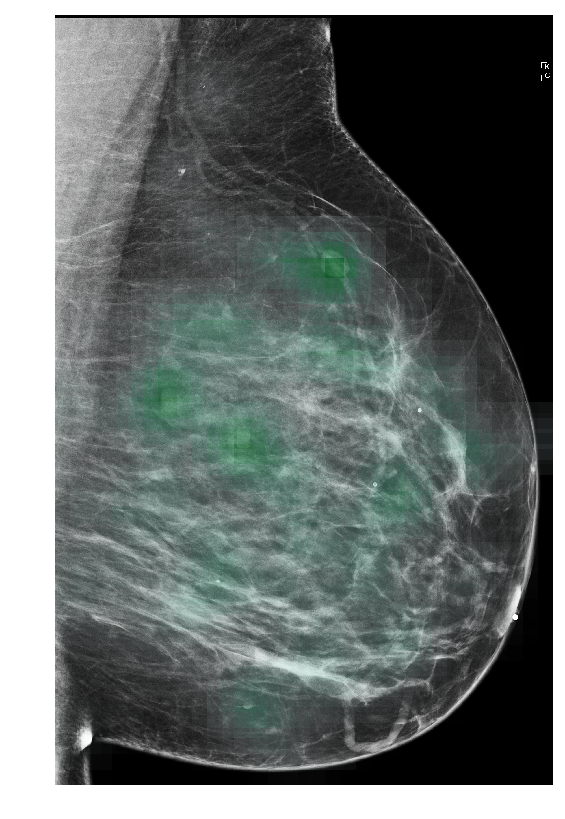}
    \hspace{-7pt}
    \includegraphics[width = 0.16\textwidth,trim={0 0 0 0}]{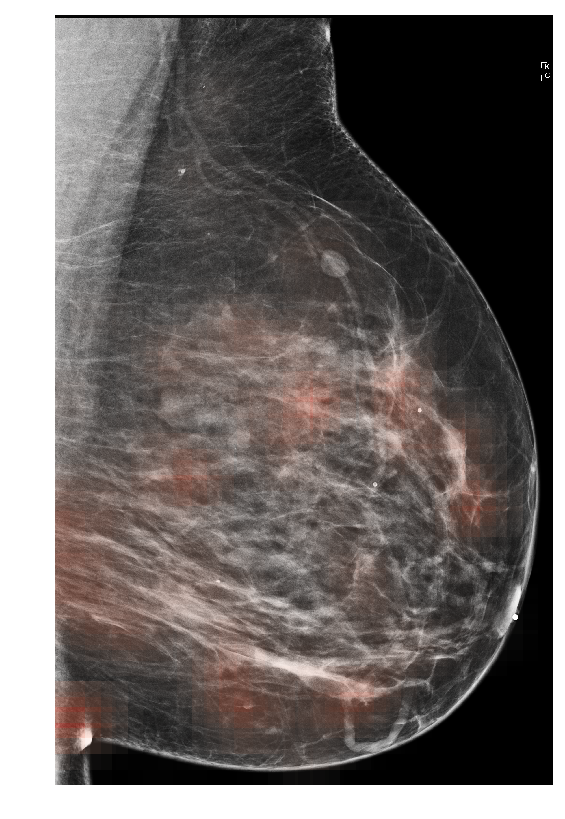}
    \caption{Example heatmaps generated by the patch-level model proposed by~\cite{wu2019deep}. The original image (left), the ``benign'' heatmap over the image (middle), and the ``malignant'' heatmap over the image (right).}
    \label{fig:heatmap_example}
\end{figure}

We also assessed how much performance of GMIC could be improved by utilizing pixel-level labels during training. Following~\cite{wu2019deep}, we used the pixel-level labels to train a patch-level model which classifies $256 \times 256$-pixel patches of mammograms, making two predictions: the presence or absence of malignant and benign findings in a given patch. We then apply the patch-level classifier to each full-resolution image in a sliding window fashion to create two heatmaps (illustrated in Figure~\ref{fig:heatmap_example}), one containing an estimated probability of a malignant finding for each pixel, and the other containing an estimated probability of a benign finding. In this comparison study, we concatenated the input images with these two heatmaps\footnote{The two heatmap channels are only used by the global network $f_g$. The local network $f_l$ does not use them.} to train 30 GMIC-ResNet-18 models (referred as GMIC-ResNet-18-heatmap models) using the hyperparameter optimization setting described in Section~\ref{sec:classification_performance}. We reported the test performance of the \textit{top-5} GMIC-heatmap models that achieved the highest validation AUC on identifying breasts with malignant lesions. The \textit{top-5} GMIC-ResNet-18-heatmap models achieved a mean AUC of $0.927\pm0.04$ / $0.792\pm0.008$ in identifying breasts with malignant/benign lesions, outperforming the vanilla GMIC models ($0.913\pm0.007$ / $0.791\pm0.005$). The ensemble of the \textit{top-5} GMIC-ResNet-18-heatmap models achieved an AUC of $0.931/0.80$ in identifying breasts with malignant/benign lesions matching the performance of vanilla GMIC models ($0.930$/$0.80$). While augmenting GMIC with heatmaps improves its classification performance, the improvement is marginal especially when comparing to the ensemble of models. We conjecture that, for a sufficiently large dataset, image-level labels alone are powerful enough to capture most of the signal, and additional localization information from the pixel-level segmentation labels only slightly improves the performance of GMIC. In fact, sometimes it might even be biasing the model towards ignoring mammographically-occult findings.

\begin{figure*}[ht!]
    \includegraphics[width = 0.47\textwidth,trim={0 0 0 0}]{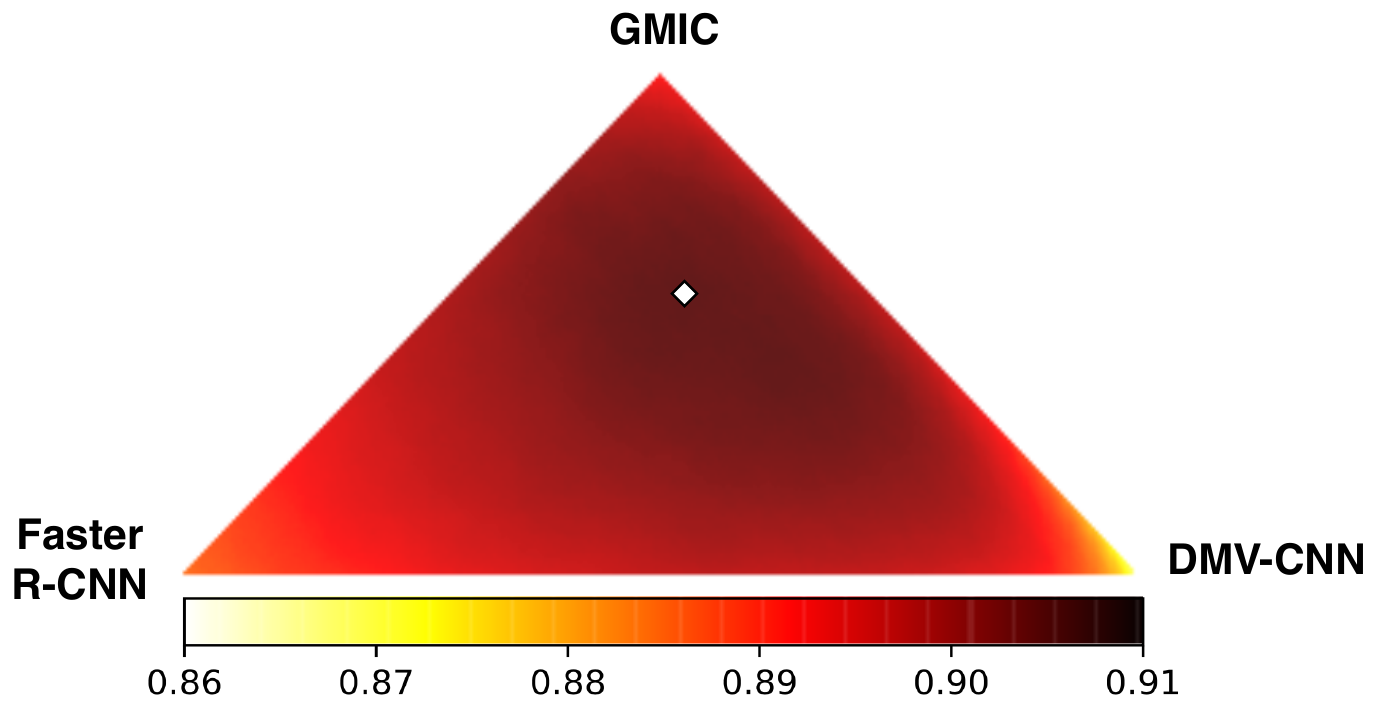}  
    \hspace{15pt}
    \includegraphics[width = 0.47\textwidth,trim={0 0 0 0}]{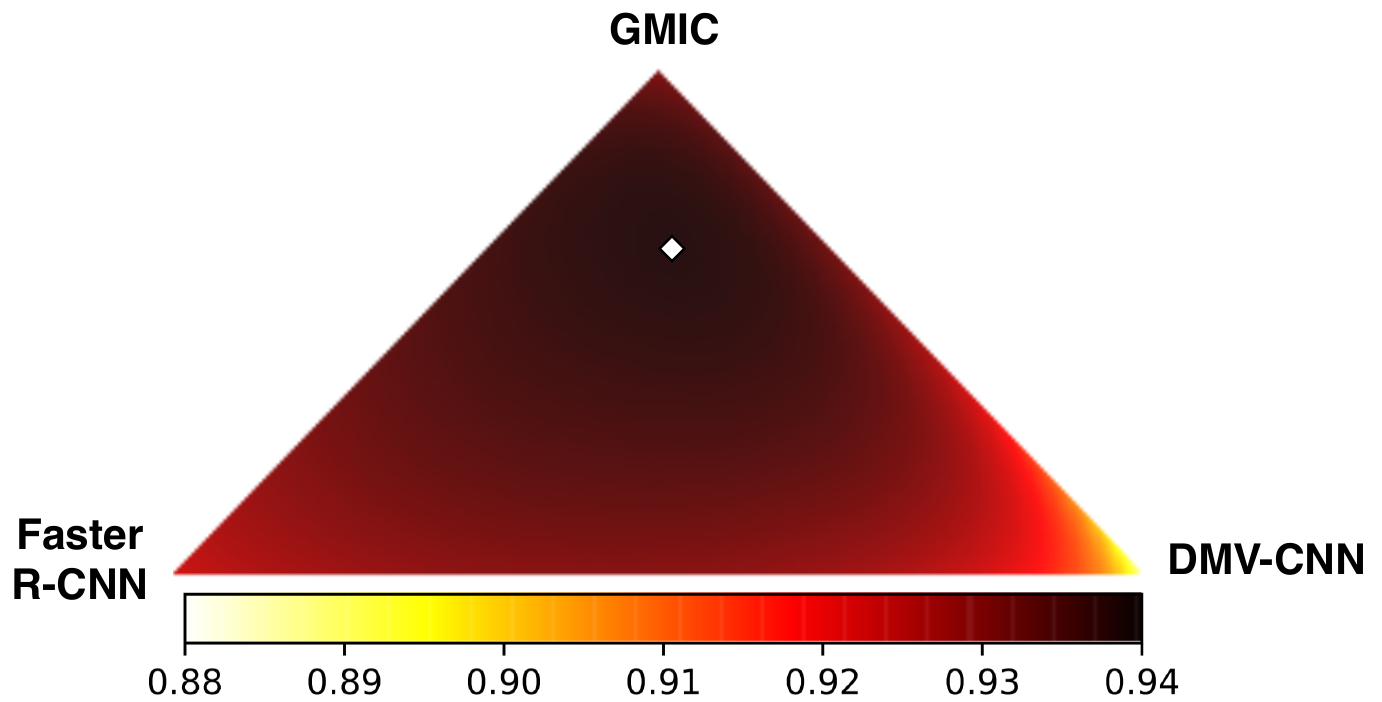}
    \caption{We visualize the AUC of identifying breasts with malignant findings achieved by the ensemble model with varying $\lambda_1$, $\lambda_2$, and $\lambda_3$ on the reader study dataset (left) and the test set (right). The optimal combination of $\lambda_1$, $\lambda_2$, and $\lambda_3$ that achieves highest AUC is highlighted in white diamond. The weight associated with GMIC is the largest among the three models for both datasets. On the reader study dataset, the optimal combination ($\lambda_1 = 0.56$, $\lambda_2=0.2$, $\lambda_3=0.24$) achieves an AUC of 0.905. On the test set, the optimal combination ($\lambda_1 = 0.65$, $\lambda_2=0.16$, $\lambda_3=0.19$) achieves an AUC of 0.939.}
    \label{fig:super_ensemble}
\end{figure*}

\paragraph{Ensembling GMIC with Other Models} 
In order to estimate a lower bound of what level of performance is possible to achieve on this task, we build a large ``super-ensemble'' of models by aggregating the predictions of: a) an ensemble of \textit{top-5} GMIC-ResNet-18, b) an ensemble of 5 DMV-CNN model (with heatmaps) \citep{wu2019deep}, and c) an ensemble of 3 Faster R-CNN models~\citep{fevry2019improving}.
Similar to the human-machine hybrid model, the predictions of the ensemble model are defined as $\hat{\mathbf{y}}_\text{ensemble} = \lambda_1 \hat{\mathbf{y}}_\text{GMIC} + \lambda_2 \hat{\mathbf{y}}_\text{Faster R-CNN} +  \lambda_3 \hat{\mathbf{y}}_\text{DMV-CNN}$ where $\lambda_1 + \lambda_2 + \lambda_3 = 1$. On the test set, the ensemble model with equal weights associated with each of its components ($\lambda_1=\lambda_2=\lambda_3=\frac{1}{3}$) achieves an AUC of 0.936 in identifying breasts with malignant lesions. We note that the improvement against \textit{top-5} GMIC-ResNet-18-ensemble (0.930) is small. We also note that utilizing this ensemble might be impractical, due to its complexity and computational cost.

We also checked what would be the AUC of this ensemble if we could tune the weighting coefficients of the ensemble on the test set. In Figure~\ref{fig:super_ensemble}, we visualize its classification performance on the reader study dataset and the full test set for different combinations of $\lambda_1$, $\lambda_2$ and $\lambda_3$. For the optimal combinations of $\lambda_1$, $\lambda_2$, and $\lambda_3$ that achieve the highest AUC on both datasets, the weight associated with GMIC ($\lambda_1$) is the largest, however, the two other weights are also non-negligible, suggesting that the three types of models are complementary, even though the improvement in terms of AUC is small.

\section{Related Work} 
\label{sec:related_works}

\subsection{High-resolution 2D Medical Image Classification}
The increased resolution level of medical images has posed new challenges for machine learning. Early works on applying deep neural networks to medical image classification typically utilize a CNN acting on the entire image to generate a prediction, resembling approaches developed for object classification in natural images. For instance, \cite{roth2015anatomy} adopted a 5-layer CNN to perform anatomical classification of CT slices. A similar approach was adopted by \cite{codella2015deep} to recognize melanoma on dermoscopy images. More recently, \cite{rajpurkar2017chexnet} fine-tuned a 121-layer DenseNet~\citep{huang2016densely} to classify thorax disease on chest X-ray images. However, this line of work suffers from two drawbacks. Unlike many natural images in which ROIs are sufficiently large, ROIs in medical images are typically small and sparsely distributed over the image. Applying a CNN indiscriminately over the entire image may include a considerable level of noise outside the ROI. Moreover, input images are commonly downsampled to fit in GPU memory. Aggressively downsampling medical images could distort important details making the correct diagnosis difficult~\citep{high_resolution}.

In another line of research, input images are uniformly divided into small patches. A classifier is trained and applied to each patch, and patch-level predictions are aggregated to form an image-level prediction. This family of methods has been commonly applied to the segmentation and classification of pathology images~\citep{campanella2019clinical,sun2019accurate,sun2019prediction}. \cite{coudray2018classification} used Inception V3~\citep{szegedy2016rethinking} on tiles of whole-slide histopathology images to detect adenocarcinoma and squamous cell carcinoma. \cite{sun2019accurate} proposed a multi-scale patch-level classifier using dilated convolutions to localize gastric cancer regions. For breast cancer screening, \cite{wu2019deep} utilized patch-level predictions as additional input channels to classify screening mammograms. A major limitation of these methods is that many of them require lesion locations to train the patch-level classifiers, which might be expensive to obtain. Moreover, global information such as the image structure could be lost by dividing input images into small patches.

Instead of applying patch-level model on all tiles, several methods have been proposed to select patches that are related to the classification task. \cite{zhong2019attention} suggested selecting important patches based on a coarse attention map generated by applying an UNet~\citep{ronneberger2015u} on downsampled input images. \cite{guo2019deep} adopted a similar strategy to detect strut points on intravascular optical coherence tomography images. \cite{guan2018diagnose} further developed this idea and proposed the attention guided convolution neural network (AG-CNN) that explicitly merges information from both the global image and a refined local patch to detect thorax disease on chest X-ray images. Our work is perhaps most similar to \cite{guan2018diagnose}. While AG-CNN only selects one patch for each class, our method is able to selectively aggregate information from a variable number of patches, which enables the model to learn from broader source of signal.

\subsection{Breast Cancer Classification in Mammography}
Early works on breast cancer screening
exam classification were computer-aided detection (CAD) systems built with handcrafted features~\citep{li2001false,wu2007bilateral,masotti2009computer,oliver2010review}. Despite their popularity, clinical study has suggested that CAD systems do not improve diagnostic accuracy~\citep{lehman2015diagnostic}. With the advances in deep learning in the last decade~\citep{lecun2015deep}, neural networks have been extensively applied to assist radiologists in interpreting screening mammograms~\citep{zhu2017deep,wu2018breast,rampun2019breast,mckinney2020international}. In particular, \cite{high_resolution} adopted a multi-view CNN that jointly utilizes information from four standard views to classify the BI-RADS category associated with mammograms. To accurately detect small lesions on mammograms, segmentation labels have been utilized to train patch-level classifiers~\citep{lotter2017multi,kooi2017classifying,shen2017end, teare2017malignancy, zhu2017deep, wu2019deep}. \cite{hagos2018improving} further designed a multi-input CNN that learns symmetrical difference among patches to detect breast masses. Another popular way of utilizing segmentation labels is to train anchor-based object detection models. For instance, \cite{ribli2018detecting} and \cite{fevry2019improving} fine-tuned a Faster RCNN~\citep{ren2015faster} to localize lesions on mammograms. \cite{xiao2019learning} integrated object detector in a Siamese structure with explicit loss terms to differentiate anchor proposals containing lesion from those with only normal tissues. We refer the readers to~\cite{hamidinekoo2018deep,gao2019new,geras2019artificial} for comprehensive reviews of prior works on machine learning for mammography.

\subsection{Weakly Supervised Object Detection} 
Recent progress demonstrates that CNN classifiers, trained with image-level labels, are able to perform semantic segmentation at the pixel level~\citep{oquab2015object,pinheiro2015image,bilen2016weakly,zhou2016learning,diba2017weakly,zeng2019joint}. This is commonly achieved in two steps. First, a backbone CNN converts the input image to a saliency map which highlights the discriminative regions. A global pooling operator then collapses the saliency map into scalar predictions, which makes the entire model trainable end-to-end. \cite{durand2017wildcat} devised a new pooling operator that performs feature pooling on both spatial space and class space. \cite{wei2018revisiting} augmented the backbone network using convolution filters with varying dilation rates to address scale variation among object classes. \cite{zhu2019learning} refined segmentation masks using pseudo-supervision from noisy segment proposals. 

Weakly supervised object detection (WSOD) has become increasingly popular in the field of medical image analysis as it eliminates the reliance of models on segmentation labels which are often expensive to obtain. WSOD has been broadly utilized in medical applications including disease classification~\citep{yao2018weakly,liu2019align}, cell segmentation~\citep{li2019weakly,yoo2019pseudoedgenet}, and lesion detection~\citep{xu2014weakly,luo2019deep,wu2019weakly}. \cite{schlemper2019attention} designed a novel attention gate unit that can be integrated with standard CNN classifiers to localize objects of interest in ultrasound images. \cite{ouyang2019weakly} proposed a spatial smoothing regularization to model the uncertainty associated with the segmentation mask. \cite{kervadec2019constrained} demonstrated that regularization terms stemming from inequality constraints can significantly improve the localization performance of a weakly supervised model. While many works still rely on weak localization labels such as point annotations~\citep{yoo2019pseudoedgenet} and scribbles~\citep{ji2019scribble} to produce saliency maps, our approach requires only image-level labels that indicate the presence of an object of a given class. In addition, to make an image-level prediction, most existing models only utilize global information from the saliency maps which often neglect fine-grained details. In contrast, our model also leverages local information from ROI patches using a dedicated network. In Section \ref{sec:ablation_study}, we empirically demonstrate that the ability to focus on fine visual detail is important for classification. 

\section{Discussion and Conclusion}
\label{sec:discussion}

\begin{figure}
    \centering
\includegraphics[width=1.0\columnwidth,trim=3mm 0mm 7mm 0mm]{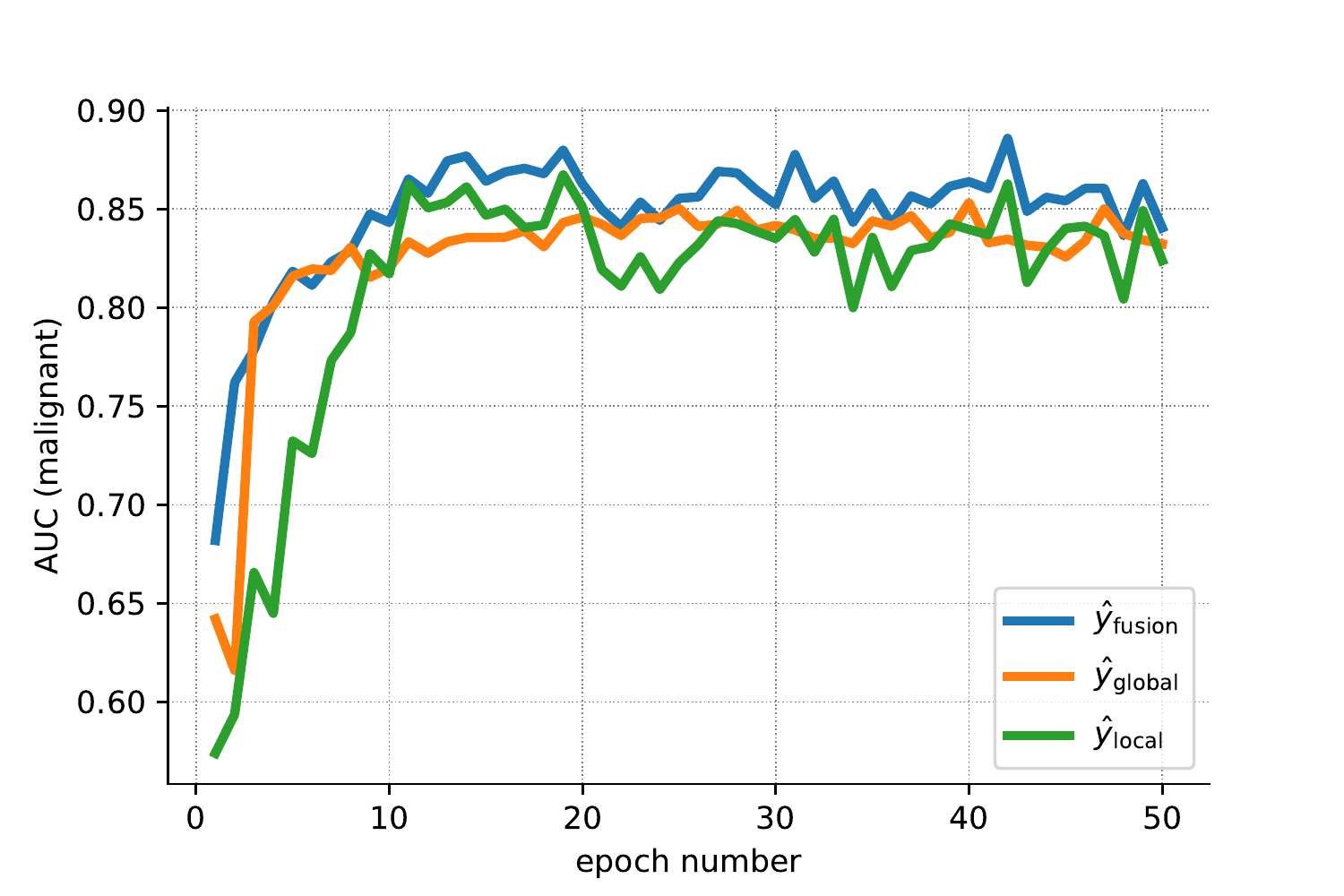}
    \caption{Learning curves for a GMIC-ResNet-18 model. The AUC for malignancy prediction on the validation set is shown for $\hat{\mathbf{y}}_\text{fusion}$, $\hat{\mathbf{y}}_\text{global}$, and $\hat{\mathbf{y}}_\text{local}$.}
  \label{fig:learning_curve}
\end{figure}

Medical images differ from typical natural images in many ways such as much higher resolutions and smaller ROIs. Moreover, both the global structure and local details play essential roles in the classification of medical images. Because of these differences, deep neural network architectures that work well for natural images might not be applicable to many medical image classification tasks. In this work, we present a novel framework, GMIC, to classify high-resolution screening mammograms. GMIC first applies a low-capacity, yet memory-efficient, global module on the whole image to extract the global context and generate saliency maps that provide coarse localization of possible benign/malignant findings. It then identifies the most informative regions in the image and utilizes a local module with higher capacity to extract fine-grained visual details from the chosen regions. Finally, it employs a fusion module that aggregates information from both global context and local details to produce the final prediction.

Our approach is well-suited for the unique properties of medical images. GMIC is capable of processing input images in a memory-efficient manner, thus being able to handle medical images in their original resolutions while still using a high-capacity neural network to pick up on fine visual details. Moreover, despite being trained with only image-level labels, GMIC is able to generate pixel-level saliency maps that provide additional interpretability.

We applied GMIC to interpret screening mammograms: predicting the presence or absence of malignant and benign lesions in a breast. Evaluated on a large mammography dataset, the proposed model outperforms the ResNet-34 while being \textbf{4.3x} faster and using $\mathbf{76.1\%}$ fewer memory of GPU. Moreover, we also demonstrated that our model can generate predictions that are as accurate as radiologists, given equivalent input information. Given its generic design, the proposed model could be widely applicable to various high-resolution image classification tasks. In future research, we would like to extend this framework to other imaging modalities such as ultrasound, tomosynthesis, and MRI.

In addition, we note that training GMIC is slightly more complex than training a standard ResNet model. As shown in Figure~\ref{fig:learning_curve}, the learning speeds for the global and local module are different. As learning of the global module stabilizes, the saliency maps tend to highlight a fixed set of regions in each example, which decreases the diversity of patches provided to the local module. This causes the local module to overfit, causing its validation AUC to decrease. We speculate that GMIC could benefit from a curriculum that optimally coordinates the learning of both modules. A learnable strategy such as the one proposed in~\cite{katharopoulos2019processing} could help to jointly train both global and local module.

\subsubsection*{Acknowledgments}
The authors would like to thank Joe Katsnelson, Mario Videna and Abdul Khaja for supporting our computing environment. We also gratefully acknowledge the support of Nvidia Corporation with the donation of some of the GPUs used in this research. This work was supported in part by grants from the National Institutes of Health (R21CA225175 and P41EB017183).

\bibliographystyle{model2-names}
\biboptions{authoryear}
\bibliography{gmic}

\clearpage


\begin{figure*}[ht]
  \centering
 \includegraphics[width=0.98\textwidth, trim=0 0 0 20]{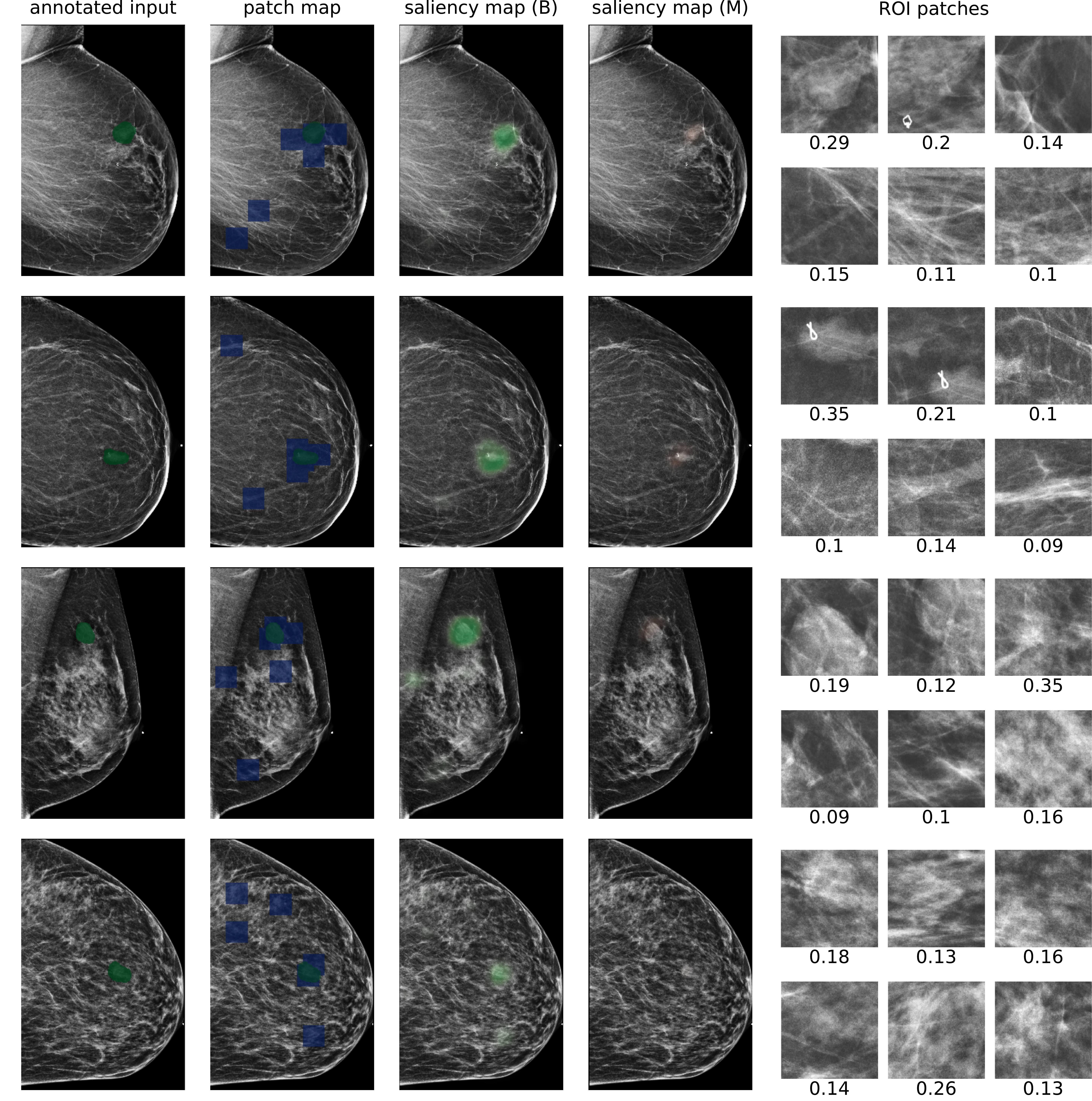}
  \caption{Additional visualizations of benign examples. We follow the same layout as described in Figure~\ref{vis_plot}. Input images are annotated with segmentation labels (green=benign, red=malignant). ROI patches are shown with their attention scores.}
 \label{fig:add_vis_ben}
\end{figure*}

\begin{figure*}
  \centering
 \includegraphics[width=0.94\textwidth, trim=50 0 0 20]{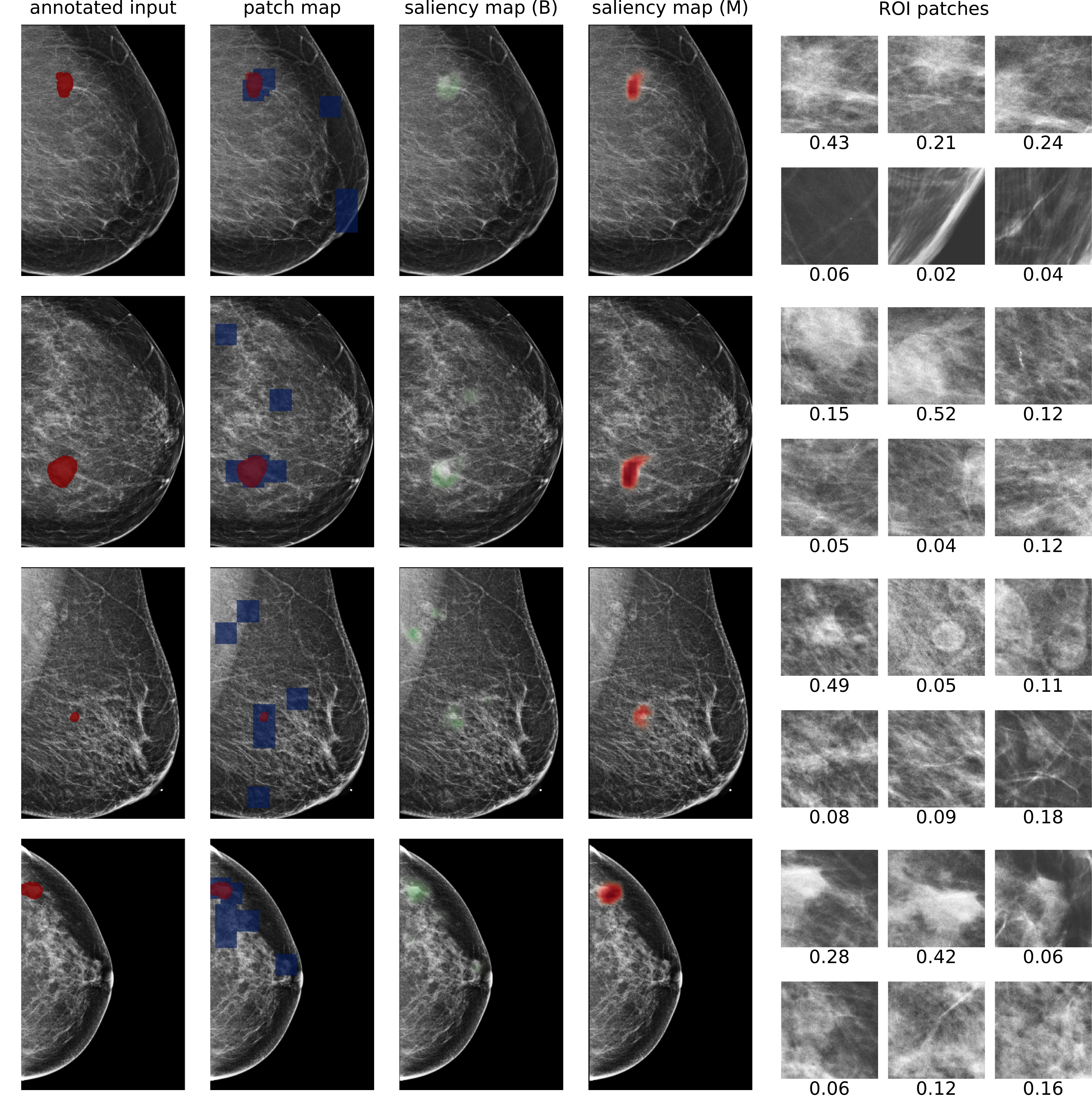}
  \caption{Additional visualizations of malignant examples. We follow the same layout as described in Figure~\ref{vis_plot}. Input images are annotated with segmentation labels (green=benign, red=malignant). ROI patches are shown with their attention scores.}
      \label{fig:add_vis_mal}
\end{figure*}

\end{document}